\documentclass[11pt]{article}
\usepackage[english]{babel}
\usepackage[normalem]{ulem}

\usepackage[dvipsnames,table]{xcolor}
\usepackage{tikz}
\usepackage{pgfplots}
\usepgfplotslibrary{fillbetween}
\pgfplotsset{compat=1.18}
\definecolor{dustygreen}{RGB}{92, 130, 98}
\definecolor{denimblue}{RGB}{36, 92, 155}
\definecolor{deepblue}{RGB}{26, 56, 106}
\definecolor{burntpeach}{RGB}{207, 122, 76}
\definecolor{coolgray}{RGB}{112, 128, 144}
\definecolor{denimlight}{RGB}{94, 146, 196}

\usepackage{makecell}

\usepackage{changepage}

\usepackage{bigints}
\usepackage[letterpaper,margin=1in,bottom=1.06in,includefoot]{geometry}
\usepackage{amsmath,amsthm,bbm,graphicx}

\usepackage{eqnarray}
\usepackage{amsmath}
\usepackage{mathtools}
\usepackage{graphicx}
\usepackage{centernot}
\usepackage{bbm}

\usepackage{mathrsfs}
\usepackage{amsthm}

\usepackage{setspace}

\usepackage[utf8]{inputenc}
\usepackage[T1]{fontenc}
\usepackage{newtxtext}
\usepackage{newtxmath}
\usepackage{mathtools}
\usepackage[thinc]{esdiff}

\usepackage{hhline}

\usepackage{natbib}
\bibpunct[, ]{(}{)}{,}{a}{}{,}%

\newtheorem{theorem}{Theorem}[section]
\newtheorem{lemma}[theorem]{Lemma}

\newtheorem{corollary}[theorem]{Corollary}
\newtheorem{proposition}[theorem]{Proposition}
\newtheorem{assumption}[theorem]{Assumption}

\newcounter{property}

\newcommand\cvar[0]{\text{CVaR}}
\newcommand\opt[0]{\text{OPT}}
\newcommand{\Var}[0]{\text{Var}}

\newcommand{\VaR}{\mathrm{VaR}}

\newcommand{\RegretCVaR}[2]{\mathcal{R}_{\delta,n}\left(#2, #1\right)}

\newcommand{\Pcomp}[2]{\mathrm{PS}_{\delta,n}\!\left(#1, #2\right)}
\newcommand{\SRcomp}[2]{\mathrm{SR}_{\delta,n}\!\left(#1, #2\right)}

\newcommand{\Qc}[2]{Q_{#1}\!\left(#2\right)}
\newcommand{\qv}[2]{q_{#1}\!\left(#2\right)}
\newcommand{\phiv}[3]{\phi_{#1}\!\left(#2, #3\right)}
\newcommand{\xopt}[1]{x^{*}_{#1}}

\newcommand{\rate}{r_n}
\newcommand{\Kprice}{K}

\newcommand{\disp}[2]{\mathcal{D}_{#1}\!\left(#2\right)}
\newcommand{\dispA}[3]{\mathcal{D}_{#1}\!\left(#2 \,;\, #3\right)}

\newtheorem{definition}[theorem]{Definition}
\newtheorem{example}[theorem]{Example}

\usepackage{enumitem}
\counterwithin{equation}{section}
\usepackage[linktocpage=true,pagebackref=true,colorlinks=true,allcolors=blue,bookmarks=true,bookmarksopen,bookmarksnumbered]{hyperref}

\newif\ifshowextra
\showextratrue

\title{\texorpdfstring{How Risky Is Your Data-Driven Policy?}{How Risky Is Your Data-Driven Policy?}}
\author{Omar Besbes and Alon Rieger}
\date{}

\begin{document}

\onehalfspacing

\maketitle

\begin{abstract}

Data-driven algorithms are ubiquitous across industries, and there is an extensive literature on analyzing their \emph{average} performance. The risk profile of such algorithms, while critical, is much less understood. 
This paper quantifies the risk of data-driven policies and isolates its sources.
We evaluate a policy through the conditional value-at-risk (CVaR) of its loss over the training sample, any policy randomization, and the future outcome, at a fixed unknown distribution. Regret against an oracle that knows the distribution decomposes into expected post-sample regret (PS), the average outcome-only CVaR gap of the selected action, and a sample-risk premium (SR), the additional tail risk omitted by averaging those evaluations. The analogous premium vanishes under expected loss.

We characterize when the premium vanishes and how large it is relative to post-sample regret. For losses with a continuous density near the tail cutoff, we identify conditions under which both components decay at the same rate, yet the limiting ratio of the premium to post-sample regret can reach the thousands. When a probability mass at the cutoff moves with the action, the premium can decay more slowly and dominate post-sample regret.  We establish conditions under which sample average approximation learns the oracle action at rate $n^{-1/2}$ in both settings, but its combined-risk regret is of order $n^{-1}$ in the first and $n^{-1/2}$ in the second. In finite-sample numerical examples of SAA in both settings, the premium accounts for approximately $74\%$--$97\%$ of regret.

\end{abstract}

\textbf{keywords}:  data-driven decisions, conditional value-at-risk, sample average approximation, regret decomposition

\section{Introduction}

Data-driven decision procedures are now deployed at scale. A retailer fits an order quantity to last year's sales for each of thousands of products; a cloud provider sizes capacity from recent usage; an insurer sets a retention from a window of historical claims. Each procedure is retrained and redeployed many times, and in each deployment the realized loss depends on two things: the data the procedure was trained on, and the outcome it then faces. 
Under an expected-loss objective, averaging performance conditional on the training sample recovers the overall expected loss. Under a tail-risk objective, averaging conditional risk evaluations need not recover the risk of the learning procedure. We ask how large this difference can be and when it becomes a leading component of regret.

Tail risk is the natural criterion when the costly events are rare: a stockout that loses a customer, a capacity shortfall that breaches a service level, a reserve that falls short of a claim. Conditional Value-at-Risk ($\cvar$), the average loss in the worst $(1-\delta)$ fraction of cases, is a standard expression of that criterion. When the distribution of the outcome is known, the risk-averse problem is well understood. The action is fixed in advance, and the only way to be unlucky is to draw a bad outcome.

Learning the action from a finite sample changes what unlucky means. In the retailer's case, a slow year produces a low order, and an ordinary week then arrives as a stockout. The demand was not unusual; the sample was. Conversely, a week of unusually high demand does little harm when the sample ran high as well and the order already covers it. For a learned policy, large losses can reflect a mismatch between the sample that chose the action and the outcome that tests it, and any randomization inside the policy is one more draw of the same kind. 
If risk means being unlucky, the unlucky event is the combined draw of the sample, any randomization, and the future outcome. We call the $\cvar$ of the loss over that combined draw the \emph{combined-risk} $\cvar$, following the total $\cvar$ of \citet{LuShanthikumarShen2015}. It is what a practitioner estimates by pooling realized losses across deployments, as opposed to the within-deployment $\cvar$ of the action a particular sample happened to produce.

The question becomes precise once the regret is split. The regret of a learning policy against an oracle that knows the distribution is the sum of two nonnegative pieces. The first is the \emph{expected post-sample regret}: the gap between the outcome-only $\cvar$ of the selected action and the oracle value, averaged over samples and any policy randomization. This component already captures the effect of learning on the quality of the selected action. It is the quantity that most of the data-driven optimization literature studies. 
The second is a \emph{sample-risk premium}: the difference between combined-risk $\cvar$ and the average outcome-only $\cvar$ of the selected actions. The analogous premium is zero under an expected-loss criterion. Under $\cvar$, it can dominate regret even when the selected actions are nearly optimal. The premium can be the whole of the regret. Section~\ref{sec:decomposition} gives a two-action example in which every action the policy returns is optimal, so expected post-sample regret is zero, while the combined-risk evaluation still pays a strictly positive premium. The training sample selects different actions on different draws.

Three questions organize the paper. When is the premium zero, and what exactly is charged when it is not? How do the two pieces scale with the sample size, and which features of the loss, the distribution, and the policy set their relative order? And how does the standard policy, sample average approximation, fare under the combined criterion? Each question is about the learning procedure rather than about any one action it produces. The usual assessment of a data-driven policy fixes the sample and evaluates the action it induced; that assessment sees the first piece and not the second. We answer the three questions for a general loss and a scalar action, at a fixed unknown distribution with no parametric family and no prior. 

\subsection{Main contributions}

\emph{Quantifying the sample-risk premium.} 
Building on total $\cvar$ and known conditional-risk comparisons, discussed in Section~\ref{sec:related}, we quantify the premium relative to oracle regret for scalar decisions under a general loss, derive its sharp smooth and atomic expansions, and establish the resulting regret rates for sample average approximation. The premium measures the additional tail risk from evaluating those losses together, under one threshold shared across samples, beyond the average risk of the selected actions. It vanishes exactly when there is a threshold that minimizes the variational objective for almost every selected action, and therefore when the action is fixed; under expectation the analogous premium vanishes for every policy. The expected post-sample regret vanishes when the returned action is an oracle action. 
A nonasymptotic upper bound relates the premium to the dispersion of the induced loss cutoffs. The local results below then identify when the premium is comparable to, dominates, or persists beyond post-sample regret.

\emph{The order and magnitude of each component.}
We determine how the two pieces scale with the sample size, and which features of the loss, the distribution, and the policy set their relative order. We characterize four representative regimes at a fixed distribution, under their respective regularity and policy-concentration conditions. The smooth regime is the one in which the cutoff moves to first order with the action and the loss has a density there. 
Both components are then of order $r_n^2$ when the action concentrates at rate $r_n$. The premium is an additional variance penalty: a common tail threshold across samples gives a larger local action curvature than a threshold re-optimized for each selected action. The ratio of these curvatures is $1+K$. For centered limiting fluctuations, the ratio of the premium to expected post-sample regret tends to $K$. For squared loss with lognormal outcomes, Table~\ref{tab:K_delta} gives $K\approx3900$ at $\delta=0.99$ and $K\approx50000$ at $\delta=0.999$. These are asymptotic coefficients; they do not specify the premium at a given sample size. The atomic regime is the one in which an atom sits at the cutoff and moves with the action, and there the expected post-sample regret is of order $r_n^2$ and the premium is of order $r_n$. The mixing regime is the one in which the policy keeps visiting several optimal actions with no common minimizing threshold, and there the premium tends to a positive constant. The displaced regime is the one in which the policy stays off a unique well-separated optimum, and there the expected post-sample regret is bounded away from zero.

\emph{Sample average approximation.}
Under the conditions of Proposition~\ref{prop:saa_concentration_rate}, SAA learns the oracle action at rate $n^{-1/2}$ in both regimes. Nevertheless, its combined-risk regret is $\Theta(n^{-1})$ in the smooth regime and $\Theta(n^{-1/2})$ in the atomic regime, where the premium dominates. We identify the limiting variance of the action and the resulting atomic premium constant. Finite-sample examples show that the premium accounts for most regret in both regimes: Monte Carlo estimates give shares of about $74\%$--$94\%$ in bounded smooth squared-loss examples, while calculations from the exact distribution of the uniform-demand newsvendor order give shares of $76\%$ at $n=110$ and $95\%$ at $n=1630$. These results show why the action's learning rate alone does not describe the policy's tail-risk performance.

\subsection{Related literature}\label{sec:related}
Coherent risk measures~\citep{ArtznerDEH99} and the variational representation of $\cvar$~\citep{RockafellarU02} are the language of the papers below. 
The closest connections concern total-$\cvar$ and risk estimation, conditional-risk and quantile-score decompositions, Bayesian and contextual risk, and empirical optimization.

\paragraph{The total-CVaR perspective.}
A plug-in evaluation treats the estimated distribution as known, so the risk it reports is the risk of the future outcome alone. The papers below bring the risk contributed by the data into the evaluation. \citet{LuShanthikumarShen2015} do so by changing the objective. Under the name total $\cvar$, they apply $\cvar$ to the profit of the historical sample and next-period demand taken together, explicitly contrast that criterion with expected conditional $\cvar$, and optimize operational statistics for it, for a repeated newsvendor with parametric location-scale demand. Building on this criterion, we compare the sample-risk premium with oracle regret for scalar decisions under a general loss, derive its smooth and atomic expansions, and establish the resulting regret rates for sample average approximation.

\citet{PiteraSchmidt2018} show that plug-in estimators of value-at-risk and expected shortfall can underestimate risk. They correct it by an acceptance criterion: the reported position has to be acceptable as a risk capital. \citet{FrancioniHerzog2012} require that the next observation exceed the estimated quantile with the nominal probability, for every value of the parameter. The quantile of the fitted distribution fails that requirement, and for the normal distribution a distortion that uses the sampling distribution of the estimate restores it. \citet{GerrardTsanakas2011} study how often a threshold set to a nominal tail probability fails, as in a capital or inventory rule. Parameter error can raise that frequency. For increasing transformations of location-scale families the frequency can be computed without knowing the true parameter. A tighter nominal level for the plug-in threshold and a threshold taken from the Bayesian predictive distribution both achieve a prescribed failure probability; under the prior proportional to $1/\sigma$, the predictive threshold meets the nominal level exactly. \citet{BignozziTsanakas2016} subtract the estimated capital from the loss and measure the risk that remains. That remainder is the further capital that makes the position acceptable. Rules based on the parametric bootstrap and on the predictive distribution reduce it, and for location-scale families a capital estimator can remove it. 

The risk-estimation papers above calibrate data-dependent capital, quantile, or threshold rules to a nominal standard, accounting for sampling randomness in that calibration. Our analysis uses the regret decomposition to quantify how much of the excess risk comes from the premium, including when the selected actions approach an oracle action. 

The premium also differs from the optimizer's curse: selection using noisy value estimates can make the selected alternative's estimated performance optimistic~\citep{SmithWinkler2006}. Here the premium is a difference between true population risk evaluations of a policy, and can be positive even when every selected action is optimal.

\paragraph{Conditional risk and quantile-score decompositions.}
The nonnegativity of the premium follows from known conditional-risk comparisons. For a coherent distortion risk measure, the risk of an outcome is at least the average of the risks computed after extra information \citep{IancuPetrikSubramanian2015}. Expected shortfall has the same property because it is concave under mixtures \citep{EmbrechtsMaoWangWang2021,PertaiaUryasev2019,Tselishchev2019}, and the same comparison follows for optimized certainty equivalents from their infimum-of-expectations representation, under the required integrability conditions~\citep{BenTalTeboulle2007}. Taking the information to be the action makes the difference the premium. The same difference is $(1-\delta)^{-1}$ times the resolution term of a quantile score, the gain from letting the cutoff depend on side information \citep{BentzienFriederichs2014,EhmOvcharov2017,Pohle2020}. For a fixed policy, this identity follows by taking its realized loss as the forecast target, its action as side information, and the standard pinball loss as the quantile score. The local behavior of quantile-score excess, including quadratic and linear growth under different distributional conditions, is also studied in \citet{SteinwartChristmann2011}. 
Our analysis relates the premium to the movement of the loss cutoff, the curvature of the outcome-only objective, and the dispersion of the learned action. Comparing it with post-sample regret yields the smooth and atomic orders and the coefficient $K$ under the respective regularity conditions. The mixing regime further identifies when the premium remains positive as post-sample regret vanishes. 

\paragraph{Composite and Bayesian risk.}
Bayesian approaches condition on the observed sample and represent uncertainty about the outcome distribution through a posterior. Some apply an outer risk measure to a parameter-dependent performance criterion; others evaluate risk under the posterior predictive distribution of future outcomes. These constructions differ in how they aggregate parameter and outcome uncertainty.

\citet{QianWangWen2019} nest two risk measures. The inner measure prices the decision when the parameter of the outcome distribution is known. The outer measure prices the fact that the parameter must be estimated. Stochastic programming, robust optimization, and distributionally robust optimization are the cases in which one of the two measures is an expectation or a worst-case value. \citet{LinRenZhou2022} apply a risk functional, nested over the Bayesian posterior of an unknown transition distribution, to the expected total cost of a finite-horizon Markov decision process, and solve the $\cvar$ case by an approximate dynamic program. \citet{MaChenXu2026} use the same nesting for stochastic control: the inner risk is conditional on a latent environment parameter, the outer risk is taken under the posterior, and the action may depend on that posterior. \citet{BodnarLindholmNiklassonThorsen2020,BodnarLindholmNiklassonThorsen2022} and \citet{KimKang2021} compute portfolio value-at-risk and $\cvar$ from the posterior predictive distribution of returns, so the allocation is a function of the observed sample. \citet{JaiswalHonnappaRao2023} replace the expectation in the Bayes risk by an entropic risk over the posterior and give rates for the approximate posterior and for the decision, with the newsvendor among the examples. \citet{SadatsafaviLeeGustafson2022} measure the value of a larger development sample by the net-benefit gap between the predictions of a fitted risk model and the correct predictions. 

Our evaluation takes a frequentist perspective before the sample is observed. The distribution $F$ is fixed and unknown, and no prior is placed on it. Combined-risk $\cvar$ is the tail risk of the loss under the combined draw of the sample given $F$, the action that sample induces, and the future outcome. 
The distinction is in the conditioning: posterior and predictive evaluations condition on the observed sample, whereas combined-risk $\cvar$ evaluates repeated draws of the sample and the future outcome at the same fixed $F$. It is distinct from the nested value that first replaces each action by its outcome-only $\cvar$ and then takes the tail of that number: either one can be larger.

\paragraph{Contextual risk.}
\citet{TaoDelageXu2025} study when a nested assessment of contextual and outcome uncertainty agrees with one risk evaluation of the two sources taken together. Their revised analysis also considers integrated learning and optimization. Here the information is a training sample, independent of the future outcome at a fixed $F$, and the distribution of the learned action changes with the sample size. Our focus is on regret against a distribution-knowing oracle and on how its post-sample and sample-risk components compare as the sample grows.

\paragraph{Outcome-only empirical optimization.}
A large literature approximates an unknown distribution by the sample and then evaluates risk on future outcomes only. Sampling variability enters as estimation error. For $\cvar$ objectives, \citet{MengSunGoh2010} show that under continuous distributions, sample-average optimal values converge to the population optimum and limit points of sample-average stationary points are stationary for the true problem. \citet{Shapiro13} shows that, under mild moment and regularity conditions, limit points of sample-average solutions lie in the set of outcome-only optima. \citet{DentchevaPenevRuszczynski2017} and \citet{GuiguesKS18} refine this analysis with limit theorems for the estimation error, and \citet{BanholzerFW22} give rates of convergence in mean under Lipschitz and moment conditions. Data-driven inventory applications include expected-profit optimization with an empirical value-at-risk constraint~\citep{LinChenLiShen2022}, support-vector formulations of $\cvar$ newsvendor criteria~\citep{Chen2024}, a feature-based $\cvar$ newsvendor~\citep{LiuZhu2024}, and high-probability near-optimality guarantees for risk-averse multiperiod inventory~\citep{JiangTYZ26}. Across this literature, the guarantees concern an optimizer of an outcome-only risk. \citet{LeeParkShin2020} give high-probability excess population-risk guarantees for empirical optimized-certainty-equivalent minimization, including $\cvar$, while \citet{SomaYoshida2020} directly bound expected excess population $\cvar$. These results control the post-sample component, directly or after averaging over training samples. Such guarantees account for sampling variability through the population risk of the selected action. They do not by themselves characterize the additional sample-risk premium, which can determine the order of combined-risk regret. Our SAA analysis connects the concentration of the learned action to the rates of both components.

Related work studies the distribution of cumulative regret in sequential learning. \citet{FanG25} and \citet{SimchiLeviZhengZhu2024} show how expected-regret performance can conceal substantial tail risk arising from the learning process. Here we study the loss of a learned action on an independent future outcome and decompose its combined-risk CVaR regret into expected post-sample regret and a sample-risk premium.

The companion paper~\citep{BesbesRiegerNewsvendor} specializes the decomposition and the asymptotic regimes developed here to the newsvendor. It studies worst-case regret over bounded demand distributions and optimizes policies within a class based on order statistics. For every fixed $\delta\in(0,1)$, the minimax rate over all data-driven policies is $\Theta(n^{-1/4})$, while at $\delta=0$ it is $\Theta(n^{-1/2})$, and uniform windows of width $n^{3/4}$ attain the rate $\Theta(n^{-1/4})$. The present paper develops the structural and local asymptotic analysis for a general loss at a fixed distribution.

\section{Problem Formulation}\label{sec:problem}

We consider a stochastic optimization problem with a nonempty Borel action set $\mathcal{X}\subseteq\mathbb{R}$. The decision maker chooses $x\in\mathcal{X}$ before a random outcome $\xi\in\mathbb{R}$ is realized. We use $c(x,\xi)$ to denote the loss incurred by the decision maker.

\paragraph{CVaR.}
We adopt Conditional Value-at-Risk ($\cvar$) as our performance criterion, a tail-risk measure that focuses on large losses.
\begin{definition}[CVaR]
For $\delta \in (0,1)$ and an integrable random variable $X$ with CDF $F_X$, define
\begin{equation}\label{eqn:cvar_definition}
\cvar_\delta[X] ~~=~~ \frac{1}{1-\delta} \cdot \int_{\delta}^{1} F_X^{-1}\left(u\right)\,du \ .
\end{equation}
Here, $F_X^{-1}(u) = \inf\{y: F_X(y) \geq u\}$ is the $u^{th}$ quantile of $X$.
\end{definition}
The Value-at-Risk at level $\delta$, denoted by $\mathrm{VaR}_\delta[X]$, is the
corresponding quantile,
\[
    \mathrm{VaR}_\delta[X] := F_X^{-1}(\delta).
\]
Equivalently, it is the smallest loss threshold $t$ such that
$\mathbb{P}(X\leq t)\geq\delta$.

Throughout, $\delta\in(0,1)$. The excluded endpoint $\delta=0$ is the risk-neutral
case, $\cvar_0[X]=\mathbb{E}[X]$, in which the criterion is the expectation and the
premium studied below vanishes identically by the tower property.

Informally, $\cvar$ is the expected loss conditional on the loss lying in the worst $(1-\delta)$-tail of its distribution. When the distribution has no probability mass at the $\delta^{th}$ quantile, CVaR admits the following {commonly studied} expression $\cvar_\delta[X] = \mathbb{E}[X \mid X \geq F_X^{-1}\left(\delta\right)]$. When there is an atom at the cutoff, the definition includes only the fraction of that atom needed to obtain a tail of probability $1-\delta$.

An equivalent representation of $\cvar$ was shown by~\citet[Theorem 10]{RockafellarU02}:
\begin{equation}\label{eqn:cvar_representation}
    \cvar_\delta[X] ~~=~~ \inf_{t \in \mathbb{R}}\left\{t + \frac{1}{1-\delta} \cdot \mathbb{E}\left[\left(X-t\right)^+\right] \right\}  \ .
\end{equation}
For an integrable $X$ and $\delta\in(0,1)$, the infimum is attained at $t=\VaR_\delta[X]$. If several thresholds attain it, this Value-at-Risk is the smallest of them. We call that threshold the cutoff of the loss, and the cutoff of an action $x$ is the cutoff of $c(x,\xi)$.

The same representation applies when the loss is generated by a random action. Below we evaluate it over the training sample, any policy randomization, and the future outcome together.

\paragraph{Data and policies.}
In the present paper, we consider the case in which the distribution $F$ is \emph{unknown} to the decision maker, but the decision maker has access to $n$ i.i.d.\ samples $\xi_1^n := (\xi_1,\ldots,\xi_n)$ from $F$. 
We consider a class $\Pi$ of \emph{data-driven policies}, each a measurable probability kernel from $\mathbb{R}^n$ to $\mathcal{X}$. A policy maps the sample to a distribution in $\mathcal{P}(\mathcal{X})$, the set of probability distributions on $\mathcal{X}$. We suppress dependence on $n$ in the policy notation. In particular, a policy $\pi\in\Pi$ specifies
\[
    \pi:\xi_1^n \mapsto H_{\xi_1^n} \ , 
\]
where $H_{\xi_1^n}\in\mathcal{P}(\mathcal{X})$ is the distribution of the action conditional on the sample.
For an observed sample $\xi_1^n$, the distribution $\pi(\xi_1^n)$ specifies a possibly randomized action. We identify a deterministic policy with the point mass at its prescribed action. Note that even for a deterministic policy, the resulting decision $x^\pi$ is a random variable because it depends on the random sample $\xi_1^n$. The decision maker commits to the policy before observing the sample.

We let $\xi_1^n \sim F$ generate the sample, $x^\pi \sim \pi(\xi_1^n)$ denote the resulting decision, and $\xi \sim F$ denote the future outcome realization, drawn independently of the sample and of any randomization in the policy; the induced loss random variable is $c(x^\pi, \xi)$.

\paragraph{Oracle benchmark and combined-risk performance.}
When $F$ is known, the decision maker minimizes $\cvar_{\delta,\xi \sim F}[c(x,\xi)]$ over $x$; we will refer to the optimal value as the \emph{oracle} value defined by 
\[   \opt_\delta(F) ~~=~~ \inf_{x \in \mathcal{X}} \cvar_{\delta,\xi \sim F}\left[c\left(x, \xi\right)\right] \ . \]
Here the decision $x$ is deterministic, so $\cvar_{\delta,\xi \sim F}[c(x,\xi)]$ has a single source of randomness---the outcome realization $\xi \sim F$. Independent randomization over actions cannot improve this value, as the comparison in Section~\ref{sec:decomposition} shows. The conditions maintained below are the following.
\begin{assumption}[Maintained conditions]\label{ass:maintained}
The loss $c$ is jointly measurable, with $\mathbb{E}_{\xi\sim F}|c(x,\xi)|<\infty$ for every $x\in\mathcal{X}$ and $\mathbb{E}|c(x^\pi,\xi)|<\infty$, the second expectation over the sample, policy randomization, and future outcome. The future outcome $\xi\sim F$ is independent of the sample and of any policy randomization. The oracle value satisfies $\opt_\delta(F)>-\infty$.
\end{assumption}

It is convenient to name the fixed-action quantities in which the entire analysis below is expressed. For an action $x \in \mathcal{X}$, let
\begin{equation}\label{eqn:variational_objective}
    \phiv{F}{x}{t} ~~:=~~ t + \frac{1}{1-\delta} \cdot \mathbb{E}_{\xi \sim F}\left[\left(c\left(x,\xi\right)-t\right)^+\right]
\end{equation}
denote the variational objective appearing in equation~\eqref{eqn:cvar_representation}, and let
\begin{equation}\label{eqn:q_c_def}
    \Qc{F}{x} ~~:=~~ \cvar_{\delta,\xi \sim F}\left[c\left(x,\xi\right)\right] ~~=~~ \inf_{t \in \mathbb{R}} \phiv{F}{x}{t}
    \ , \qquad
    \qv{F}{x} ~~:=~~ \VaR_{\delta,\xi \sim F}\left[c\left(x,\xi\right)\right]
\end{equation}
denote, respectively, the outcome-only $\cvar$ and the outcome-only $\VaR$ of the loss at the fixed action $x$. The infimum defining $\Qc{F}{x}$ is attained at $t = \qv{F}{x}$. In this notation the oracle value is $\opt_\delta(F) = \inf_{x \in \mathcal{X}} \Qc{F}{x}$, and whenever this infimum is attained we write $\xopt{F}$ for an optimal action. We state $\xopt{F}$ with the subscript when it matters for the understanding, and otherwise write $x^*$. Proposition~\ref{prop:multiple_optima} treats the case of several optimal actions.

In contrast to the oracle, the data-driven setting carries two layers of randomness: the sample $\xi_1^n$, which feeds into the action $x^\pi$, and the independent future outcome $\xi$, both governed by $F$.
Following the total-CVaR perspective of~\citet{LuShanthikumarShen2015}, expressed there as lower-tail CVaR of profit, we evaluate upper-tail CVaR of loss over both sources of randomness.
The tail-risk \emph{value} of a policy under $F$ is therefore
\begin{equation}\label{eqn:policy_joint_cvar}
    \cvar_\delta[c\left(x^\pi, \xi\right)] ~:=~ \inf_{t \in \mathbb{R}}\left\{t + \frac{1}{1-\delta} \cdot \mathbb{E}_{\xi_1^n \sim F, x^\pi \sim \pi(\xi_1^n), \xi \sim F}\left[\left(c\left(x^\pi,\xi\right)-t\right)^+\right] \right\} \ .
\end{equation}
Throughout the paper, $\cvar_\delta[\cdot]$ without a distribution subscript denotes evaluation over the combined randomness of sample, policy, and outcome; outcome-only evaluation at a fixed action is written $\cvar_{\delta,\xi\sim F}[\cdot]$, equivalently $\Qc{F}{\cdot}$.

\paragraph{Lipschitz condition.}
Assumption~\ref{ass:maintained} is in force throughout. The smooth and atomic results in Section~\ref{sec:decomposition} use, in addition, the following Lipschitz condition. There is a nonnegative random variable $L(\xi)$ with $\mathbb{E}[L(\xi)]<\infty$ such that
\begin{equation}\label{eqn:lipschitz}
    \bigl|c(x,\xi)-c(y,\xi)\bigr|
    ~~\leq~~
    L(\xi)\,|x-y|
\end{equation}
for $F$-almost every $\xi$, simultaneously for all $x,y\in\mathcal{X}$. Additional moment and regularity conditions are stated with the relevant results. The SAA results in Section~\ref{sec:saa} use their own stated assumptions.\footnote{Joint measurability makes $x\mapsto\phiv{F}{x}{t}$ measurable for each $t$. Since $t\mapsto\phiv{F}{x}{t}$ is finite and convex, hence continuous, its infimum over $\mathbb{R}$ agrees with the infimum over $\mathbb{Q}$, so $\Qc{F}{\cdot}$ is measurable, and the outcome-only $\cvar$ of the induced action is a random variable.}

We define the regret of a data-driven policy as the nonnegative difference between the $\cvar$ induced by the policy and the oracle value $\opt_\delta(F)$ when the distribution is known
\begin{equation}\label{eqn:regret_cvar_def}
    \RegretCVaR{F}{\pi} ~~=~~  \cvar_\delta[c\left(x^\pi, \xi\right)] - {\opt_\delta(F)}  \ .
\end{equation}
Nonnegativity follows from the comparison in Section~\ref{sec:decomposition}. This regret evaluates the learning procedure before the sample and any policy randomization are realized. The tail of the combined loss depends on both the sample-induced action and the future outcome, which remain independent. Variation across samples can increase combined-risk $\cvar$ even when every selected action is optimal under its outcome-only $\cvar$.

For an operational interpretation, consider independent deployments under the same distribution $F$ and policy $\pi$, with the sample size $n$ fixed. Each deployment draws a fresh training sample, selects an action, and records the loss from an independent future outcome. As the number of deployments grows, the empirical $\cvar$ of these losses converges to the combined-risk value in equation~\eqref{eqn:policy_joint_cvar}. If instead the realized sample and action are held fixed and outcomes are independently redrawn from $F$, the empirical $\cvar$ converges to the outcome-only value $\Qc{F}{x^\pi}$.

\section{Decomposition of combined-risk regret}\label{sec:decomposition}
After the sample is observed and the policy has selected an action, one may instead evaluate tail risk only over the future realization. Fix $\pi\in\Pi$. Independence of the future outcome gives the \emph{post-sample outcome-only CVaR} $\Psi_n^\pi=\Qc{F}{x^\pi}$, with conditional variational representation
\begin{equation}\label{eqn:conditional_psi}
\Psi_n^\pi
~:=~
\inf_{t \in \mathbb{R}}
\left\{
t+\frac{1}{1-\delta}\,
\mathbb{E}_{\xi \sim F}\!\left[
\left.\left(c\left(x^\pi,\xi\right)-t\right)^+\,\right|\,\xi_1^n,\,x^\pi
\right]
\right\} \ .
\end{equation}
This value is fixed once the action is selected and random across samples and policy randomization.

\paragraph{Comparison and decomposition.}
The following comparison and resulting identity isolate the two components of combined-risk regret. Let $X:=c(x^\pi,\xi)$, and let $t^\ast$ attain the variational representation of $\cvar_\delta[X]$ in equation~\eqref{eqn:cvar_representation}. For every realized $(\xi_1^n,x^\pi)$,
\[
\Psi_n^\pi
~\leq~
t^\ast
+
\frac{1}{1-\delta}\,
\mathbb{E}\!\left[
(X-t^\ast)^+
\,\middle|\,
\xi_1^n,x^\pi
\right],
\]
because the conditional variational formula defining $\Psi_n^\pi$ minimizes over all thresholds, and $t^\ast$ is one feasible threshold. Taking expectations over $(\xi_1^n,x^\pi)$ yields
\begin{equation}\label{eqn:cond_vs_joint}
\mathbb{E}_{\xi_1^n,\,x^\pi}\!\left[\Psi_n^\pi\right]
~\leq~
\cvar_\delta[c\left(x^\pi,\xi\right)].
\end{equation}

Adding and subtracting
$\mathbb{E}_{\xi_1^n,\,x^\pi}[\Psi_n^\pi]$ in~\eqref{eqn:regret_cvar_def} gives
\begin{equation}\label{eqn:regret_psi_decomp}
\RegretCVaR{F}{\pi}
=
\underbrace{\Bigl(
\mathbb{E}_{\xi_1^n,\,x^\pi}\!\left[\Psi_n^\pi\right]
-
\opt_\delta\left(F\right)
\Bigr)}_{\text{$\mathrm{PS}_{\delta, n}(\pi,F)$-- expected post-sample regret}}
+
\underbrace{\Bigl(
\cvar_\delta[c\left(x^\pi,\xi\right)]
-
\mathbb{E}_{\xi_1^n,\,x^\pi}\!\left[\Psi_n^\pi\right]
\Bigr)}_{\mathrm{SR}_{\delta,n}(\pi,F) \text{-- sample-risk premium}}.
\end{equation}
Both brackets are nonnegative. For every realized $(\xi_1^n,x^\pi)$ the induced action is a
feasible deterministic action, so $\Psi_n^\pi\geq\opt_\delta(F)$, and taking expectations
establishes nonnegativity of the first. We call it the \emph{expected post-sample regret}
$\Pcomp{\pi}{F}$: the expected outcome-only tail-risk gap of the selected action relative to the oracle, averaged over samples and policy randomization. The second
bracket is nonnegative by equation~\eqref{eqn:cond_vs_joint}, and it measures something the
first cannot see. 
Combined-risk $\cvar$ evaluates the induced loss over the sample, the policy randomization, and the future outcome together. Its excess over the average outcome-only $\cvar$ can be positive even when every selected action is optimal, as Example~\ref{ex:premium_without_bad_action} below shows. Under expected cost this excess vanishes by the tower property. We call this additional severity the
\emph{sample-risk premium} $\SRcomp{\pi}{F}$.
For independent deployments under the same $F$, $\pi$, and $n$, the average of the outcome-only values $\Psi_n^\pi$ converges to $\mathbb{E}[\Psi_n^\pi]$. The gap between this limit and the combined-risk $\cvar$ is $\SRcomp{\pi}{F}$.

\paragraph{The premium as the price of one shared cutoff.}
The premium admits a form that makes this additional severity explicit. Because the induced action $x^\pi$ is independent of the future outcome $\xi\sim F$, both terms in the premium can be expressed through the fixed-action variational objective $\phiv{F}{x}{t}$ from equation~\eqref{eqn:variational_objective}. Combined-risk $\cvar$ in equation~\eqref{eqn:policy_joint_cvar} optimizes over a single threshold that has to serve every realized action at once, so that it equals $\inf_{t}\mathbb{E}_{x^\pi}[\phiv{F}{x^\pi}{t}]$. 
The post-sample evaluation minimizes the threshold separately for each realized action, giving $\Psi_n^\pi=\inf_t\phiv{F}{x^\pi}{t}=\Qc{F}{x^\pi}$. Subtracting the second from the first,
\begin{equation}\label{eqn:sr_variational}
    \SRcomp{\pi}{F}
    ~=~
    \inf_{t \in \mathbb{R}}\mathbb{E}_{x^\pi}\!\left[\phiv{F}{x^\pi}{t}\right]
    ~-~
    \mathbb{E}_{x^\pi}\!\left[\inf_{t \in \mathbb{R}}\phiv{F}{x^\pi}{t}\right]
    ~=~
    \inf_{t \in \mathbb{R}}\mathbb{E}_{x^\pi}\!\left[\phiv{F}{x^\pi}{t}\right]
    ~-~
    \mathbb{E}_{x^\pi}\!\left[\Qc{F}{x^\pi}\right]
\end{equation}
The premium is the cost of using a single threshold across the induced loss distributions: it is the gap between the infimum of an expectation and the expectation of the pointwise infimum. The threshold is an auxiliary variable in the CVaR representation.
Example~\ref{ex:premium_without_bad_action} makes that gap strict. Both actions are optimal for outcome-only $\cvar$, with minimizing cutoffs $1$ and $0$, so the difference in~\eqref{eqn:sr_variational} is positive and the combined-risk regret equals the premium.

\begin{example}[Premium without a bad action]\label{ex:premium_without_bad_action}
Fix $\delta=1/2$, let $\mathcal X=\{0,1\}$, and let $\xi$ equal $0$ or $2$ with probabilities $3/4$ and $1/4$. Set $c(0,\xi)=1$ and $c(1,\xi)=\xi$. Both actions have outcome-only $\cvar$ equal to $1$, with cutoffs $1$ and $0$. For any $n\ge1$, consider the deterministic policy that selects action $0$ if the first training observation is $2$ and action $1$ otherwise. Against an independent future outcome, its losses $0$, $1$, and $2$ have probabilities $9/16$, $4/16$, and $3/16$. Its combined-risk $\cvar$ is therefore $5/4$: expected post-sample regret is zero, while the sample-risk premium is $1/4$.
\end{example}

The next proposition characterizes when a common threshold minimizes the objective for almost every induced action.

\paragraph{When each component vanishes.}
The following proposition characterizes when each component vanishes. Before we state it, we
write
\[
    \mathcal{T}_\delta(x) ~~:=~~ \arg\min_{t}\phiv{F}{x}{t}
\]
for the set of minimizing thresholds at the action $x$. The conditional tails induced by $x^\pi$ are \emph{aligned} if one threshold lies in $\mathcal{T}_\delta(x^\pi)$ for almost every realized action, and \emph{misaligned} otherwise.

\begin{proposition}[Endpoints of the two components]
\label{prop:endpoints}
Suppose that $x^\pi$ is independent of $\xi\sim F$ and that
\[
    \mathbb{E}_{x^\pi, \xi}|c(x^\pi,\xi)| < \infty
\ . \]
Then the following hold.
\begin{enumerate}
    \item $\SRcomp{\pi}{F}=0$ if and only if the conditional tails induced by
    $x^\pi$ are aligned. In particular, the premium vanishes whenever $x^\pi$
    is degenerate.

    \item Suppose in addition that the infimum defining $\opt_\delta(F)$ is
    attained, and let
    $\mathcal{X}^*_F:=\arg\min_{x\in\mathcal{X}}\Qc{F}{x}$ denote the set of
    oracle actions. Then $\Pcomp{\pi}{F}=0$ if and only if
    $x^\pi\in\mathcal{X}^*_F$ almost surely. If the oracle action is unique,
    this holds if and only if $x^\pi=\xopt{F}$ almost surely.
\end{enumerate}
\end{proposition}
In words: $\mathrm{PS}$ vanishes if and only if the policy plays an oracle action almost surely; $\mathrm{SR}$ vanishes if and only if the induced loss distributions admit a common minimizing threshold, and a degenerate action is always aligned. Distinct selected VaR values can still yield zero premium when their sets of minimizing thresholds overlap. The proof is in Appendix~\ref{prf:endpoints}.
A complementary bound controls the premium at every sample size through the distribution of the induced cutoff.
\begin{proposition}[An upper bound on the sample-risk premium]\label{prop:premium_ceiling}
Under the independence and integrability hypotheses of Proposition~\ref{prop:endpoints}, let $Y=q_F(x^\pi)$ and $L_\delta=\max\{1,\delta/(1-\delta)\}$. Then $Y$ is integrable and
\begin{equation}\label{eqn:premium_ceiling}
0\le\SRcomp{\pi}{F}
\le \cvar_\delta[Y]-\mathbb E Y
\le L_\delta\inf_{t\in\mathbb R}\mathbb E|Y-t|.
\end{equation}
\end{proposition}
The proof is in Appendix~\ref{prf:premium_ceiling}. Equality holds between the premium and $\cvar_\delta[Y]-\mathbb E Y$ when the loss is conditionally deterministic. That middle term keeps the asymmetric tail weighting that the absolute-deviation bound discards.
Corollary~\ref{cor:premium_rate} records the rate that follows from the bound: under the Lipschitz condition~\eqref{eqn:lipschitz}, if $x^\pi=x^*+r_nZ_n$, $r_n\to0$, $\sup_n\mathbb E|Z_n|<\infty$, and $q_F$ is differentiable at $x^*$, then $\mathrm{SR}=O(r_n)$. Its proof is in Appendix~\ref{prf:premium_rate}.
The atomic regime below attains this upper order. The rate conclusion uses the cutoff and moment conditions; other orders remain possible outside the regimes below.

We first establish the orders of the two components for policies concentrating on a unique oracle action. We then quantify their leading coefficients in the smooth and atomic regimes, and examine when a component remains positive in the limit.

\subsection{The asymptotic order of the components}
The curvature of the outcome-only objective determines the local cost of action error. The movement of the cutoff and the behavior of the loss distribution around it determine the premium. The following results translate a policy concentration rate $\rate$ into orders of regret under the stated moment and regularity conditions. Section~\ref{sec:saa} establishes such concentration for SAA under additional hypotheses.

\begin{theorem}[Smooth regime]
\label{prop:asymptotic_characterization}
Under Assumption~\ref{ass:maintained}, suppose that equation~\eqref{eqn:lipschitz} holds.
Fix $F$ and $\delta\in(0,1)$, and let $x^*$ be the unique oracle action. Suppose $x^*$ is interior and that $Q_F$ is twice differentiable at $x^*$ with $0<Q_F''(x^*)<\infty$. Write $q^*:=\qv{F}{x^*}$. Suppose $c(x,\xi)$ has a density $g_x(t)$ for $(x,t)$ near $(x^*,q^*)$, that the map $(x,t)\mapsto g_x(t)$ is continuous at $(x^*,q^*)$, and that $0<g_{x^*}(q^*)<\infty$. Suppose the policy concentrates on $x^*$ at rate $\rate$,
\begin{equation}\label{eqn:concentration}
    x^\pi ~~=~~ x^*+\rate Z_n \ , \qquad \rate\searrow0 \ , \qquad Z_n\Rightarrow Z \ ,
\end{equation}
and that $q_F'(x^*)\neq 0$, so the cutoff moves to first order with the action. Suppose $\bigl\{|Z_n|^{2}\bigr\}$ is uniformly integrable and $Z$ is nondegenerate. Then
\[
    \Pcomp{\pi}{F} ~~=~~ \Theta\!\left(\rate^{2}\right) \ ,
    \qquad
    \SRcomp{\pi}{F} ~~=~~ \Theta\!\left(\rate^{2}\right) \ .
\]
\end{theorem}
The proof is in Appendix~\ref{prf:asymptotic_characterization}.
Theorem~\ref{prop:asymptotic_characterization} characterizes the regime in which the policy
concentrates around a unique oracle action. The relative order of the two
regret components is determined by three features: the local growth of the
outcome-only $\cvar$ objective away from the oracle action, the order at which the $\VaR$
moves with the action, and the local growth of the variational objective
around the $\VaR$. When the outcome-only $\cvar$ objective is locally
quadratic, the variational objective is locally quadratic around the $\VaR$,
and the $\VaR$ moves to first order with the action, the two regret components
are of the same order.

A squared loss against a skewed outcome illustrates the smooth regime when the loss density satisfies the theorem's conditions, the cutoff slope is nonzero, and the oracle is unique and interior with positive finite curvature. Table~\ref{tab:K_delta} evaluates the local coefficient for this loss. Symmetry can instead make $q_F'(x^*)=0$, outside the theorem's scope.  Proposition~\ref{prop:smooth_brackets} identifies the limiting ratio under these same hypotheses. For centered limiting fluctuations, the coefficient can be thousands even though the two components have the same order.

Under the moving-atom, straddling, and isolation conditions below, the premium is linear in the concentration scale, while post-sample regret remains quadratic.
\begin{theorem}[Atomic regime]
\label{cor:atom_category}
Under Assumption~\ref{ass:maintained}, suppose that equation~\eqref{eqn:lipschitz} holds.
Let $x^*$ be the unique interior oracle action, and suppose $Q_F$ is twice differentiable there with $0<Q_F''(x^*)<\infty$. Suppose the loss places an atom at a level $\ell(x)$, continuously differentiable near $x^*$, with $\ell(x^*)=q^*:=\qv{F}{x^*}$ and $\ell'(x^*)\neq 0$, and that the atom strictly straddles the tail level,
\[
    \mathbb{P}\bigl(c(x^*,\xi)<q^*\bigr)
    ~<~ \delta ~<~
    \mathbb{P}\bigl(c(x^*,\xi)\le q^*\bigr) \ .
\]
As the action approaches the oracle and the margin around $\ell(x)$ shrinks, the probability strictly above the atom and the probability at or above a level just below it settle at their oracle values,
\begin{equation}\label{eqn:atom_isolation}
\begin{aligned}
    \lim_{\left(x,u\right)\to\left(x^*,0^+\right)}\mathbb{P}\bigl(c\left(x,\xi\right)>\ell\left(x\right)+u\bigr)
    ~&=~
    \mathbb{P}\bigl(c\left(x^*,\xi\right)>q^*\bigr) \ , \\
    \lim_{\left(x,u\right)\to\left(x^*,0^+\right)}\mathbb{P}\bigl(c\left(x,\xi\right)\geq\ell\left(x\right)-u\bigr)
    ~&=~
    \mathbb{P}\bigl(c\left(x^*,\xi\right)\geq q^*\bigr) \ .
\end{aligned}
\end{equation}
These limits are the isolation assumption. Other losses may approach the atom; only the probability in a shrinking band is controlled.
Suppose that the policy concentrates on $x^*$ as in equation~\eqref{eqn:concentration}, with $\bigl\{|Z_n|^{2}\bigr\}$ uniformly integrable and $Z$ nondegenerate. Then
\[
    \Pcomp{\pi}{F} ~~=~~ \Theta\!\left(\rate^{2}\right) \ ,
    \qquad
    \SRcomp{\pi}{F} ~~=~~ \Theta\!\left(\rate\right) \ .
\]
\end{theorem}
The proof is in Appendix~\ref{prf:atom_category}.
The profit formulation of the newsvendor illustrates an atom created by the loss: when demand meets or exceeds the order, the loss is constant at a level that moves with the order. At the oracle, this atom strictly straddles the tail level; continuous demand gives the isolation condition~\citep{BesbesRiegerNewsvendor}. With a unique interior oracle, a demand distribution differentiable there with a positive finite derivative, and a policy satisfying the theorem's concentration and moment conditions, the problem lies in the atomic regime. Multiple optimal orders can instead produce the mixing behavior discussed below. More generally, an atom yields the stated orders only under the full hypotheses of Theorem~\ref{cor:atom_category}.
Under these hypotheses, the sample-risk premium is the leading component of combined-risk regret: its ratio to expected post-sample regret diverges.

\subsection{Mean squared error plus a $K$-weighted variance penalty}\label{sec:fixed_n}

In the smooth regime, the common rate of decay does not determine the relative magnitude of the two components. 
The next proposition shows that $\mathrm{PS}$ is asymptotically proportional to the mean squared error of the action about $x^*$, while $\mathrm{SR}$ is asymptotically proportional to its variance, and identifies the two constants.

\begin{proposition}[Smooth-regime coefficients]\label{prop:smooth_brackets}\label{cor:smooth_brackets}
Suppose that the hypotheses of Theorem~\ref{prop:asymptotic_characterization} hold. Write $q^*:=\qv{F}{x^*}$, $g:=g_{x^*}(q^*)$, $\mathrm{MSE}:=\mathbb{E}[(x^\pi-x^*)^{2}]$, and $V:=\Var(x^\pi)$. Then
\begin{equation}\label{eqn:smooth_brackets}
    \frac{\Pcomp{\pi}{F}}{\mathrm{MSE}} ~~\longrightarrow~~ \frac{1}{2}Q_F''\!\left(x^*\right) \ ,
    \qquad
    \frac{\SRcomp{\pi}{F}}{V} ~~\longrightarrow~~ \frac{g\cdot q_F'\!\left(x^*\right)^{2}}{2\left(1-\delta\right)} \ .
\end{equation}
\end{proposition}
The proof is in Appendix~\ref{prf:smooth_local}.
The ratio of the two constants is the relative variance penalty,
\begin{equation}\label{eqn:K}
    \boxed{\;\Kprice \;=\; \dfrac{g\cdot q_F'\!\left(x^*\right)^{2}}{\left(1-\delta\right)\,Q_F''\!\left(x^*\right)}\;} \ .
\end{equation}
Together, these limits give the local expansion
\begin{equation}\label{eqn:reduction}
    \Pcomp{\pi}{F}+\SRcomp{\pi}{F}
    ~~=~~
    \tfrac{1}{2}Q_F''\!\left(x^*\right)\bigl[\mathrm{MSE}+\Kprice\cdot V\bigr]
    +
    o\!\left(\rate^{2}\right) \ .
\end{equation}
Since $\mathrm{MSE}=\bigl(\mathbb{E}[x^\pi-x^*]\bigr)^2+V$, variance receives weight $1+\Kprice$ relative to squared bias in the leading term.

\paragraph{Why the premium can be large.}
The coefficient compares two local curvatures. Let $A_F$ be the action curvature when the threshold is held at $q^*=q_F(x^*)$. The smooth local expansion gives
\begin{equation}\label{eqn:frozen_curvature}
\text{$\displaystyle \begin{aligned}
\phi_F(x^*+h,q^*)&=Q_F(x^*)+\tfrac12 A_Fh^2+o(h^2),\\
A_F&=Q_F''(x^*)+\frac{g\,q_F'(x^*)^2}{1-\delta}.
\end{aligned}$}
\end{equation}

Thus $1+\Kprice=A_F/Q_F''(x^*)$. Post-sample evaluation allows a separate minimizing threshold for each selected action. Combined evaluation uses one threshold across samples; locally, it adjusts to the mean action but not to each fluctuation. In the leading regret term, squared bias is therefore charged at curvature $Q_F''(x^*)$ and variance at $A_F$. The additional variance curvature in the premium is $A_F-Q_F''(x^*)$. For the squared loss with lognormal outcomes in Table~\ref{tab:K_delta} at $\delta=0.99$, the optimized and fixed-threshold curvatures are approximately $1.85$ and $7234$. Appendix~\ref{prf:frozen_curvature} derives the expansion under the existing smooth-regime hypotheses.

To express the relative magnitude of the two components, write
\begin{equation}\label{eqn:premium_ratio}
    \rho_n
    ~~:=~~
    \frac{\SRcomp{\pi}{F}}{\Pcomp{\pi}{F}}
\end{equation}
whenever $\Pcomp{\pi}{F}>0$. Combined regret is then $(1+\rho_n)$ times $\Pcomp{\pi}{F}$. Under the smooth-regime hypotheses, the ratio converges to a positive finite constant whose magnitude can be large.

\begin{corollary}[The limiting premium ratio in the smooth regime]\label{cor:rho_limit}
Under the hypotheses of Proposition~\ref{prop:smooth_brackets}, suppose in addition that $\mathbb{E}[Z]=0$. Then
\begin{equation}\label{eqn:rho_limit}
    \rho_n ~~\longrightarrow~~ \Kprice \ ,
\end{equation}
with $\Kprice$ the relative variance penalty in equation~\eqref{eqn:K}. Without the centering assumption the limit is $\Kprice\cdot\Var(Z)/\mathbb{E}[Z^{2}]$.
\end{corollary}
The proof is in Appendix~\ref{prf:rho_limit}.
A nonzero mean of the limiting scaled error $Z$ lowers the limiting ratio by the factor $\Var(Z)/\mathbb{E}[Z^2]$. For fixed $(c,F,\delta)$, $\Kprice$ is the largest limiting ratio under Proposition~\ref{prop:smooth_brackets}, attained when $\mathbb{E}[Z]=0$.
Table~\ref{tab:K_delta} evaluates $\Kprice$ for squared loss using the formula in Proposition~\ref{prop:K_squared}. In the tabulated examples, $\Kprice$ increases over the reported tail levels. It reaches the tens of thousands for the lognormal family, while lying below one for the gamma family at lower tail levels.

For centered limiting fluctuations, the premium's share of combined-risk regret converges to $\Kprice/(1+\Kprice)$. Thus the lognormal entry $\Kprice\approx3900$ at $\delta=0.99$ corresponds to a limiting share of approximately $99.97\%$.
The coefficient $\Kprice$ describes a local limit; the results do not specify the sample size at which $\rho_n$ is close to it. Finite-sample accuracy depends on both the spread of the learned action and the neighborhood over which the density and cutoff expansions are accurate.

\begin{table}[ht]
\centering
\small
\begin{tabular}{l rrrrr}
\hline
\textbf{Family}
& $\boldsymbol{\delta=0.80}$
& $\boldsymbol{\delta=0.90}$
& $\boldsymbol{\delta=0.95}$
& $\boldsymbol{\delta=0.99}$
& $\boldsymbol{\delta=0.999}$
\\
\hline
Exponential & $1.56$ & $3.93$ & $8.76$ & $47.9$ & $492$ \\
Gamma & $0.228$ & $0.413$ & $0.637$ & $1.33$ & $2.93$ \\
Lognormal & $34.7$ & $136$ & $428$ & $3900$ & $49700$ \\
Pareto & $4.46$ & $13.4$ & $36.6$ & $330$ & $7200$ \\
Weibull & $6.21$ & $22.1$ & $70.5$ & $884$ & $28700$ \\
\hline
\end{tabular}
\caption{The relative variance penalty $\Kprice$ at the oracle action for $c(x,\xi)=(\xi-x)^{2}/2$, from~\eqref{eqn:K_squared}, with the oracle action and cutoff located numerically. For the limiting-ratio interpretation, take a compact action interval with the oracle in its interior and a policy satisfying Theorem~\ref{prop:asymptotic_characterization} with centered limiting fluctuations. Then the entries are the limit of $\rho_n=\SRcomp{\pi}{F}/\Pcomp{\pi}{F}$. The coefficient is invariant to location--scale changes of the outcome for this squared loss. Exponential has rate $1$; gamma has shape $4$ and scale $1/8$; lognormal has $\log\xi\sim\mathcal N(0,1.8^2)$; Pareto is Type~I with tail index $3$ and scale $1$; Weibull has shape $0.7$ and scale $1$.}
\label{tab:K_delta}
\end{table}

The zero-slope endpoint is different. If the other hypotheses of Theorem~\ref{prop:asymptotic_characterization} hold but $q_F'(x^*)=0$, then $\mathrm{SR}=o(r_n^2)$ while $\mathrm{PS}=\Theta(r_n^2)$. Proposition~\ref{prop:flat_cutoff} records these orders. Its proof evaluates the common threshold at $q^*$. Thus action variability need not create a premium of the same order as post-sample regret.

\subsection{When the cutoff sits on an atom}\label{sec:atom}
The reduction in equation~\eqref{eqn:reduction} prices a smooth cutoff by mean squared error and a variance. 
Under the hypotheses of Theorem~\ref{cor:atom_category}, the unique minimizing threshold near $x^*$ is the atom level, $q_F(x)=\ell(x)$. Hence $q_F'(x^*)=\ell'(x^*)\neq0$, and the variational objective grows linearly away from this threshold. The leading premium is then a weighted absolute deviation of the induced cutoff. This dispersion is invariant under deterministic translations of the scaled cutoff fluctuations, while the leading post-sample term continues to charge mean squared error. Write $p:=\mathbb{P}(c(x^*,\xi)=q^*)$ for the mass at the cutoff, and
\begin{equation}\label{eqn:atom_tau}
    \tau
    ~:=~
    \frac{\delta-\mathbb{P}\bigl(c(x^*,\xi)<q^*\bigr)}{p}
\end{equation}
for the fraction of that mass lying below $\delta$. The fraction above $\delta$ is $1-\tau$. For any integrable random variable $Y$, define $\mathcal{W}(Y):=p\inf_{t\in\mathbb{R}}\mathbb{E}[\tau(Y-t)^++(1-\tau)(t-Y)^+]$. A $\tau$-quantile attains the infimum. Letting $q_\tau$ be such a quantile of the induced $\VaR$, this gives
\begin{equation}\label{eqn:atom_weight}
    \mathcal{W}\!\left(\qv{F}{x^\pi}\right)
    ~~=~~
    p\,\mathbb{E}\!\left[
        \tau\bigl(\qv{F}{x^\pi}-q_\tau\bigr)^{+}
        +(1-\tau)\bigl(q_\tau-\qv{F}{x^\pi}\bigr)^{+}
    \right]
    \ .
\end{equation}

\begin{proposition}[Atom at the cutoff]\label{prop:atom_brackets}\label{cor:atom_brackets}
Suppose that the hypotheses of Theorem~\ref{cor:atom_category} hold, and write $\mathrm{MSE}$ as in Proposition~\ref{prop:smooth_brackets}. Then
\begin{equation}\label{eqn:atom_brackets}
    \frac{\Pcomp{\pi}{F}}{\mathrm{MSE}}
    ~~\longrightarrow~~
    \tfrac{1}{2}Q_F''\!\left(x^*\right)
    \ ,
    \qquad
    \frac{\SRcomp{\pi}{F}}{\mathcal{W}\!\left(\qv{F}{x^\pi}\right)}
    ~~\longrightarrow~~
    \frac{1}{1-\delta}
    \ .
\end{equation}
\end{proposition}
The proof is in Appendix~\ref{prf:atom_brackets}. The induced $\VaR$ moves with the action because $q_F'(x^*)\neq 0$. When the two gaps in the straddling hypothesis are equal, $\tau=1/2$ and $\mathcal{W}\!\left(\qv{F}{x^\pi}\right)$ is $(p/2)$ times the mean absolute deviation of the induced $\VaR$ about a median.
The two limits have different remainders, so they are recorded separately,
\begin{equation}\label{eqn:atom_reduction}
    \Pcomp{\pi}{F}
    ~~=~~
    \tfrac{1}{2}Q_F''\!\left(x^*\right)\,\mathrm{MSE}
    +o(\rate^{2})
    \ ,
    \qquad
    \SRcomp{\pi}{F}
    ~~=~~
    \frac{\mathcal{W}\!\left(\qv{F}{x^\pi}\right)}{1-\delta}
    +o(\rate)
    \ .
\end{equation}
The error in the premium is of order $o(\rate)$. That remainder can exceed the quadratic post-sample term, so the sum is an expansion of total regret up to $o(\rate)$. At small action spreads the linear premium dominates the quadratic post-sample term, the same comparison the atomic regime records asymptotically.

The cutoff expansion holds in first mean: $\qv{F}{x^\pi}-q^*=q_F'(x^*)\,\rate Z_n+o_{L^1}(\rate)$.  Weak convergence and uniform integrability give convergence of the scaled cutoff distributions in the Wasserstein-$1$ metric. The weighted deviation is continuous in this metric, positively homogeneous, and invariant under deterministic translations. Applying it with the same level $\tau$ to $q_F'(x^*)Z$ yields
\begin{equation}\label{eqn:atom_rho}
    \rate\,\rho_n
    ~~\longrightarrow~~
    \frac{2\,\mathcal{W}\!\left(q_F'(x^*)\,Z\right)}{(1-\delta)\,Q_F''\!\left(x^*\right)\,\mathbb{E}[Z^{2}]}
    \ .
\end{equation}

The weights are a property of $(c,F,\delta)$ at the oracle action and carry no sample size. The numerator of $\rho_n$ is homogeneous of degree one and the denominator of degree two, so at a concentration rate $\rate$ the ratio is of order $1/\rate$, and equation~\eqref{eqn:atom_rho} names the constant. The premium therefore outgrows $\Pcomp{\pi}{F}$ in order and not merely in constant, which is what Theorem~\ref{cor:atom_category} records.
For the newsvendor loss under the conditions of Theorem~\ref{cor:atom_category}, the ratio of the premium to expected post-sample regret grows like $1/\rate$.
\citet{BesbesRiegerNewsvendor} specialize Proposition~\ref{prop:atom_brackets} to that loss: they record the limit of $\mathrm{PS}$ over mean squared error and the limit of the premium over a weighted absolute deviation of the order.

\subsection{Regimes in which a component does not vanish}\label{sec:categories}
Uniqueness of $\xopt{F}$, which Theorems~\ref{prop:asymptotic_characterization} and~\ref{cor:atom_category} both assume, excludes a
qualitatively different regime. When several actions attain the oracle value, a data-driven policy
may converge to the set of optimal actions without converging to any single
one of them. In that case, the expected post-sample regret may vanish even
though the induced action remains nondegenerate. If no threshold minimizes the variational objective for almost every limiting action, the sample-risk premium can converge to a strictly positive constant. The following proposition formalizes this possibility.

\begin{proposition}\label{prop:multiple_optima}
Suppose that the unconditional action distribution, over the sample and any policy randomization, satisfies $x^\pi\Rightarrow x_\infty^\pi$, where $x_\infty^\pi$ is supported
on the set of population-optimal actions. 
Suppose further that $x\mapsto c(x,\xi)$ is continuous at $x_\infty^\pi$ almost surely, and that the family
\[
\left\{|c(x^\pi,\xi)|\right\}_{n\geq1}
\]
is uniformly integrable.
Then
\[
\mathrm{PS}_{\delta,n}(\pi,F)\longrightarrow0.
\]
Moreover, if 
\[\mathbb{P}\!\left(t\in\mathcal{T}_\delta(x_\infty^\pi)\right)<1 \qquad\text{for every }t\in\mathbb{R} \ , \]
then
\[
\mathrm{SR}_{\delta,n}(\pi,F)
\longrightarrow
\mathrm{SR}_{\delta,\infty}(\pi,F)>0.
\]
Consequently,
\[
\mathrm{PS}_{\delta,n}(\pi,F)=o(1),
\qquad
\mathrm{SR}_{\delta,n}(\pi,F)=\Theta(1).
\]
\end{proposition}
The proof is in Appendix~\ref{prf:multiple_optima}. 
The cost and $F$ determine the optimal actions and their sets of minimizing thresholds. The policy determines whether its limiting action remains nondegenerate on that optimal set. Under the proposition's hypotheses, this limiting mixture has vanishing post-sample regret and a positive premium precisely when it has no common minimizing threshold. A policy converging to a single optimal action has vanishing premium. For example, in the newsvendor problem, SAA at a particular Bernoulli distribution keeps mixing optimal orders with no common minimizing threshold, and its premium tends to a positive constant. A window of order-statistic indices wider than the $\sqrt{n}$ fluctuations of the binomial count concentrates on one limiting optimal order and eliminates that persistent premium~\citep{BesbesRiegerNewsvendor}.

The displaced regime keeps a unique oracle action and allows persistent displacement from it.

\begin{proposition}
\label{prop:separation_inconsistency}
Fix $F$ and suppose that the population problem has a unique optimal action
$x_F^*$ satisfying
\[
\kappa_F(\varepsilon)
:=
\inf_{\substack{x\in\mathcal X:\\ |x-x_F^*|\geq\varepsilon}}
\left\{
\cvar_{\delta,\xi\sim F}[c(x,\xi)]
-
\opt_\delta(F)
\right\}
>0
\]
for every $\varepsilon>0$. If, for some $\varepsilon>0$ and $\eta>0$,

\[
\text{\(\displaystyle \liminf_{n\to\infty}
\mathbb{P}_F\!\left(
|x^{\pi}-x_F^*|\geq\varepsilon
\right)
\geq\eta,\)}
\]
then
\[
\text{\(\displaystyle \liminf_{n\to\infty}
\mathrm{PS}_{\delta,n}(\pi,F)
\geq
\eta\,\kappa_F(\varepsilon)>0,\)}
\]
and hence
\[
\text{\(\displaystyle \mathrm{PS}_{\delta,n}(\pi,F)=\Omega(1).\)}
\]
\end{proposition}
We prove the proposition in Appendix~\ref{prf:separation_inconsistency}.
The same bound yields the form used below: if $x_F^*$ is well separated and $\mathrm{PS}_{\delta,n}(\pi,F)\to 0$, then $x^{\pi}\to x_F^*$ in probability.

A policy enters this regime when it remains away from the oracle with probability bounded below. This occurs, for example, if it converges to a surrogate-objective optimum different from $x_F^*$, or to the best action in a restricted family whose optimum differs from $x_F^*$. Shrinking an otherwise consistent action toward a distinct fixed anchor with a weight converging to a positive constant also gives a displaced limit. Confining the policy to actions outside a neighborhood of $x_F^*$, while the oracle continues to minimize over $\mathcal{X}$, forces persistent displacement directly. These mechanisms satisfy the proposition when the oracle is well separated.

\subsection{A summary of the regimes}
Theorems~\ref{prop:asymptotic_characterization} and~\ref{cor:atom_category},
Proposition~\ref{prop:multiple_optima} and Proposition~\ref{prop:separation_inconsistency}
provide a spectrum of possibilities for
the relative importance of the two regret components. The spectrum covers a wide variety of cost functions and distributions. However, it is not a partition: other combinations of the policy, the cost, and $F$ are possible, and the decomposition is not characterized for every one of them. These regimes are
summarized in Table~\ref{tab:regret_categories}.

\begin{table}[ht]
\centering
\small
\begin{tabular}{@{}l p{0.46\textwidth} c c@{}}
\hline
\textbf{Regime} & \textbf{Entered when} &
\shortstack{\textbf{Sample-risk}\\\textbf{premium}} &
\shortstack{\textbf{Expected}\\\textbf{post-sample}\\\textbf{regret}} \\
\hline
Smooth &
Unique optimum; the cutoff moves to first order and the loss has a positive finite density there; the policy concentrates at rate $\rate\searrow 0$. &
$\Theta(\rate^{2})$ & $\Theta(\rate^{2})$ \\[6pt]
Atomic &
Unique optimum; an atom at the cutoff moves with the action; the policy concentrates at rate $\rate\searrow 0$. &
$\Theta(\rate)$ & $\Theta(\rate^{2})$ \\[6pt]
Mixing &
Several actions are optimal, and the policy keeps mixing over them, with no common minimizing threshold. &
$\Theta(1)$ & $o(1)$ \\[6pt]
Displaced &
Unique well-separated optimum, and the policy is persistently bounded away from it. &
unrestricted & $\Omega(1)$ \\
\hline
\end{tabular}
\caption{Structural regimes governing the relative magnitude of the
sample-risk premium and expected post-sample regret.
In the displaced row, unrestricted means that
Proposition~\ref{prop:separation_inconsistency} does not constrain the premium.
Each row abbreviates the hypotheses of the corresponding result. The smooth and atomic rows also use positive finite curvature, the moment and nondegeneracy conditions, and, in the atomic row, strict straddling and the isolation limits.
Concentration at a rate $\rate\searrow0$ implies consistency.
Persistent displacement is strictly stronger than
failure of consistency: an inconsistent policy need only sit away from $x_F^*$ along a
subsequence, which yields $\limsup_n\Pcomp{\pi}{F}>0$ rather than the
$\Pcomp{\pi}{F}=\Omega(1)$ recorded in the row.}
\label{tab:regret_categories}
\end{table}
Under the stated hypotheses of the smooth and atomic regimes, the sample-risk premium is of at least the same order as expected post-sample regret. 
Equal orders do not imply similar magnitudes. For centered limiting fluctuations in the smooth regime, the premium ratio tends to $\Kprice$, which reaches thousands in Table~\ref{tab:K_delta}. The atomic regime attains the $O(r_n)$ upper order from Proposition~\ref{prop:premium_ceiling} under the stated cutoff and moment conditions. Other orders remain possible outside the representative regimes.
Section~\ref{sec:saa} shows that sample average approximation is consistent at a unique well-separated oracle under a compact action set and a loss that is continuous in the action and uniformly bounded. The displaced regime is then avoided. In the atomic regime the sample-risk premium is of larger order than the expected post-sample regret.

\section{Regret of SAA} \label{sec:saa}

We now establish the concentration properties of sample average approximation (SAA) needed to apply Section~\ref{sec:decomposition}, and use them to determine the rates of its two regret components. Throughout this section, $F$ and $\delta\in(0,1)$ are fixed.

Sample average approximation replaces $F$ by the empirical distribution $\hat F_n$ and minimizes the resulting $\cvar$. For $\xi\sim\hat F_n$,
\begin{equation}\label{eqn:saa_objective}
    \cvar_{\delta,\xi \sim \hat{F}_n}\left[c\left(x, \xi\right)\right] ~~=~~ \inf_t  \left\{t + \frac{1}{1-\delta} \cdot \frac{1}{n} \sum_{i=1}^n (c(x, \xi_i) - t)^+ \right \}
\end{equation}
Sort the sample losses from worst to best as $c_{[1]}(x)\ge\cdots\ge c_{[n]}(x)$. The infimum over $t$ reduces to a weighted average of the upper tail, with $m=\lceil(1-\delta)n\rceil$:
\[ \cvar_{\delta,\xi \sim \hat{F}_n}\left[c\left(x, \xi\right)\right] ~~=~~ \frac{1}{1-\delta} \cdot \left[\frac{1}{n} \cdot \sum_{i=1}^{m-1} c_{[i]}\left(x\right) + \left(1 - \delta - \frac{m-1}{n}\right) \cdot c_{[m]}\left(x\right) \right] \ . \]
An SAA action is a measurable minimizer
\[
x^{\mathrm{SAA}}
\in
\arg\min_{x\in\mathcal{X}}
\cvar_{\delta,\xi\sim\hat F_n}[c(x,\xi)].
\]
We assume a minimizer exists. If several actions attain it, $x^{\mathrm{SAA}}$ is the one returned by a fixed measurable tie-break.

SAA minimizes an empirical estimate of the outcome-only objective $\Qc{F}{\cdot}$. The ex-ante value of the procedure is the $\cvar$ of losses recorded across independent deployments, each with its own training sample. That is the object in which the sample-risk premium appears, and it is not the objective in equation~\eqref{eqn:saa_objective}.

\subsection{Post-sample consistency and the sample-risk premium}

Under the conditions below, the expected outcome-only CVaR of the action selected by SAA converges to the oracle benchmark. Thus the post-sample component in equation~\eqref{eqn:regret_psi_decomp} vanishes. 

Consistency and convergence-in-mean results are well established in SAA analysis; see, for example, \citet{BanholzerFW22}. The proposition below records a bounded, continuous setting in which expected post-sample regret vanishes, providing the starting point for analyzing the two components of combined-risk regret.

\begin{proposition}
\label{prop:saa_post_sample_consistency}
Fix a distribution $F$. Suppose that:
\begin{enumerate}
    \item The action set $\mathcal{X}$ is nonempty and compact.
    \item The mapping $x \mapsto c(x,\xi)$ is continuous on $\mathcal{X}$
    for $F$-almost every $\xi$.
    \item The loss is uniformly bounded under $F$: there exists $M<\infty$
    such that
    \[
        \sup_{x\in\mathcal{X}} |c(x,\xi)|
        \leq M
        \qquad \text{almost surely}.
    \]
\end{enumerate}
Then
\[
    \mathrm{PS}_{\delta,n}(\pi^{\mathrm{SAA}},F)
    =
    \mathbb{E}_{\xi_1^n\sim F}\!\left[
        \Qc{F}{x^{\mathrm{SAA}}}
    \right]
    -
    \opt_\delta(F)
    \longrightarrow 0.
\]
\end{proposition}
The proof is in Appendix~\ref{apx:saa_post_sample_consistency}.
Proposition~\ref{prop:saa_post_sample_consistency} controls the post-sample component. With several optimal actions, persistent mixing without a common minimizing threshold can still retain a positive premium, as Proposition~\ref{prop:multiple_optima} shows. If the oracle is unique, continuity on the compact action set makes it well separated, so Proposition~\ref{prop:separation_inconsistency} gives $x^{\mathrm{SAA}}\to x_F^*$ in probability. Boundedness and almost-sure continuity then give convergence of the induced losses in mean; continuity of CVaR in the $L^1$ metric implies that combined-risk regret, and hence both components, vanish.

\subsection{\texorpdfstring{Concentration and regret rates}{Concentration and regret rates}}
We next establish the concentration rate of SAA and determine how quickly each regret component vanishes.
In addition to the structural conditions of the smooth or atomic regime, the rate uses almost-sure action differentiability at $x_F^*$ and a local Lipschitz envelope with a moment above two. Specifically, $\partial_x c(x_F^*,\xi)$ exists for $F$-almost every $\xi$, and there is a nonnegative $\mu(\xi)$ with $\mathbb{E}[\mu(\xi)^{2+\eta}]<\infty$ for some $\eta>0$ such that, for $F$-almost every $\xi$, simultaneously for all $x,y$ in a fixed neighborhood of $x_F^*$,
\begin{equation}\label{eqn:marginal_loss_bound}
    \bigl|c(x,\xi)-c(y,\xi)\bigr|
    ~~\leq~~
    \mu(\xi)\,|x-y|.
\end{equation}
These conditions control fluctuations of the empirical criterion near the oracle. For the squared loss $(x-\xi)^2$ on a compact action interval, the derivative exists and one may take $\mu(\xi)=C(1+|\xi|)$ for a suitable constant $C$. Bounded outcomes also ensure the uniform loss bound in Proposition~\ref{prop:saa_post_sample_consistency} and all required moments. The unbounded families in Table~\ref{tab:K_delta} provide values of $\Kprice$; applying the SAA result below additionally requires its bounded-loss hypothesis.

\begin{proposition}
\label{prop:saa_concentration_rate}
Suppose the hypotheses of Proposition~\ref{prop:saa_post_sample_consistency}, and let $x_F^*$ be the unique interior, well-separated oracle action. Assume the hypotheses on $Q_F$, the cutoff, and the loss distribution in either Theorem~\ref{prop:asymptotic_characterization} or Theorem~\ref{cor:atom_category}. Suppose also that the loss satisfies~\eqref{eqn:marginal_loss_bound} with $\mathbb{E}[\mu(\xi)^{2+\eta}]<\infty$ for some $\eta>0$, and that $\partial_x c(x_F^*,\xi)$ exists for $F$-almost every $\xi$.  

Write $q^*=q_F(x_F^*)$ and let $a_F=q_F'(x_F^*)$ in the smooth regime and $a_F=\ell'(x_F^*)$ in the atomic regime. Define
\begin{equation}\label{eqn:saa_variance}
\text{$\displaystyle \begin{aligned}
\psi_F(\xi)&=a_F+\frac{\partial_xc(x_F^*,\xi)-a_F}{1-\delta}
\mathbf1_{\{c(x_F^*,\xi)>q^*\}},\\
\sigma^2&=\frac{\operatorname{Var}(\psi_F(\xi))}{Q_F''(x_F^*)^2}.
\end{aligned}$}
\end{equation}
Then $0<\sigma^2<\infty$, and
\[
\text{$\displaystyle \sqrt n\,(x^{\mathrm{SAA}}-x_F^*)\ \Rightarrow\ \mathcal{N}(0,\sigma^2).$}
\]
Moreover, $\{n(x^{\mathrm{SAA}}-x_F^*)^2\}_{n\ge1}$ is uniformly integrable. In particular,
\[
x^{\mathrm{SAA}}-x_F^*
~~=~~
\Theta_{\mathbb P}\!\left(n^{-1/2}\right).
\]
\end{proposition}

The proof is in Appendix~\ref{prf:saa_concentration_rate}. The normal limit and uniform integrability establish the concentration and moment conditions in Section~\ref{sec:decomposition} with $\rate=n^{-1/2}$. Bounded loss supplies the remainder estimates needed in the proofs of Theorems~\ref{prop:asymptotic_characterization} and~\ref{cor:atom_category}, giving
\[
\text{$\displaystyle \begin{aligned}
\text{Smooth:}\quad &\mathrm{PS}=\Theta(n^{-1}),\qquad \mathrm{SR}=\Theta(n^{-1}),\\
\text{Atomic:}\quad &\mathrm{PS}=\Theta(n^{-1}),\qquad \mathrm{SR}=\Theta(n^{-1/2}).
\end{aligned}$}
\]
The action converges at the same rate in both regimes, but combined-risk regret decays more slowly in the atomic regime because of the premium.  In the smooth regime, the centered normal limit also gives $\rho_n\to\Kprice$ by Corollary~\ref{cor:rho_limit}.

The newsvendor loss is piecewise linear in the order with uniformly bounded slopes, so $\mu$ can be chosen as a deterministic constant. Continuous demand makes the action derivative exist almost surely at the oracle and yields the isolation limits~\citep{BesbesRiegerNewsvendor}. On a compact order set with bounded continuous demand, the proposition therefore applies when the structural atomic conditions of Theorem~\ref{cor:atom_category} hold.

\subsection{\texorpdfstring{{Finite-sample illustrations}}{Finite-sample illustrations}}\label{sec:saa_finite}
{We evaluate SAA at finite sample sizes in two smooth squared-loss examples and an atomic newsvendor example.}

\paragraph{\texorpdfstring{{Smooth losses.}}{Smooth losses.}}
{Consider $c(x,\xi)=(\xi-x)^2/2$ at $\delta=0.95$, with $\mathcal X=[0,B]$. The outcome is either exponential with rate $1$, conditioned on $\xi\le12$, or lognormal with $\log \xi\sim\mathcal N(0,1.8^2)$ before conditioning on $\xi\le2000$; in each case $B$ is the upper bound. These bounded examples satisfy Proposition~\ref{prop:saa_concentration_rate}, as verified in Appendix~\ref{apx:saa_smooth_finite}. Table~\ref{tab:saa_smooth_finite} reports estimates from $12{,}000$ independent training samples for each row. Conditional risk evaluations use analytic truncated moments of the specified outcome distribution.}

\begin{table}[ht]
\centering
\small
{\begin{tabular}{lrrrr}
\hline
Outcome & $K$ & $n$ & $\mathrm{SR}/\mathrm{PS}$ & Premium share (\%) \\
\hline
Exponential & $8.76$ & $100$ & $2.90\ (0.05)$ & $74.4\ (0.31)$ \\
& & $400$ & $3.88\ (0.04)$ & $79.5\ (0.16)$ \\
& & $1600$ & $5.59\ (0.04)$ & $84.8\ (0.09)$ \\
\hline
Lognormal & $418.42$ & $100$ & $14.08\ (0.20)$ & $93.4\ (0.09)$ \\
& & $400$ & $12.52\ (0.14)$ & $92.6\ (0.08)$ \\
& & $1600$ & $14.38\ (0.15)$ & $93.5\ (0.06)$ \\
\hline
\end{tabular}}
\caption{{Smooth-regime SAA with the truncated outcome distributions specified in the text. Premium share is $\mathrm{SR}/(\mathrm{PS}+\mathrm{SR})$. Estimates use $12{,}000$ independent training samples per row. Parentheses give approximate Monte Carlo standard errors, in percentage points for shares. The limiting coefficient $K$ is computed from the population distribution. Appendix~\ref{apx:saa_smooth_finite} describes the computation.}}
\label{tab:saa_smooth_finite}
\end{table}

{The estimated premium share ranges from $74.4\%$ to $93.5\%$ across these cases. Yet the finite-sample ratios remain below $K$, especially for the lognormal outcome: at $n=1600$, the estimated ratio is $14.38$, compared with a limit of $418.42$. Thus the premium can already account for most regret while its ratio to post-sample regret is far from its local asymptotic value.}

\paragraph{\texorpdfstring{{An atom at the cutoff.}}{An atom at the cutoff.}}

{Consider the newsvendor loss $c(x,\xi)=x/2-\min(x,\xi)$, with $\xi\sim U[0,1]$, $\mathcal X=[0,1]$, and $\delta=0.9$. The unique oracle order is $x_F^*=0.05$, and the problem satisfies the atomic hypotheses.  With the smallest-minimizer tie-break, SAA returns an order with exact sampling distribution}
\begin{equation}\label{eqn:saa_beta}
\text{{$\displaystyle x^{\mathrm{SAA}}=\xi_{(k)},\qquad
k=\lceil n/20\rceil,\qquad
\xi_{(k)}\sim\operatorname{Beta}(k,n+1-k).$}}
\end{equation}
{Table~\ref{tab:saa_finite} evaluates both regret components using this distribution. The premium accounts for about $76\%$ of regret at $n=110$ and $95\%$ at $n=1630$. The scaled components approach the constants predicted by Propositions~\ref{prop:saa_concentration_rate} and~\ref{prop:atom_brackets}. Appendix~\ref{apx:saa_finite} gives the formulas and verifies the assumptions.}

\begin{table}[ht]
\centering
\small
{\begin{tabular}{rrrrr}
\hline
$n$ & $n\,\mathrm{PS}$ & $\sqrt n\,\mathrm{SR}$ & $\mathrm{SR}/\mathrm{PS}$ & Premium share \\
\hline
$110$ & $0.2548$ & $0.0781$ & $3.21$ & $76.3\%$ \\
$430$ & $0.2434$ & $0.0962$ & $8.19$ & $89.1\%$ \\
$1630$ & $0.2391$ & $0.1041$ & $17.58$ & $94.6\%$ \\
$6430$ & $0.2379$ & $0.1078$ & $36.34$ & $97.3\%$ \\
Limit & $0.2375$ & $0.1112$ & $\sim0.4683\sqrt n$ & $\to100\%$ \\
\hline
\end{tabular}}
\caption{{Finite-sample SAA regret for uniform demand. Premium share is $\mathrm{SR}/(\mathrm{PS}+\mathrm{SR})$. Values are evaluated numerically from the exact Beta distribution using incomplete Beta moments and a scalar quantile equation, without simulation.}}
\label{tab:saa_finite}
\end{table}

\section{Conclusion}
{This paper studies an ex-ante notion of tail risk for data-driven decisions. We evaluate CVaR over the combined randomness of training, policy randomization, and the future outcome, relative to an oracle that knows the distribution. The resulting combined-risk regret decomposes into two nonnegative terms: average post-sample regret and a sample-risk premium. The first averages the outcome-only CVaR gaps of the selected actions; it already captures how learning affects action quality. The second measures the additional tail risk omitted by this averaging. This premium can be positive even when every selected action is individually optimal under outcome-only CVaR.}

{The regime analysis shows how the premium can become the main component of regret. Under the smooth-regime hypotheses, the two components have the same asymptotic order. For centered limiting fluctuations, the ratio of the premium to post-sample regret tends to $\Kprice$, which can be large. The curvature comparison explains this difference: using one tail threshold across samples adds a variance penalty. The limit alone does not determine the ratio at a given sample size. Under the atomic-regime hypotheses, the premium decays more slowly than average post-sample regret and becomes the leading component. With persistent mixing among optimal actions that admit no common minimizing threshold, the premium can remain positive even as average post-sample regret vanishes.}

{Sample average approximation illustrates the distinction between action convergence and regret. Under Proposition~\ref{prop:saa_concentration_rate}, its action converges at rate $n^{-1/2}$ in both the smooth and atomic regimes, but its combined-risk regret is of order $n^{-1}$ in the first and $n^{-1/2}$ in the second. The finite-sample examples in Tables~\ref{tab:saa_smooth_finite} and~\ref{tab:saa_finite} illustrate the importance of the premium in both regimes: it accounts for approximately $74\%$--$97\%$ of regret in the reported cases. In the smooth examples, these large premium shares occur while the ratios remain below their limiting coefficients.}

{These regimes are representative settings at a fixed distribution, not an exhaustive classification. The smooth expansion also identifies a local design tradeoff: at leading order, action variance receives weight $1+\Kprice$ relative to squared bias. Comparing policies through this expansion requires both to satisfy its hypotheses. Finite-sample guarantees for policy changes motivated by this tradeoff, and statistical guarantees for estimating $\Kprice$ and the cutoff-based upper bound, remain questions for further work. Evaluating a learned policy through its average outcome-only risk can omit a leading component of excess tail risk.}

\bibliographystyle{informs2014}
\bibliography{sample}

\appendix
\section{Decomposition of regret}\label{apx:regret_decomposition}

\subsection{Proof of Proposition~\ref{prop:endpoints}}
\label{prf:endpoints}

For $x\in\mathcal X$ and $t\in\mathbb R$, write
$\phi_x(t):=\phiv{F}{x}{t}$, so that
\[
    \Qc{F}{x}=\min_{t\in\mathbb R}\phi_x(t),
    \qquad
    \mathcal T_\delta(x)=\arg\min_{t\in\mathbb R}\phi_x(t).
\]
Both the minimum and the minimizing set are nonempty for almost every realization of
$x^\pi$, by integrability.

\paragraph{Part 1.}
By the independence of $x^\pi$ and $\xi$ and equation~\eqref{eqn:sr_variational},
\[
    \SRcomp{\pi}{F}
    =
    \inf_{t\in\mathbb R}G(t),
    \qquad
    G(t)
    :=
    \mathbb E_{x^\pi}\!\left[
        \phi_{x^\pi}(t)-\min_{s\in\mathbb R}\phi_{x^\pi}(s)
    \right],
\]
and the integrand defining $G(t)$ is nonnegative for every $t$. The
quantity $\mathbb E_{x^\pi}[\min_s\phi_{x^\pi}(s)]=\mathbb E_{x^\pi}[\Qc{F}{x^\pi}]$ is finite by
integrability and does not depend on $t$, so $G$ differs from
$t\mapsto\mathbb E_{x^\pi}[\phi_{x^\pi}(t)]$ by a constant. Consequently $G$ attains its
infimum at any $t^*$ attaining the variational representation of
$\cvar_\delta[c(x^\pi,\xi)]$, and such a $t^*$ exists by integrability.

Suppose first that the tails are aligned, so that some $t^\star$ satisfies
$t^\star\in\mathcal T_\delta(x^\pi)$ almost surely. Then
$\phi_{x^\pi}(t^\star)=\min_s\phi_{x^\pi}(s)$ almost surely, hence $G(t^\star)=0$ and the
premium vanishes. Conversely, suppose the premium vanishes, so that
$G(t^*)=0$. The integrand is nonnegative, whence
$\phi_{x^\pi}(t^*)=\min_s\phi_{x^\pi}(s)$ almost surely, that is,
$t^*\in\mathcal T_\delta(x^\pi)$ almost surely, and the tails are aligned.
Finally, if $x^\pi=x_0$ almost surely, then any element of
$\mathcal T_\delta(x_0)$, a nonempty set, serves as a common threshold; the
tails are aligned and the premium vanishes.

\paragraph{Part 2.}
By definition,
\[
    \Pcomp{\pi}{F}
    =
    \mathbb E_{x^\pi}\!\left[\Qc{F}{x^\pi}\right]-\opt_\delta(F)
    =
    \mathbb E_{x^\pi}\!\left[\Qc{F}{x^\pi}-\opt_\delta(F)\right],
\]
and $\Qc{F}{x}\geq\opt_\delta(F)$ for every $x\in\mathcal X$, so the
integrand is nonnegative. Hence $\Pcomp{\pi}{F}=0$ if and only if
$\Qc{F}{x^\pi}=\opt_\delta(F)$ almost surely. Since the infimum defining
$\opt_\delta(F)$ is attained,
$\{x\in\mathcal X:\Qc{F}{x}=\opt_\delta(F)\}=\mathcal X^*_F$, so this is
exactly $x^\pi\in\mathcal X^*_F$ almost surely. If the oracle action is unique,
$\mathcal X^*_F=\{\xopt{F}\}$ and the condition reads $x^\pi=\xopt{F}$ almost
surely.

\subsection{\texorpdfstring{{Proof of Proposition~\ref{prop:premium_ceiling}}}{Proof of the premium bound}}\label{prf:premium_ceiling}

Write $Y=q_F(x^\pi)$. Markov's inequality gives
\[
|q_F(x)|\le\frac{\mathbb E_\xi|c(x,\xi)|}{\min\{\delta,1-\delta\}},
\]
so $Y$ is integrable. The one-sided threshold derivatives of $\phi_F(x,t)$ lie in $[-\delta/(1-\delta),1]$. Since $q_F(x)$ minimizes this convex function,
\[
0\le\phi_F(x,t)-Q_F(x)
\le(t-q_F(x))^++\frac{\delta}{1-\delta}(q_F(x)-t)^+.
\]
Taking expectations and infima in $t$, and using the CVaR variational formula, gives
\[
\begin{aligned}
0\le\mathrm{SR}&\le\inf_t\mathbb E\left[(t-Y)^++\frac{\delta}{1-\delta}(Y-t)^+\right]\\
&=\inf_t\left\{t+\frac{\mathbb E[(Y-t)^+]}{1-\delta}\right\}-\mathbb E Y
=\cvar_\delta[Y]-\mathbb E Y.
\end{aligned}
\]
For a conditionally deterministic loss, $Q_F(x)=q_F(x)=c(x,\xi)$ almost surely, so $\mathrm{SR}=\cvar_\delta[Y]-\mathbb E Y$. Since $L_\delta=\max\{1,\delta/(1-\delta)\}$,
\[
(t-Y)^++\frac{\delta}{1-\delta}(Y-t)^+\le L_\delta|Y-t|.
\]
Taking the expectation and the infimum in $t$ gives the last inequality in equation~\eqref{eqn:premium_ceiling}.

\subsection{\texorpdfstring{{A linear rate from the premium bound}}{A linear rate from the premium bound}}
\label{prf:premium_rate}

\begin{corollary}[Linear rate for the sample-risk premium]\label{cor:premium_rate}
Maintain the hypotheses of Proposition~\ref{prop:premium_ceiling} and the Lipschitz condition~\eqref{eqn:lipschitz}.
Let $x^*$ be such that $q_F$ is differentiable at $x^*$, and suppose
$x^\pi=x^*+r_n Z_n$ with $r_n\to 0$ and $\sup_n\mathbb E|Z_n|<\infty$.
Then $\mathrm{SR}=O(r_n)$.
\end{corollary}

\begin{proof}
Differentiability at $x^*$ gives $|q_F(x)-q_F(x^*)|\le C|x-x^*|$ in a fixed neighborhood.
The Lipschitz envelope in~\eqref{eqn:lipschitz} and Markov's inequality in the proof of Proposition~\ref{prop:premium_ceiling} give $|q_F(x)-q_F(x^*)|\le C'(1+|x-x^*|)$ globally.
Outside that neighborhood the same comparison is bounded by a constant times $|x-x^*|$.
Evaluating~\eqref{eqn:premium_ceiling} at $t=q_F(x^*)$ therefore gives
\[
\mathrm{SR}\le C''\mathbb E|x^\pi-x^*|
=C''r_n\mathbb E|Z_n|=O(r_n).
\]
\end{proof}

\subsection{Proof of Theorem~\ref{prop:asymptotic_characterization}}
\label{prf:asymptotic_characterization}

Both $\Pcomp{\pi}{F}$ and $\SRcomp{\pi}{F}$ are expectations, and the hypotheses control them near $x^*$. The argument records that local control and the orders it induces, and then checks that the contribution of the event on which the induced action leaves a neighborhood of $x^*$ is $o(\rate^{2})$.

Twice differentiability of $Q_F$ at the interior minimizer $x^*$, with $0<Q_F''(x^*)<\infty$, makes $\Qc{F}{x}-\Qc{F}{x^*}$ quadratic in $|x-x^*|$. The assumption $q_F'(x^*)\neq 0$ makes the cutoff move linearly with the action. A positive finite density of the loss, continuous at $(x^*,q^*)$, makes the variational objective from equation~\eqref{eqn:variational_objective} quadratic in the threshold. The following lemma turns those facts into bounds on a neighborhood of $x^*$: the outcome-only $\cvar$ grows like $|x-x^*|^{2}$, the cutoff moves linearly with the action, and the variational objective grows like the square of the distance from the cutoff. 

\begin{lemma}[Smooth exponents]\label{lem:smooth_exponents}
Suppose that the hypotheses of Theorem~\ref{prop:asymptotic_characterization} hold. Write $q^*:=\qv{F}{x^*}$ and $g:=g_{x^*}(q^*)$. For every $\theta>0$ there is a neighborhood $\mathcal{N}:=\left[x^*-\eta,\,x^*+\eta\right]$ of $x^*$, together with constants $0<c_2\leq C_2<\infty$ and $0<c_\gamma\leq C_\gamma<\infty$, a threshold radius $u_0>0$, and $\varepsilon\geq0$, such that for every $x\in\mathcal{N}$,
\begin{align}
    c_2\left|x-x^*\right|^{2} ~~\leq~~ \Qc{F}{x}-\Qc{F}{x^*} ~~&\leq~~ C_2\left|x-x^*\right|^{2} \ , \label{eqn:loc_P}\\
    \left|\,\qv{F}{x}-q^*-q_F'(x^*)\left(x-x^*\right)\,\right| ~~&\leq~~ \varepsilon\left|x-x^*\right| \ , \label{eqn:loc_q}
\end{align}
and such that for every $x\in\mathcal{N}$ and every $\left|u\right|\leq u_0$,
\begin{equation}\label{eqn:loc_phi}
    c_\gamma\left|u\right|^{\gamma} ~~\leq~~ \phiv{F}{x}{\qv{F}{x}+u}-\Qc{F}{x} ~~\leq~~ C_\gamma\left|u\right|^{\gamma} \ ,
\end{equation}
with $\gamma=2$ and $q_F'(x^*)\neq 0$, with $c_2$ and $C_2$ within $\theta$ of $\tfrac12 Q_F''(x^*)$, with $c_\gamma$ and $C_\gamma$ within $\theta$ of $g/(2(1-\delta))$, and with $\varepsilon$ arbitrarily small.
\end{lemma}
The proof is in Appendix~\ref{prf:smooth_brackets}.
The comparisons bound a fixed action. Under the concentration in equation~\eqref{eqn:concentration}, one has $x^\pi-x^*=\rate Z_n$, so the post-sample regret and the premium, both expectations of these comparisons at the induced action, are of order $\rate^{2}$ and $\rate^{\gamma}$ once the contribution outside $\mathcal{N}$ is of smaller order and the moments of $Z_n$ stay of order one. The passage from the local comparisons to these orders is the same at either exponent, so it is recorded once and cited by Theorem~\ref{cor:atom_category} at $\gamma=1$.

\begin{lemma}[Orders under local brackets]\label{lem:orders}
Suppose $q_F'(x^*)\neq 0$ and $\gamma\geq 1$. Suppose that on every sufficiently small neighborhood $\mathcal{N}:=\left[x^*-\eta,\,x^*+\eta\right]$ the comparisons in equations~\eqref{eqn:loc_P}--\eqref{eqn:loc_phi} hold, with the cutoff error $\varepsilon$ in~\eqref{eqn:loc_q} arbitrarily small. The nonzero slope keeps a flat cutoff from making the premium vanish while the post-sample regret stays quadratic. Write $A:=\left\{x^\pi\in\mathcal{N}\right\}$ and
\[
    R_{\mathrm{PS}} ~:=~ \mathbb{E}\!\left[\left(\Qc{F}{x^\pi}-\Qc{F}{x^*}\right)\mathbf{1}_{A^{c}}\right] \ ,
    \qquad
    R_{\mathrm{SR}} ~:=~ \sup_{\left|t-q^*\right|\leq u_0/2}\mathbb{E}\!\left[\left(\phiv{F}{x^\pi}{t}-\Qc{F}{x^\pi}\right)\mathbf{1}_{A^{c}}\right] \ ,
\]
where $u_0$ is the threshold radius in~\eqref{eqn:loc_phi}. Both remainders are nonnegative. Let the policy concentrate as
in equation~\eqref{eqn:concentration}, with $\bigl\{|Z_n|^{\max\{2,\,\gamma\}}\bigr\}$
uniformly integrable and $Z$ nondegenerate, and suppose
$R_{\mathrm{PS}}=o(\rate^{2})$ and $R_{\mathrm{SR}}=o(\rate^{\gamma})$. Then
\begin{equation}\label{eqn:asym_orders}
    \Pcomp{\pi}{F} ~~=~~ \Theta\!\left(\rate^{2}\right) \ ,
    \qquad
    \SRcomp{\pi}{F} ~~=~~ \Theta\!\left(\rate^{\gamma}\right) \ .
\end{equation}
\end{lemma}
The proof is in Appendix~\ref{prf:orders}.
Lemma~\ref{lem:smooth_exponents} gives the comparisons at $\gamma=2$. What remains is the remainder hypothesis of Lemma~\ref{lem:orders}. The two remainders are supported on $\{x^\pi\notin\mathcal{N}\}$, and the following bounds control them there.

The remainder $R_{\mathrm{SR}}$ admits a bound in terms of the induced $\VaR$ outside the neighborhood. The variational objective is convex in the threshold. Its right derivative is
\begin{equation}\label{eqn:phi_derivative}
    \partial_t^{+}\phiv{F}{x}{t} ~~=~~ 1-\frac{\mathbb{P}\left(c\left(x,\xi\right)>t\right)}{1-\delta} \ ,
\end{equation}
and the left derivative replaces $c\left(x,\xi\right)>t$ by $c\left(x,\xi\right)\ge t$. At an atom the two differ. Both are bounded by
\[
    \left|\partial_t^{\pm}\phiv{F}{x}{t}\right| ~~\leq~~ L_\delta:=\max\left\{1,\frac{\delta}{1-\delta}\right\} \ ,
\]
uniformly in $x$ and $t$, and convexity gives the same Lipschitz bound from either one. Since $\qv{F}{x}$ minimizes $t\mapsto\phiv{F}{x}{t}$, this yields
\[
    R_{\mathrm{SR}} ~~\leq~~ L_\delta\left(\frac{u_0}{2}\,\mathbb{P}\left(A^{c}\right)+\mathbb{E}\!\left[\left|\qv{F}{x^\pi}-q^*\right|\mathbf{1}_{A^{c}}\right]\right) \ .
\]
The Lipschitz constant $L_\delta$ grows like $1/(1-\delta)$. Equation~\eqref{eqn:lipschitz} gives the companion bound for the post-sample remainder. Since $|\cvar_\delta[Y]-\cvar_\delta[Y']|\leq(1-\delta)^{-1}\mathbb{E}|Y-Y'|$,
\[
    \bigl|\Qc{F}{x}-\Qc{F}{y}\bigr|
    ~~\leq~~
    \frac{\mathbb{E}[L(\xi)]}{1-\delta}\,|x-y|,
\]
and therefore $R_{\mathrm{PS}}\leq \mathbb{E}[L(\xi)]\,(1-\delta)^{-1}\,\mathbb{E}\bigl[|x^\pi-x^*|\mathbf{1}_{A^{c}}\bigr]$. Equation~\eqref{eqn:lipschitz} likewise bounds the cutoff: $|\qv{F}{x}-q^*|\leq C\bigl(1+|x-x^*|\bigr)$ for a finite constant $C$, because $c(x,\xi)\leq |c(x^*,\xi)|+L(\xi)|x-x^*|$.

Under the concentration in equation~\eqref{eqn:concentration}, these bounds have to be turned into $o(\rate^{2})$. Write $x^*:=\xopt{F}$, $q^*:=\qv{F}{x^*}$ and $\Delta:=x^\pi-x^*=\rate Z_n$, and for a radius
$\eta>0$ let $\mathcal{N}:=[x^*-\eta,x^*+\eta]$ and $A_n:=\{x^\pi\in\mathcal{N}\}$. Since
$\rate\searrow0$ and $Z_n\Rightarrow Z$, we have $\mathbb{P}(A_n)\to1$ for every fixed $\eta$.

\paragraph{Escapes.}
The bounds above and the Lipschitz assumption give, for every fixed radius $\eta>0$,
\[
    R_{\mathrm{PS}}
    ~~\leq~~
    \frac{\mathbb{E}[L(\xi)]}{1-\delta}\,\mathbb{E}\bigl[|\Delta|\mathbf{1}_{A_n^c}\bigr]
\]
and
\[
    R_{\mathrm{SR}}
    ~~\leq~~
    C\Bigl(\mathbb{P}(A_n^c)+\mathbb{E}\bigl[|\Delta|\mathbf{1}_{A_n^c}\bigr]\Bigr).
\]
Uniform integrability of $\bigl\{|Z_n|^2\bigr\}$ forces $\mathbb{E}\bigl[Z_n^2\mathbf{1}_{|Z_n|>\eta/\rate}\bigr]\to 0$. Markov's inequality gives $\mathbb{P}(A_n^c)\leq (\rate/\eta)^2$ times that expectation, hence $\mathbb{P}(A_n^c)=o(\rate^2)$. On the same event $|Z_n|\leq (\rate/\eta)Z_n^2$, so $\mathbb{E}[|\Delta|\mathbf{1}_{A_n^c}]=\rate\,\mathbb{E}[|Z_n|\mathbf{1}_{A_n^c}]\leq (\rate^2/\eta)\,\mathbb{E}[Z_n^2\mathbf{1}_{A_n^c}]=o(\rate^2)$. Therefore, for every fixed neighborhood,
\begin{equation}\label{eqn:asym_localization}
    R_{\mathrm{PS}} ~~=~~ o\!\left(\rate^{2}\right) \ ,
    \qquad
    R_{\mathrm{SR}} ~~=~~ o\!\left(\rate^{2}\right) \ .
\end{equation}
Equation~\eqref{eqn:asym_localization} is then the remainder hypothesis of Lemma~\ref{lem:orders} at $\gamma=2$, so $\Pcomp{\pi}{F}=\Theta(\rate^{2})$ and $\SRcomp{\pi}{F}=\Theta(\rate^{2})$.

\subsection{Proof of Lemma~\ref{lem:orders}}\label{prf:orders}

The comparisons in equations~\eqref{eqn:loc_P}--\eqref{eqn:loc_phi} bound a fixed action. The post-sample regret is the expectation of $\Qc{F}{x^\pi}-\Qc{F}{x^*}$, so on the event $A:=\{x^\pi\in\mathcal{N}\}$ the first comparison bounds it by a second moment of $\Delta:=x^\pi-x^*$, and the contribution of $A^c$ is a remainder. The premium is the price of one shared threshold against the action-specific cutoffs $\qv{F}{x^\pi}$, so on $A$ it is the distance from $\qv{F}{x^\pi}$ to a single constant. For $\gamma\geq 1$, a random variable $W$ and an event $A$, let
\begin{equation}\label{eqn:dispersion}
    \dispA{\gamma}{W}{A} ~~:=~~ \inf_{t\in\mathbb{R}}\mathbb{E}\!\left[\left|W-t\right|^{\gamma}\mathbf{1}_{A}\right] \ ,
\end{equation}
and write $\disp{\gamma}{W}$ when $A$ is the sure event. This is the best approximation of $W$ by a single constant, measured in $\gamma$-th absolute moment. For $\gamma=2$ it is the variance, attained at the mean; for $\gamma=1$ it is the mean absolute deviation about the median. The functional is invariant under deterministic shifts,
\[
    \dispA{\gamma}{W+b}{A} ~~=~~ \dispA{\gamma}{W}{A} \qquad \text{for every constant } b \ ,
\]
and it vanishes precisely when $W$ is almost surely constant on $A$.

The expansion of the cutoff carries the distance for $\qv{F}{x^\pi}$ over to the same distance for $\Delta$. The resulting bounds hold at a single sample size, and the orders are obtained by scaling them once the remainders are of smaller order. The next lemma records both bounds and this transfer. The exponent $\gamma\geq 1$ is free: the smooth regime takes $\gamma=2$, and the atomic regime takes $\gamma=1$.

\begin{lemma}[Local brackets at a fixed sample size]\label{lem:fixed_n}
Suppose that on a neighborhood $\mathcal{N}:=\left[x^*-\eta,\,x^*+\eta\right]$ the comparisons~\eqref{eqn:loc_P}--\eqref{eqn:loc_phi} hold for an exponent $\gamma\geq 1$, with $q_F'(x^*)\neq 0$, and constants $0<c_2\leq C_2<\infty$ and $0<c_\gamma\leq C_\gamma<\infty$. Suppose also that the neighborhood is small enough that the induced cutoffs stay inside the threshold radius on which~\eqref{eqn:loc_phi} applies,
\begin{equation}\label{eqn:loc_radius}
    \left(\left|q_F'(x^*)\right|+\varepsilon\right)\eta ~~\leq~~ \frac{u_0}{4}\left(\frac{c_\gamma}{2C_\gamma}\right)^{1/\gamma} \ .
\end{equation}
Write $\Delta:=x^\pi-x^*$ and $A:=\left\{x^\pi\in\mathcal{N}\right\}$. Then
\begin{equation}\label{eqn:bracket_P}
    c_2\,\mathbb{E}\!\left[\left|\Delta\right|^{2}\mathbf{1}_{A}\right]
    ~~\leq~~
    \Pcomp{\pi}{F}
    ~~\leq~~
    C_2\,\mathbb{E}\!\left[\left|\Delta\right|^{2}\mathbf{1}_{A}\right] ~+~ R_{\mathrm{PS}} \ ,
\end{equation}
and
\begin{equation}\label{eqn:bracket_SR}
    c_\gamma\,\dispA{\gamma}{\qv{F}{x^\pi}}{A}
    ~~\leq~~
    \SRcomp{\pi}{F}
    ~~\leq~~
    C_\gamma\,\dispA{\gamma}{\qv{F}{x^\pi}}{A} ~+~ R_{\mathrm{SR}} \ ,
\end{equation}
where $R_{\mathrm{PS}}$ and $R_{\mathrm{SR}}$ are the remainders in the statement of Lemma~\ref{lem:orders}; both are nonnegative. Moreover, the dispersion of the induced $\VaR$ is itself controlled by that of $\Delta$,
\begin{equation}\label{eqn:bracket_q}
    \left|\;\dispA{\gamma}{\qv{F}{x^\pi}}{A}^{1/\gamma} ~-~ \left|q_F'(x^*)\right|\cdot\dispA{\gamma}{\Delta}{A}^{1/\gamma}\;\right|
    ~~\leq~~
    \varepsilon\left(\mathbb{E}\!\left[\left|\Delta\right|^{\gamma}\mathbf{1}_{A}\right]\right)^{1/\gamma} \ .
\end{equation}
\end{lemma}
The proof is in Appendix~\ref{prf:fixed_n}.
Write $x^*:=\xopt{F}$, $q^*:=\qv{F}{x^*}$ and $\Delta:=x^\pi-x^*=\rate Z_n$, and for a radius
$\eta>0$ let $\mathcal{N}:=[x^*-\eta,x^*+\eta]$ and $A_n:=\{x^\pi\in\mathcal{N}\}$. Since
$\rate\searrow0$ and $Z_n\Rightarrow Z$, we have $\mathbb{P}(A_n)\to1$ for every fixed $\eta$.

The comparisons in equations~\eqref{eqn:loc_P}--\eqref{eqn:loc_phi} are the hypothesis of Lemma~\ref{lem:fixed_n}, so the
brackets in equations~\eqref{eqn:bracket_P}, \eqref{eqn:bracket_SR} and~\eqref{eqn:bracket_q}
are available at every sample size, and all that remains is to evaluate them at
$\Delta=\rate Z_n$ and pass to the limit. The evaluation is stated at the free exponents
of those comparisons. In the smooth regime $\gamma=2$, and in the atomic regime $\gamma=1$. For these exponents equation~\eqref{eqn:asym_localization} supplies
$R_{\mathrm{PS}}=o(\rate^{2})$ and $R_{\mathrm{SR}}=o(\rate^{\gamma})$, since $o(\rate^{2})$ is $o(\rate^{\gamma})$ when $\gamma\le 2$. For $\gamma>2$ the argument uses the hypothesis $R_{\mathrm{SR}}=o(\rate^{\gamma})$ of the lemma. The second-moment escape does not give that stronger remainder.

\paragraph{An elementary observation.}
The one estimate the fixed-$n$ lemma does not supply is that a dispersion functional cannot
degenerate along a weakly convergent sequence. Fix $\gamma\geq1$ and let $Y_n\Rightarrow Y$ with
$\{|Y_n|^\gamma\}$ uniformly integrable and $Y$ nondegenerate. Then
\begin{equation}\label{eq:distance_to_constants}
    0<
    \liminf_{n\to\infty}
    \disp{\gamma}{Y_n}
    \leq
    \limsup_{n\to\infty}
    \disp{\gamma}{Y_n}
    <\infty.
\end{equation}
Indeed, the upper bound follows by taking $t=0$ in equation~\eqref{eqn:dispersion}, since uniform
integrability gives $\Lambda:=\sup_n\mathbb{E}[|Y_n|^\gamma]<\infty$. If the lower bound
failed, there would exist $t_n$ with
$\mathbb{E}[|Y_n-t_n|^\gamma]\to0$. Finiteness of $\Lambda$ makes $\{t_n\}$
bounded, so along a convergent subsequence with $t_n\to t$ we would have
$Y_n-t_n\to0$ in probability, and hence $Y=t$ almost surely, contradicting
nondegeneracy.

Restricting the functional to a high-probability event cannot destroy this, and the resulting
bound is uniform in the event, which is what lets a radius be chosen below without
circularity. Let $\{E_n\}$ be any events with $\mathbb{P}(E_n)\to1$. Then
\begin{equation}\label{eq:restricted_dispersion}
    \liminf_{n\to\infty}\dispA{\gamma}{Y_n}{E_n}
    ~~\geq~~
    \liminf_{n\to\infty}\disp{\gamma}{Y_n} \ .
\end{equation}
To see this, choose $M$ with
$\sup_n\mathbb{P}(|Y_n|>M)\leq1/4$, which is possible because $\{Y_n\}$ is bounded in
probability, and set $S:=M+(4\Lambda)^{1/\gamma}$. For $n$ large enough that
$\mathbb{P}(E_n)\geq3/4$ we have $\mathbb{P}(E_n\cap\{|Y_n|\leq M\})\geq1/2$, so for $|t|>S$
\[
    \mathbb{E}\!\left[\left|Y_n-t\right|^{\gamma}\mathbf{1}_{E_n}\right]
    ~\geq~ \tfrac{1}{2}\left(S-M\right)^{\gamma}
    ~=~ 2\Lambda
    ~>~ \dispA{\gamma}{Y_n}{E_n} \ ,
\]
the last inequality by the feasible choice $t=0$. The infimum
in equation~\eqref{eqn:dispersion} may therefore be taken over $|t|\leq S$, with $S$ free of $n$ and of
the event. For such $t$,
\[
    \mathbb{E}\!\left[\left|Y_n-t\right|^{\gamma}\mathbf{1}_{E_n^{c}}\right]
    ~\leq~ 2^{\gamma-1}\left(\mathbb{E}\!\left[\left|Y_n\right|^{\gamma}\mathbf{1}_{E_n^{c}}\right]+S^{\gamma}\mathbb{P}\!\left(E_n^{c}\right)\right)
    ~=:~ \varepsilon_n ~\longrightarrow~ 0 \ ,
\]
uniformly in $t$, by uniform integrability and $\mathbb{P}(E_n^c)\to0$. Hence
$\dispA{\gamma}{Y_n}{E_n}\geq\disp{\gamma}{Y_n}-\varepsilon_n$,
which gives equation~\eqref{eq:restricted_dispersion}.

\paragraph{Scaling, and the choice of radius.}
The functional $\dispA{\gamma}{\cdot}{A}$ is positively homogeneous of degree $\gamma$:
replacing $W$ by $cW$ with $c>0$ reindexes the infimum in equation~\eqref{eqn:dispersion} by
$t\mapsto ct$ and rescales the objective by $c^{\gamma}$. Since $\Delta=\rate Z_n$,
\begin{equation}\label{eq:asym_scaling}
    \mathbb{E}\!\left[\left|\Delta\right|^{2}\mathbf{1}_{A_n}\right]
    ~=~ \rate^{2}\,\mathbb{E}\!\left[\left|Z_n\right|^{2}\mathbf{1}_{A_n}\right] \ ,
    \qquad
    \dispA{\gamma}{\Delta}{A_n} ~=~ \rate^{\gamma}\dispA{\gamma}{Z_n}{A_n} \ .
\end{equation}
Uniform integrability of $\{|Z_n|^{\max\{2,\gamma\}}\}$, weak convergence
$Z_n\Rightarrow Z$ and $\mathbb{P}(A_n)\to1$ give
\begin{equation}\label{eq:asym_moments}
    \mathbb{E}\!\left[\left|Z_n\right|^{2}\mathbf{1}_{A_n}\right]
    \longrightarrow\mathbb{E}\!\left[\left|Z\right|^{2}\right] \ ,
    \qquad
    \mathbb{E}\!\left[\left|Z_n\right|^{\gamma}\mathbf{1}_{A_n}\right]
    \longrightarrow\mathbb{E}\!\left[\left|Z\right|^{\gamma}\right] \ ,
\end{equation}
both limits finite, and the first strictly positive because $Z$ is nondegenerate.
Uniform integrability of $\{|Z_n|^{\max\{2,\gamma\}}\}$ implies that of
$\{|Z_n|^{p}\}$ for every $p\leq\max\{2,\gamma\}$, since the family is bounded in
$L^{\max\{2,\gamma\}}$; in particular, the same argument gives
\begin{equation}\label{eq:asym_first_moment}
    \mathbb{E}\!\left[Z_n\mathbf{1}_{A_n}\right] ~\longrightarrow~ \mathbb{E}\!\left[Z\right] \ ,
\end{equation}
which is used only in the proof of Corollary~\ref{cor:rho_limit}. Applying equation~\eqref{eq:distance_to_constants} to $Y_n=Z_n$, which converges weakly
to the nondegenerate $Z$ with $\{|Z_n|^{\gamma}\}$ uniformly integrable, the quantity
\[
    \varrho ~:=~ \left(\frac{\liminf_{n}\disp{\gamma}{Z_n}}{\mathbb{E}\left[\left|Z\right|^{\gamma}\right]}\right)^{1/\gamma}
\]
lies in $(0,1]$; it is at most one because $t=0$ is feasible in equation~\eqref{eqn:dispersion}, and it
does not depend on the radius $\eta$. Fix once and for all an $\varepsilon$ with
$0<\varepsilon<|q_F'(x^*)|\varrho/2$. Choose $\eta$ small enough that the comparisons in equations~\eqref{eqn:loc_P}--\eqref{eqn:loc_phi} hold on $\mathcal{N}$ with that
$\varepsilon$ and that equation~\eqref{eqn:loc_radius} holds. The second-moment escape~\eqref{eqn:asym_localization} then gives $R_{\mathrm{PS}}=o(\rate^{2})$ and $R_{\mathrm{SR}}=o(\rate^{2})$ on the corresponding $A_n$. The premium remainder required by the lemma is $o(\rate^{\gamma})$, which this supplies when $\gamma\le 2$ and which is otherwise the hypothesis.

\paragraph{Expected post-sample regret.}
The bracket in equation~\eqref{eqn:bracket_P} together with equation~\eqref{eq:asym_scaling} gives
\[
    c_2\,\rate^{2}\,\mathbb{E}\!\left[\left|Z_n\right|^{2}\mathbf{1}_{A_n}\right]
    ~\leq~ \Pcomp{\pi}{F} ~\leq~
    C_2\,\rate^{2}\,\mathbb{E}\!\left[\left|Z_n\right|^{2}\mathbf{1}_{A_n}\right]+R_{\mathrm{PS}} \ .
\]
The moment converges to a finite positive limit by equation~\eqref{eq:asym_moments} and
$R_{\mathrm{PS}}=o(\rate^{2})$ by equation~\eqref{eqn:asym_localization}, so
$\Pcomp{\pi}{F}=\Theta(\rate^{2})$.

\paragraph{Sample-risk premium.}
By equation~\eqref{eqn:bracket_q}, homogeneity and equation~\eqref{eq:asym_scaling},
\[
    \left|\;\dispA{\gamma}{\qv{F}{x^\pi}}{A_n}^{1/\gamma}
    ~-~\left|q_F'(x^*)\right|\rate\,\dispA{\gamma}{Z_n}{A_n}^{1/\gamma}\;\right|
    ~~\leq~~ \varepsilon\,\rate\left(\mathbb{E}\!\left[\left|Z_n\right|^{\gamma}\mathbf{1}_{A_n}\right]\right)^{1/\gamma} \ .
\]
For the upper bound, $\dispA{\gamma}{Z_n}{A_n}\leq\mathbb{E}[|Z_n|^{\gamma}\mathbf{1}_{A_n}]$
and equation~\eqref{eq:asym_moments} give
$\dispA{\gamma}{\qv{F}{x^\pi}}{A_n}^{1/\gamma}=O(\rate)$. For the lower bound,
divide by $\rate$ and use equation~\eqref{eq:restricted_dispersion} with $Y_n=Z_n$ and
$E_n=A_n$, together with equation~\eqref{eq:asym_moments} and the definition of $\varrho$:
\[
    \liminf_{n\to\infty}\frac{\dispA{\gamma}{\qv{F}{x^\pi}}{A_n}^{1/\gamma}}{\rate}
    ~~\geq~~ \left(\left|q_F'(x^*)\right|\varrho-\varepsilon\right)\left(\mathbb{E}\!\left[\left|Z\right|^{\gamma}\right]\right)^{1/\gamma}
    ~~>~~ 0 \ ,
\]
strictly positive by the choice $\varepsilon<|q_F'(x^*)|\varrho/2$. Hence
$\dispA{\gamma}{\qv{F}{x^\pi}}{A_n}=\Theta(\rate^{\gamma})$, and
the bracket in equation~\eqref{eqn:bracket_SR},
\[
    c_\gamma\,\dispA{\gamma}{\qv{F}{x^\pi}}{A_n}
    ~\leq~ \SRcomp{\pi}{F} ~\leq~
    C_\gamma\,\dispA{\gamma}{\qv{F}{x^\pi}}{A_n}+R_{\mathrm{SR}} \ ,
\]
with $R_{\mathrm{SR}}=o(\rate^{\gamma})$ by the hypothesis of the lemma, which equation~\eqref{eqn:asym_localization} verifies when $\gamma\le 2$, gives
$\SRcomp{\pi}{F}=\Theta(\rate^{\gamma})$. Dividing the two orders
yields $\rho_n=\Theta(\rate^{\gamma-2})$ and completes equation~\eqref{eqn:asym_orders}.

\subsection{Proof of Lemma~\ref{lem:fixed_n}}\label{prf:fixed_n}
Throughout write $x^*:=\xopt{F}$, $q^*:=\qv{F}{x^*}$, $\Delta:=x^\pi-x^*$ and
$A:=\{x^\pi\in\mathcal{N}\}$. For a fixed action $x$ set
\[
h_x(u) ~:=~ \phiv{F}{x}{\qv{F}{x}+u}-\Qc{F}{x} \ .
\]
Since $(c(x,\xi)-t)^+$ is convex in $t$, the map $t\mapsto\phiv{F}{x}{t}$ is convex, and hence
so is $h_x$. By definition of $\qv{F}{x}$ as a minimizer, $h_x\geq0=h_x(0)$. A convex
function attaining its minimum at the origin is nonincreasing on $(-\infty,0]$ and
nondecreasing on $[0,\infty)$, so $|u|\mapsto h_x(u)$ is nondecreasing on each side of the
origin. We use this monotonicity twice below. If $\mathbb{P}(A)=0$ the terms supported on $A$ vanish, and the inequalities reduce to the remainder bounds $0\leq\Pcomp{\pi}{F}\leq R_{\mathrm{PS}}$ and $0\leq\SRcomp{\pi}{F}\leq R_{\mathrm{SR}}$. Assume $\mathbb{P}(A)>0$.

\paragraph{Geometry of the neighborhood.}
Comparing the two bounds in~\eqref{eqn:loc_phi} at any $u\neq 0$ gives $c_\gamma\leq C_\gamma$. The radius condition~\eqref{eqn:loc_radius} keeps every induced cutoff inside the interval on which those bounds apply. Its right-hand side is at most $u_0/4$. Set $\kappa:=(|q_F'(x^*)|+\varepsilon)\eta$. By the $\VaR$ expansion~\eqref{eqn:loc_q}, for every
$x\in\mathcal{N}$,
\begin{equation}\label{eq:q_range}
\left|\qv{F}{x}-q^*\right|
~\leq~
\left|q_F'(x^*)\right|\left|x-x^*\right|+\varepsilon\left|x-x^*\right|
~\leq~
\kappa \ ,
\end{equation}
and the radius condition~\eqref{eqn:loc_radius} gives $\kappa\leq u_0/4$. Let
$T:=[q^*-u_0/2,\,q^*+u_0/2]$. Two consequences follow. First, for $t\in T$ and
$x\in\mathcal{N}$,
\[
\left|t-\qv{F}{x}\right|
~\leq~
\left|t-q^*\right|+\left|\qv{F}{x}-q^*\right|
~\leq~
\tfrac{u_0}{2}+\tfrac{u_0}{4}
~\leq~
u_0 \ ,
\]
so the growth bounds~\eqref{eqn:loc_phi} apply. Second, for $t\notin T$ and $x\in\mathcal{N}$,
\begin{equation}\label{eq:far_threshold}
\left|t-\qv{F}{x}\right|
~\geq~
\left|t-q^*\right|-\kappa
~>~
\tfrac{u_0}{2}-\tfrac{u_0}{4}
~=~
\tfrac{u_0}{4} \ .
\end{equation}

\paragraph{Expected post-sample regret.}
By Proposition~\ref{prop:endpoints} the integrand $\Qc{F}{x^\pi}-\Qc{F}{x^*}$ is
nonnegative, so splitting on $A$,
\[
\Pcomp{\pi}{F}
=
\mathbb{E}\!\left[\left(\Qc{F}{x^\pi}-\Qc{F}{x^*}\right)\mathbf{1}_{A}\right]
+
R_{\mathrm{PS}} \ ,
\]
with $R_{\mathrm{PS}}\geq0$. Applying~\eqref{eqn:loc_P} on $A$ to the first term and
discarding $R_{\mathrm{PS}}$ in the lower bound yields~\eqref{eqn:bracket_P}.

\paragraph{The premium as a threshold problem.}
By equation~\eqref{eqn:sr_variational}, and because $\mathbb{E}[\Qc{F}{x^\pi}]$ is finite and
free of $t$,
\[
\SRcomp{\pi}{F}
~=~
\inf_{t\in\mathbb{R}}\mathbb{E}\!\left[\phiv{F}{x^\pi}{t}-\Qc{F}{x^\pi}\right]
~=~
\inf_{t\in\mathbb{R}}\left\{J(t)+J^{c}(t)\right\} \ ,
\]
where
\[
J(t):=\mathbb{E}\!\left[\left(\phiv{F}{x^\pi}{t}-\Qc{F}{x^\pi}\right)\mathbf{1}_{A}\right] \ ,
\qquad
J^{c}(t):=\mathbb{E}\!\left[\left(\phiv{F}{x^\pi}{t}-\Qc{F}{x^\pi}\right)\mathbf{1}_{A^{c}}\right] \ .
\]
Both are nonnegative, since $\qv{F}{x}$ minimizes $t\mapsto\phiv{F}{x}{t}$ pointwise. Write
$G(t):=\mathbb{E}[|t-\qv{F}{x^\pi}|^{\gamma}\mathbf{1}_{A}]$, so that
$\inf_{t}G(t)=\dispA{\gamma}{\qv{F}{x^\pi}}{A}$.

\paragraph{Restricting the threshold to $T$.}
We first check that $\inf_{t\in\mathbb{R}}J(t)=\inf_{t\in T}J(t)$. For $t\notin T$,
combining~\eqref{eq:far_threshold} with the monotonicity of $h_x$ in $|u|$ and the lower
bound in~\eqref{eqn:loc_phi} evaluated at $|u|=u_0/4$ gives
\[
J(t) ~\geq~ c_\gamma\left(\tfrac{u_0}{4}\right)^{\gamma}\mathbb{P}(A) \ .
\]
On the other hand, taking $t=q^*\in T$ and using~\eqref{eq:q_range} with the upper bound
in~\eqref{eqn:loc_phi},
\[
J(q^*)
~\leq~
C_\gamma\,\mathbb{E}\!\left[\left|q^*-\qv{F}{x^\pi}\right|^{\gamma}\mathbf{1}_{A}\right]
~\leq~
C_\gamma\kappa^{\gamma}\mathbb{P}(A)
~\leq~
C_\gamma\left(\tfrac{u_0}{4}\right)^{\gamma}\frac{c_\gamma}{2C_\gamma}\mathbb{P}(A)
~=~
\frac{c_\gamma}{2}\left(\tfrac{u_0}{4}\right)^{\gamma}\mathbb{P}(A) \ ,
\]
where the third inequality is the radius condition~\eqref{eqn:loc_radius}. Since
$\mathbb{P}(A)>0$, the value at $q^*$ is strictly below the bound available off $T$, so no
$t\notin T$ can be optimal. The same comparison applies to $G$, and in addition the convex
map $t\mapsto G(t)$ has a minimizer in the convex hull of the essential range of
$\qv{F}{x^\pi}$ on $A$, which by~\eqref{eq:q_range} lies in $[q^*-\kappa,q^*+\kappa]\subseteq T$.
Hence $\inf_{t\in T}G(t)=\inf_{t\in\mathbb{R}}G(t)=\dispA{\gamma}{\qv{F}{x^\pi}}{A}$, and the
infimum is attained at some $\hat{t}\in T$ because $G$ is continuous on the compact set $T$.

\paragraph{Premium: lower bound.}
Dropping the nonnegative term $J^{c}$ and using the restriction just established,
\[
\SRcomp{\pi}{F}
~\geq~
\inf_{t\in\mathbb{R}}J(t)
~=~
\inf_{t\in T}J(t)
~\geq~
c_\gamma\inf_{t\in T}G(t)
~=~
c_\gamma\,\dispA{\gamma}{\qv{F}{x^\pi}}{A} \ ,
\]
where the second inequality holds because $J(t)\geq c_\gamma G(t)$ pointwise on $T$
by~\eqref{eqn:loc_phi}.

\paragraph{Premium: upper bound.}
Evaluating the infimum at the feasible threshold $\hat{t}\in T$ and
using $J(\hat t)\leq C_\gamma G(\hat t)$ on $T$,
\[
\SRcomp{\pi}{F}
~\leq~
J(\hat{t})+J^{c}(\hat{t})
~\leq~
C_\gamma\,\dispA{\gamma}{\qv{F}{x^\pi}}{A}+R_{\mathrm{SR}} \ ,
\]
since $\hat{t}\in T$ and $R_{\mathrm{SR}}$ is by definition the supremum of $J^{c}$ over $T$.
This establishes~\eqref{eqn:bracket_SR}.

\paragraph{Dispersion of the induced $\VaR$.}
Write $\|Y\|:=(\mathbb{E}[|Y|^{\gamma}\mathbf{1}_{A}])^{1/\gamma}$, which is a seminorm because
$\gamma\geq1$. On $A$ decompose $\qv{F}{x^\pi}-q^*=q_F'(x^*)\Delta+e$ where, by~\eqref{eqn:loc_q},
$|e|\leq\varepsilon|\Delta|$ and hence $\|e\|\leq\varepsilon\|\Delta\|$. Substituting
$t=q^*+s$ and using shift invariance,
\[
\dispA{\gamma}{\qv{F}{x^\pi}}{A}^{1/\gamma}=\inf_{s}\left\|q_F'(x^*)\Delta+e-s\right\| \ ,
\qquad
\left|q_F'(x^*)\right|\dispA{\gamma}{\Delta}{A}^{1/\gamma}=\inf_{s}\left\|q_F'(x^*)\Delta-s\right\| \ .
\]
For each $s$ the triangle inequality gives
$\bigl|\,\|q_F'(x^*)\Delta+e-s\|-\|q_F'(x^*)\Delta-s\|\,\bigr|\leq\|e\|$, and two infima whose
integrands differ by at most $\|e\|$ uniformly differ by at most $\|e\|$. This
is~\eqref{eqn:bracket_q}.

\subsection{Proof of Lemma~\ref{lem:smooth_exponents}}\label{prf:smooth_brackets}

Fix $\theta>0$. Since $x^*$ is an interior minimizer of $Q_F$ and $Q_F$ is twice differentiable
at $x^*$, $Q_F'(x^*)=0$, and Taylor's theorem with Peano remainder gives
\[
Q_F(x)-Q_F(x^*)=\tfrac{1}{2}Q_F''(x^*)(x-x^*)^{2}+o\!\left((x-x^*)^{2}\right) \ ,
\]
so~\eqref{eqn:loc_P} holds with constants within $\theta$ of
$\tfrac{1}{2}Q_F''(x^*)$ once $\eta$ is small enough; positivity of $Q_F''(x^*)$ makes the
lower constant positive for $\theta$ small. The expansion~\eqref{eqn:loc_q} with
$q_F'(x^*)\neq0$ and $\varepsilon$ arbitrarily small is the definition of that
derivative, and the expansion persists on every smaller neighborhood.
Differentiability of $q_F$ at $x^*$ supplies continuity of $q_F$ at $x^*$. For the threshold direction, differentiating the
variational objective twice in $t$ gives
$\partial_{t}^{2}\phiv{F}{x}{t}=g_x(t)/(1-\delta)$, while
$\partial_{t}\phiv{F}{x}{\qv{F}{x}}=0$ by optimality. Joint continuity of $(x,t)\mapsto g_x(t)$ at
$(x^*,q^*)$, positivity and finiteness of $g$, both assumed in Theorem~\ref{prop:asymptotic_characterization}, and continuity of $q_F$ at $x^*$ together imply
that $g_x(\tau)$ lies within $\theta'$ of $g$ for all $x\in\mathcal{N}$ and all $\tau$ between
$\qv{F}{x}$ and $\qv{F}{x}+u$ with $|u|\leq u_0$, provided $\eta$ and $u_0$ are small enough.
Taylor's theorem in $t$ then yields
\[
h_x(u)=\tfrac{1}{2}\frac{g_x(\tau)}{1-\delta}u^{2} \ ,
\]
so~\eqref{eqn:loc_phi} holds with $\gamma=2$ and constants within $\theta$ of
$g/(2(1-\delta))$. Choose $\eta$ small enough that~\eqref{eqn:loc_radius} holds; shrinking the neighborhood does not disturb
these bounds. One consequence of equation~\eqref{eqn:loc_phi} is that the minimizing set of the variational objective from equation~\eqref{eqn:variational_objective} is a singleton on the neighborhood: if it contained an interval, the left-hand side of equation~\eqref{eqn:loc_phi} would be strictly positive at an interior $u$ while the right-hand side would vanish. Hence that set equals $\{\qv{F}{x}\}$ for every $x\in\mathcal{N}$.

\subsection{Proof of Proposition~\ref{prop:smooth_brackets}}\label{prf:smooth_local}

{Under Theorem~\ref{prop:asymptotic_characterization}, $\gamma=2$, so the dispersion is a restricted variance.} Minimizing $t\mapsto\mathbb{E}[(W-t)^{2}\mathbf{1}_{B}]$ at
$t=\mathbb{E}[W\mathbf{1}_{B}]/\mathbb{P}(B)$ gives the identity
\[
    \dispA{2}{W}{B} ~=~ \mathbb{E}\!\left[W^{2}\mathbf{1}_{B}\right]-\frac{\left(\mathbb{E}\left[W\mathbf{1}_{B}\right]\right)^{2}}{\mathbb{P}(B)} \ ,
\]
so~\eqref{eq:asym_moments} at $\gamma=2$ and the first-moment
limit~\eqref{eq:asym_first_moment}, together with $\mathbb{P}(A_n)\to1$, upgrade
the liminf bound of~\eqref{eq:restricted_dispersion} to convergence:
$\dispA{2}{Z_n}{A_n}\to\mathbb{E}[Z^{2}]-\mathbb{E}[Z]^{2}=\Var(Z)$ and
$\mathbb{E}[Z_n^{2}\mathbf{1}_{A_n}]\to\mathbb{E}[Z^{2}]$, the latter strictly positive.
The same uniform integrability gives the unrestricted moments $\mathbb{E}[Z_n^{2}]\to\mathbb{E}[Z^{2}]$ and $\mathbb{E}[Z_n]\to\mathbb{E}[Z]$, hence $\Var(Z_n)\to\Var(Z)$.

Fix $\theta>0$ and let $\eta$ be the radius supplied by the smooth case in Appendix~\ref{prf:smooth_brackets}
for that $\theta$, so that $c_2$ and $C_2$ lie within $\theta$ of
$\tfrac{1}{2}Q_F''(x^*)$, $c_\gamma$ and $C_\gamma$ within $\theta$ of $g/(2(1-\delta))$, and
$\varepsilon\leq\theta$. The bracket~\eqref{eqn:bracket_q} at
$\gamma=2$ gives
\[
    \left|\;\dispA{2}{\qv{F}{x^\pi}}{A_n}^{1/2}-\left|q_F'(x^*)\right|\rate\,\dispA{2}{Z_n}{A_n}^{1/2}\;\right|
    ~\leq~ \theta\,\rate\left(\mathbb{E}\!\left[Z_n^{2}\mathbf{1}_{A_n}\right]\right)^{1/2} \ ,
\]
so that, letting $n\to\infty$ at fixed $\theta$, both the limit inferior and the limit superior
of $\dispA{2}{\qv{F}{x^\pi}}{A_n}^{1/2}/\rate$ lie within $\theta\,\mathbb{E}[Z^{2}]^{1/2}$ of
$\left|q_F'(x^*)\right|\Var(Z)^{1/2}$. No claim is made that the limit exists at fixed $\theta$; the
bounds below are applied to the two extreme limits and collapse only as $\theta\downarrow0$.

Read separately, before the ratio is formed, the two brackets of Lemma~\ref{lem:fixed_n} are the limits in Proposition~\ref{prop:smooth_brackets}. The post-sample bracket, with $R_{\mathrm{PS}}=o(\rate^{2})$, pins $\Pcomp{\pi}{F}/\rate^{2}$ within $\theta$ of $\tfrac12 Q_F''(x^*)\,\mathbb{E}[Z^{2}]$ once $n$ is large. The premium bracket, fed with the display above, likewise pins $\SRcomp{\pi}{F}/\rate^{2}$ within a multiple of $\theta$ of
\[
    \frac{g\,q_F'\!\left(x^*\right)^{2}}{2\left(1-\delta\right)}\cdot\Var(Z) \ .
\]
Letting $n\to\infty$ and then $\theta\downarrow0$ gives
\[
    \frac{\Pcomp{\pi}{F}}{\rate^{2}} ~~\longrightarrow~~ \frac{Q_F''\!\left(x^*\right)}{2}\cdot\mathbb{E}\!\left[Z^{2}\right] \ ,
    \qquad
    \frac{\SRcomp{\pi}{F}}{\rate^{2}} ~~\longrightarrow~~ \frac{g\,q_F'\!\left(x^*\right)^{2}}{2\left(1-\delta\right)}\cdot\Var(Z) \ .
\]
Since $\mathrm{MSE}=\rate^{2}\,\mathbb{E}[Z_n^{2}]$ and $V=\rate^{2}\,\Var(Z_n)$, dividing through by these moments is~\eqref{eqn:smooth_brackets}.

\subsection{Proof of Corollary~\ref{cor:rho_limit}}\label{prf:rho_limit}

Dividing the two limits in the proof of Proposition~\ref{prop:smooth_brackets} yields
$\rho_n\to\Kprice\cdot\Var(Z)/\mathbb{E}[Z^{2}]$, by the definition of $\Kprice$ in~\eqref{eqn:K}. When $\mathbb{E}[Z]=0$ the ratio of the
variance to the second moment is one, which is~\eqref{eqn:rho_limit}.

\subsection{\texorpdfstring{{The fixed-threshold curvature in equation~\eqref{eqn:frozen_curvature}}}{Fixed-threshold curvature}}\label{prf:frozen_curvature}

{Let $Q''=Q_F''(x^*)$, $a_F=q_F'(x^*)$, and $b=g/(1-\delta)$. The smooth hypotheses give $Q_F(x^*+h)=Q_F(x^*)+\tfrac12 Q'' h^2+o(h^2)$ and $q_F(x^*+h)=q^*+a_F h+o(|h|)$. Integrating the threshold derivative, using joint continuity of the loss density at $(x^*,q^*)$, gives
\[
\phi_F(x,q_F(x)+v)-Q_F(x)=\tfrac12 b v^2+o(v^2)
\]
as $(x,v)\to(x^*,0)$, with a remainder uniform over nearby $x$. Consequently,
\[
\phi_F(x^*+h,q^*+u)=Q_F(x^*)+\tfrac12 Q'' h^2+\tfrac12b(u-a_F h)^2+o(h^2+u^2).
\]
Taking $u=0$ gives $A_F=Q''+b a_F^2$ and $1+K=A_F/Q''$. The regret expansion~\eqref{eqn:reduction} can therefore be written
\[
\mathcal R_{\delta,n}(\pi,F)
=\tfrac12\left\{Q''\bigl(\mathbb E[x^\pi-x^*]\bigr)^2+A_F\operatorname{Var}(x^\pi)\right\}+o(r_n^2).
\]
This argument uses the local expansion, rather than existence of classical second partial derivatives of $\phi_F$ throughout a neighborhood.}

\subsection{The zero-slope cutoff endpoint}\label{prf:flat_cutoff}

\begin{proposition}[Zero-slope cutoff]\label{prop:flat_cutoff}
Suppose that the hypotheses of Theorem~\ref{prop:asymptotic_characterization} hold, except that $q_F'(x^*)=0$. Then
\[
    \SRcomp{\pi}{F} ~~=~~ o\!\left(\rate^{2}\right) \ ,
\]
while
\[
    \Pcomp{\pi}{F}
    ~~\sim~~
    \tfrac12 Q_F''\!\left(x^*\right)\,\rate^{2}\,\mathbb{E}[Z^{2}]
    ~~=~~
    \Theta\!\left(\rate^{2}\right) \ .
\]
\end{proposition}

\begin{proof}
For any $\varepsilon>0$, differentiability gives $|q_F(x)-q^*|\le\varepsilon|x-x^*|$ on a sufficiently small neighborhood. Evaluate the common threshold in equation~\eqref{eqn:sr_variational} at $q^*$. The local threshold curvature is bounded, so its contribution is at most $C\varepsilon^2\mathbb E[(x^\pi-x^*)^2]$, with $C$ bounded as the neighborhood shrinks. The escape contribution is $o(\rate^{2})$ by the same envelope and uniform-integrability argument as in equation~\eqref{eqn:asym_localization}. Hence
\[
0\le\limsup_n\frac{\SRcomp{\pi}{F}}{\rate^{2}}
\le C\varepsilon^2\mathbb E[Z^2].
\]
Letting $\varepsilon\downarrow0$ gives $\SRcomp{\pi}{F}=o(\rate^{2})$. The post-sample expansion uses only the curvature of $Q_F$ and the moment conditions and still yields $\Pcomp{\pi}{F}\sim Q_F''(x^*)\rate^{2}\mathbb E[Z^2]/2$.
\end{proof}

\subsection{\texorpdfstring{{The relative variance penalty for squared loss}}{The relative variance penalty for squared loss}}\label{apx:location_statements}

{Table~\ref{tab:K_delta} evaluates the relative variance penalty~\eqref{eqn:K}} for the squared loss $c(x,\xi)=(\xi-x)^{2}/2$. The second derivative of this loss in the deviation $\xi-x$ is the constant $1$. In the tail expectation $T(x)=\mathbb{E}\bigl[(\xi-x)\,\mathbf{1}_{\{|\xi-x|>\sqrt{2q_F(x)}\}}\bigr]$, the factor $\xi-x$ has action derivative $-1$, so that contribution is dominated by the constant $1$. A finite second moment makes the factor itself integrable. Differentiating $T$ also moves the tail boundaries; the proof handles that term by Leibniz's rule.

\begin{proposition}[{The relative variance penalty for squared loss}]\label{prop:K_squared}
Fix $\delta\in(0,1)$ and $c(x,\xi)=(\xi-x)^{2}/2$, and assume $\mathbb{E}[\xi^{2}]<\infty$. Let $x^*$ be an interior oracle action, write $q^*:=\qv{F}{x^*}$ and $d:=\sqrt{2q^*}$, and let $f_+:=f(x^*+d)$, $f_-:=f(x^*-d)$. Suppose $F$ admits a density $f$ continuous in a neighborhood of each of $x^*+d$ and $x^*-d$, both interior to the support of $F$, {with $f_++f_->0$. Then}
\begin{equation}\label{eqn:K_squared}
\text{{$\displaystyle \Kprice=\frac{d(f_--f_+)^2}{(1-\delta)(f_++f_-)+4df_+f_-}.$}}
\end{equation}
{The two curvatures are}
\begin{equation}\label{eqn:squared_curvatures}
\text{{$\displaystyle \begin{aligned}
Q_F''(x^*)&=1+\frac{4df_+f_-}{(1-\delta)(f_++f_-)}\ge1,\\
A_F&=1+\frac{d(f_++f_-)}{1-\delta}<\infty.
\end{aligned}$}}
\end{equation}
In particular $\Kprice=0$ if and only if $f_+=f_-$.
\end{proposition}

\subsection{Proof of Proposition~\ref{prop:K_squared}}\label{prf:var_slope}

The loss equals $t$ at the deviations $\pm\sqrt{2t}$, so
\[
    \mathbb{P}\bigl(c(x,\xi)\leq t\bigr)
    ~=~
    F\bigl(x+\sqrt{2t}\bigr)-F\bigl(x-\sqrt{2t}\bigr) \ .
\]
The left-hand side is $0$ at $t=0$ and the interior-support hypothesis places the $\delta$-crossing in the open support. Since $f_++f_->0$, the right-hand side is strictly increasing there, and $q:=q_F(x)$ is its unique nonnegative root. Differentiating in $t$ at $t=q$, with $\frac{\mathrm{d}}{\mathrm{d}t}\sqrt{2t}=1/\sqrt{2t}$, gives the density of the loss at the cutoff,
\[
    g ~=~ \frac{f_++f_-}{d} \ .
\]
Set $G(x,q):=F(x+\sqrt{2q})-F(x-\sqrt{2q})-\delta$. Continuity of $f$ near the two critical levels makes $G$ continuously differentiable near $(x^*,q^*)$, with $\partial_x G=f_+-f_-$ and $\partial_q G=g>0$. The implicit function theorem yields
\begin{equation}\label{eqn:var_slope}
    q_F'(x^*)
    ~=~
    \frac{f_--f_+}{g}
\end{equation}
and makes $q_F$ continuously differentiable near $x^*$.

Write $\phi(x,t)=t+(1-\delta)^{-1}\mathbb{E}\bigl[\bigl((\xi-x)^{2}/2-t\bigr)_+\bigr]$. The integrand vanishes at both critical levels, so
\[
    \partial_x\phi(x,t)
    ~=~
    -\frac{1}{1-\delta}\,\mathbb{E}\bigl[(\xi-x)\,\mathbf{1}_{\{c(x,\xi)>t\}}\bigr] \ .
\]
The envelope theorem at the minimizing threshold $t=q(x)$ gives $Q'(x)=-(1-\delta)^{-1}T(x)$, where
\[
    T(x)
    ~:=~
    \mathbb{E}\bigl[(\xi-x)\,\mathbf{1}_{\{|\xi-x|>\sqrt{2q(x)}\}}\bigr] \ .
\]
The interior oracle condition is $T(x^*)=0$. On that tail the derivative of $\xi-x$ in the action is $-1$. For every action near $x^*$ this derivative is bounded by the constant $1$, which does not depend on the action; $\mathbb{E}[\xi^{2}]<\infty$ makes the integrand integrable. The boundaries $x\pm\sqrt{2q(x)}$ are continuously differentiable because $q_F$ is, and $f$ is continuous there, so Leibniz's rule applies:
\begin{align*}
    T'(x^*)
    &~=~
    -\bigl(1+\tfrac{q'}{d}\bigr)\,d\,f_+
    -\bigl(1-\tfrac{q'}{d}\bigr)\,d\,f_-
    -\mathbb{E}\bigl[\mathbf{1}_{\{c(x^*,\xi)>q^*\}}\bigr] \\
    &~=~
    -d\bigl(f_++f_-\bigr)-(1-\delta)+q'(f_--f_+) \ .
\end{align*}
{The indicator has probability $1-\delta$ because the loss has a density at its Value-at-Risk. Since $q'=(f_--f_+)/g$ and $g=(f_++f_-)/d$, the expression for $T'$ gives the first identity in~\eqref{eqn:squared_curvatures}. It establishes positive finite curvature from the remaining hypotheses. Adding $gq'^2/(1-\delta)$ yields the fixed-threshold curvature $A_F$, and substituting in~\eqref{eqn:K} gives~\eqref{eqn:K_squared}. A positive affine transformation of the loss multiplies both curvatures by the same positive factor. For squared loss, a location--scale change of the outcome replaces $d$ by a positive multiple and $f_\pm$ by the reciprocal multiple, leaving $K$ unchanged.}

\subsection{Proof of Theorem~\ref{cor:atom_category}}\label{prf:atom_category}

Write $p:=\mathbb{P}(c(x^*,\xi)=q^*)$ and
\begin{equation}\label{eqn:atom_slopes}
    s_+ ~:=~ \frac{\mathbb{P}(c(x^*,\xi)\leq q^*)-\delta}{1-\delta} \ ,
    \quad
    s_- ~:=~ \frac{\delta-\mathbb{P}(c(x^*,\xi)<q^*)}{1-\delta} \ ,
\end{equation}
so that $s_++s_-=p/(1-\delta)$. Both are positive under the straddling hypothesis of Theorem~\ref{cor:atom_category}.

\begin{lemma}[Atom exponents]\label{lem:atom_exponents}
Suppose the hypotheses of Theorem~\ref{cor:atom_category}. For every $\theta>0$ there is a neighborhood of $x^*$ on which~\eqref{eqn:loc_P}--\eqref{eqn:loc_phi} hold with $\gamma=1$ and $q_F'(x^*)=\ell'(x^*)\neq0$, with $\varepsilon\leq\theta$, and with $c_\gamma$ and $C_\gamma$ within $\theta$ of $\min\{s_+,s_-\}$ and $\max\{s_+,s_-\}$.
\end{lemma}

Lemma~\ref{lem:atom_exponents} gives~\eqref{eqn:loc_P}--\eqref{eqn:loc_phi} with $\gamma=1$, $q_F'(x^*)=\ell'(x^*)\neq0$ and, for every $\theta>0$, with $\varepsilon\leq\theta$. Lemma~\ref{lem:orders} at this exponent gives $\Pcomp{\pi}{F}=\Theta(\rate^{2})$ and $\SRcomp{\pi}{F}=\Theta(\rate)$.

\subsection{Proof of Lemma~\ref{lem:atom_exponents}}

Write $S_x(t):=\mathbb{P}(c(x,\xi)>t)$ for the survival function of the loss at the action $x$.
Throughout, $\theta>0$ is fixed and $\eta$ and $u_0$ are shrunk finitely many times.

\paragraph{The threshold direction.}
Differentiating the variational objective in the threshold, as in
equation~\eqref{eqn:phi_derivative}, gives the right derivative $\partial_t^{+}\phiv{F}{x}{t}=1-S_x(t)/(1-\delta)$. Where $S_x$ is continuous this is the ordinary derivative, and $t\mapsto\phiv{F}{x}{t}$ is convex and Lipschitz, so for
every $w>0$,
\[
\begin{aligned}
\phiv{F}{x}{\ell(x)+w}-\phiv{F}{x}{\ell(x)}
~&=~
\int_0^{w}\left(1-\frac{S_x\left(\ell(x)+v\right)}{1-\delta}\right)dv \ , \\
\phiv{F}{x}{\ell(x)-w}-\phiv{F}{x}{\ell(x)}
~&=~
\int_0^{w}\left(\frac{S_x\left(\ell(x)-v\right)}{1-\delta}-1\right)dv \ .
\end{aligned}
\]
The integrands are monotone in $v$, since $S_x$ is nonincreasing. Bounding each by its values
at $v\to0^+$ and at $v=w\leq u_0$ gives, for every $x\in\mathcal{N}$ and every $0<w\leq u_0$,
\[
\left(1-\frac{S_x\left(\ell(x)\right)}{1-\delta}\right)w
~\leq~
\phiv{F}{x}{\ell(x)+w}-\phiv{F}{x}{\ell(x)}
~\leq~
\left(1-\frac{S_x\left(\ell(x)+u_0\right)}{1-\delta}\right)w \ ,
\]
and
\[
\left(\frac{\mathbb{P}\left(c\left(x,\xi\right)\geq\ell(x)\right)}{1-\delta}-1\right)w
~\leq~
\phiv{F}{x}{\ell(x)-w}-\phiv{F}{x}{\ell(x)}
~\leq~
\left(\frac{\mathbb{P}\left(c\left(x,\xi\right)\geq\ell(x)-u_0\right)}{1-\delta}-1\right)w \ .
\]
The continuity of the rest of the loss distribution at the cutoff, assumed in Theorem~\ref{cor:atom_category}, is the pair of limits
\begin{equation}
\begin{aligned}
    \lim_{\left(x,u\right)\to\left(x^*,0^+\right)}\mathbb{P}\left(c\left(x,\xi\right)>\ell\left(x\right)+u\right) ~&=~ \mathbb{P}\left(c\left(x^*,\xi\right)>q^*\right) \ , \\
    \lim_{\left(x,u\right)\to\left(x^*,0^+\right)}\mathbb{P}\left(c\left(x,\xi\right)\geq\ell\left(x\right)-u\right) ~&=~ \mathbb{P}\left(c\left(x^*,\xi\right)\geq q^*\right) \ .
\end{aligned}
\end{equation}
The four coefficients converge, as $\eta$ and $u_0$ shrink, to $s_+$, $s_+$, $s_-$ and $s_-$
respectively: this is exactly~\eqref{eqn:atom_isolation} together with the
definitions in~\eqref{eqn:atom_slopes}. Choose $\eta$ and $u_0$ small enough that all four lie
within $\theta$ of their limits. Since $s_+>0$ and $s_->0$ by the straddling hypothesis, both
one-sided increments are strictly positive for $\theta$ small, so $\ell(x)$ is the unique
minimizer of $t\mapsto\phiv{F}{x}{t}$, that is,
\[
    \qv{F}{x} ~=~ \ell(x) \qquad\text{for every }x\in\mathcal{N} \ .
\]
The two displays are therefore exactly the growth condition~\eqref{eqn:loc_phi} with
$\gamma=1$, $c_\gamma\geq\min\{s_+,s_-\}-\theta$ and $C_\gamma\leq\max\{s_+,s_-\}+\theta$.
The slope identity in~\eqref{eqn:atom_slopes} is immediate:
\[
    s_++s_-
    ~=~
    \frac{\mathbb{P}\left(c\left(x^*,\xi\right)\leq q^*\right)-\mathbb{P}\left(c\left(x^*,\xi\right)<q^*\right)}{1-\delta}
    ~=~
    \frac{\mathbb{P}\left(c\left(x^*,\xi\right)=q^*\right)}{1-\delta}
    ~=~
    \frac{p}{1-\delta} \ ,
\]
and $s_+=s_-$ precisely when the two gaps are equal, in which case each equals $p/(2(1-\delta))$.

\paragraph{The action direction.}
Since $\qv{F}{x}=\ell(x)$ on $\mathcal{N}$ and $\ell$ is continuously differentiable near
$x^*$, once $\eta$ is small enough one has $\left|\ell'(x)-q_F'(x^*)\right|\leq\varepsilon$ throughout
$\mathcal{N}$, with $q_F'(x^*)=\ell'(x^*)\neq0$, $\varepsilon\leq\theta$ and $\varepsilon<\left|q_F'(x^*)\right|/2$.
The mean value theorem gives the expansion~\eqref{eqn:loc_q} and the pairwise bound
\[
    \bigl|\,\ell(x)-\ell(y)-q_F'(x^*)(x-y)\,\bigr|
    ~~\leq~~
    \varepsilon\,\left|x-y\right|
    \qquad\text{for all }x,y\in\mathcal{N} \ .
\]
For the growth of the
outcome-only $\cvar$ objective, $x^*$ is an interior minimizer of $Q_F$ and $Q_F$ is twice
differentiable there, so $Q_F'(x^*)=0$ and Taylor's theorem with Peano remainder
gives~\eqref{eqn:loc_P} with constants within $\theta$ of
$\tfrac{1}{2}Q_F''(x^*)$, positive for $\theta$ small because $Q_F''(x^*)>0$. This part of the
argument is identical to the corresponding step in Appendix~\ref{prf:smooth_brackets}; only the threshold direction differs. Choose $\eta$ small enough that~\eqref{eqn:loc_radius} holds. Its
right-hand side is positive because $c_\gamma>0$.

\subsection{Proof of Proposition~\ref{prop:atom_brackets}}\label{prf:atom_brackets}

Write $x^*:=\xopt{F}$, $q^*:=\qv{F}{x^*}$ and $\Delta:=x^\pi-x^*=\rate Z_n$, and for a radius
$\eta>0$ let $\mathcal{N}:=[x^*-\eta,x^*+\eta]$ and $A_n:=\{x^\pi\in\mathcal{N}\}$. Since
$\rate\searrow0$ and $Z_n\Rightarrow Z$, we have $\mathbb{P}(A_n)\to1$ for every fixed $\eta$.

The post-sample limit is the bracket already used for Proposition~\ref{prop:smooth_brackets}. It depends only on the curvature of $Q_F$ at $x^*$, which the atom hypotheses supply, and with $\mathrm{MSE}=\rate^{2}\,\mathbb{E}[Z_n^{2}]$ it is the first limit in~\eqref{eqn:atom_brackets}.

For the premium, fix $\theta\in\bigl(0,\min\{s_+,s_-\}\bigr)$ and take the neighborhood of Lemma~\ref{lem:atom_exponents} on which $q_F(x)=\ell(x)$ and the one-sided increments of $t\mapsto\phiv{F}{x}{t}$ lie within $\theta$ of $s_+$ and $s_-$. On that neighborhood the excess $\phiv{F}{x}{\ell(x)+u}-\Qc{F}{x}$ therefore lies within $\theta|u|$ of $s_+u^{+}+s_-\left(-u\right)^{+}$. The weighted-optimal threshold need not be a median, so the error $\theta|u|$ does not pass through the minimum as $\theta$ times the unweighted absolute deviation. Write $s_{\min}:=\min\{s_+,s_-\}>0$ and
\[
    W_s(t)
    ~~:=~~
    \mathbb{E}\!\left[\bigl(s_+\bigl(t-\ell(x^\pi)\bigr)^{+}+s_-\bigl(\ell(x^\pi)-t\bigr)^{+}\bigr)\mathbf{1}_{A_n}\right] \ .
\]
Let $J(t)$ be the expected excess of the variational objective on $A_n$. The pointwise bound gives $|J(t)-W_s(t)|\leq\theta\,\mathbb{E}[|t-\ell(x^\pi)|\,\mathbf{1}_{A_n}]$. The weighted deviation is at least $s_{\min}$ times that absolute deviation, so $|J(t)-W_s(t)|\leq(\theta/s_{\min})W_s(t)$ for every threshold in the interval $T=[q^*-u_0/2,\,q^*+u_0/2]$ of the proof of Lemma~\ref{lem:fixed_n}. Shrink the action neighborhood so that~\eqref{eqn:loc_radius} holds. That proof shows that both $J$ and $W_s$ are minimized inside $T$: off $T$ the excess exceeds the value at $q^*$. The relative bound is applied for $t\in T$, and the contribution of $A_n^c$ is added afterwards. The infima of $J$ and of $W_s$ over $T$ therefore stand in the ratio $1\pm\theta/s_{\min}$. Uniform integrability of $\bigl\{|Z_n|^{2}\bigr\}$ gives that of $\bigl\{|Z_n|\bigr\}$, and the contribution of $A_n^c$ is $o(\rate)$ by~\eqref{eqn:asym_localization}. On $A_n$ one has $\qv{F}{x^\pi}=\ell(x^\pi)$, so the expansion of $\ell$ gives $\qv{F}{x^\pi}=q^*+\rate\,\ell'(x^*)Z_n+\rate e_n$ with $\mathbb{E}[|e_n|]\to0$ after $\theta\downarrow0$. Since $q_F=\ell$ on the neighborhood, $q_F'(x^*)=\ell'(x^*)$. {Weak convergence of $Z_n$ and uniform integrability of $|Z_n|$ give convergence of their distributions in Wasserstein-$1$. The first-mean cutoff expansion transfers this convergence to the scaled cutoffs. For any coupling, the weighted deviation changes by at most $p\max\{\tau,1-\tau\}\mathbb E|Y-Y'|$, so it is Lipschitz in Wasserstein-$1$. Homogeneity and translation invariance therefore give $\mathcal W(q_F(x^\pi))/r_n\to\mathcal W(q_F'(x^*)Z)$.} Sending $\theta\downarrow0$ removes the relative factor and yields
\[
    \frac{\SRcomp{\pi}{F}}{\rate}
    ~~\longrightarrow~~
    \frac{\mathcal{W}\!\left(q_F'(x^*)\,Z\right)}{1-\delta} \ .
\]
{Since $p>0$, $\tau\in(0,1)$, $q_F'(x^*)\ne0$, and $Z$ is nondegenerate, $\mathcal W(q_F'(x^*)Z)>0$. Dividing produces the second limit in~\eqref{eqn:atom_brackets}.}

The same uniform integrability gives $\disp{1}{Z_n}\to\disp{1}{Z}$ by the argument already used for~\eqref{eq:restricted_dispersion}, and the expansion of $\ell$ multiplies that deviation by $\left|\ell'(x^*)\right|$. The resulting liminf and limsup of $\SRcomp{\pi}{F}/\disp{1}{x^\pi}$ lie between $\min\{s_+,s_-\}\left|\ell'(x^*)\right|$ and $\max\{s_+,s_-\}\left|\ell'(x^*)\right|$.

The scaled ratio~\eqref{eqn:atom_rho} is the quotient of the two limits just obtained. The post-sample limit yields $\Pcomp{\pi}{F}/\rate^{2}\to\tfrac12 Q_F''(x^*)\,\mathbb{E}[Z^{2}]$, so
\[
    \rate\,\rho_n
    ~=~
    \frac{\SRcomp{\pi}{F}/\rate}{\Pcomp{\pi}{F}/\rate^{2}}
    ~~\longrightarrow~~
    \frac{2\,\mathcal{W}\!\left(q_F'(x^*)\,Z\right)}{(1-\delta)\,Q_F''\!\left(x^*\right)\,\mathbb{E}[Z^{2}]} \ .
\]

\subsection{Proof of Proposition~\ref{prop:multiple_optima}}\label{prf:multiple_optima}

The proof proceeds by coupling $x^\pi$ with its weak limit. Continuity and
uniform integrability then imply $L^1$ convergence of the induced losses.
The $L^1$ continuity of $\cvar$ allows us to pass both terms in the regret
decomposition to the limit, after which the strict positivity result follows
from Lemma~\ref{lem:strict_sample_risk_premium} below.

\paragraph{Convergence of the induced losses.}
By the Skorokhod representation theorem, we may construct random variables
$x_n^\pi$ and $x_\infty^\pi$ with the same respective distributions as $x^\pi$ and
$x_\infty^\pi$ such that
\[
    x_n^\pi\longrightarrow x_\infty^\pi
    \qquad\text{almost surely}.
\]
Let $\xi\sim F$ be independent of $(x_n^\pi,x_\infty^\pi)$. By the assumed
continuity of $x\mapsto c(x,\xi)$ at $x_\infty^\pi$,
\[
    c(x_n^\pi,\xi)\longrightarrow c(x_\infty^\pi,\xi)
    \qquad\text{almost surely}.
\]
The family $\{|c(x_n^\pi,\xi)|\}_{n\geq1}$ is uniformly integrable because
$x_n^\pi$ has the same distribution as $x^\pi$. Hence, by uniform
integrability,
\begin{equation}\label{eqn:mixing_loss_l1}
    \mathbb E\left[
        |c(x_n^\pi,\xi)-c(x_\infty^\pi,\xi)|
    \right]
    \longrightarrow0.
\end{equation}

\paragraph{Expected post-sample regret.}
For integrable random variables $Y$ and $Y'$,
\[
    |\cvar_\delta[Y]-\cvar_\delta[Y']|
    \leq
    \frac{1}{1-\delta}\mathbb E|Y-Y'|.
\]
Applying this inequality conditionally on $(x_n^\pi,x_\infty^\pi)$ and then taking
expectations, {\eqref{eqn:mixing_loss_l1}} gives
\[
\mathbb E\!\left[
    \left|
    \cvar_{\delta,\xi\sim F}[c(x_n^\pi,\xi)]
    -
    \cvar_{\delta,\xi\sim F}[c(x_\infty^\pi,\xi)]
    \right|
\right]
\longrightarrow0.
\]
Since $x_\infty^\pi$ is supported on population-optimal actions,
\[
    \cvar_{\delta,\xi\sim F}[c(x_\infty^\pi,\xi)]
    =
    \opt_\delta(F)
    \qquad\text{almost surely}.
\]
Therefore,
\[
    \mathrm{PS}_{\delta,n}(\pi,F)\longrightarrow0.
\]

\paragraph{Sample-risk premium.}
Applying the same $L^1$ continuity of $\cvar$ directly to {\eqref{eqn:mixing_loss_l1}} yields
\[
    \cvar_{\delta,x_n^\pi,\xi}[c(x_n^\pi,\xi)]
    \longrightarrow
    \cvar_{\delta,x_\infty^\pi,\xi}[c(x_\infty^\pi,\xi)].
\]
Together with the preceding convergence of the expected outcome-only
$\cvar$, this implies
\[
\mathrm{SR}_{\delta,n}(\pi,F)
\longrightarrow
\cvar_{\delta,x_\infty^\pi,\xi}[c(x_\infty^\pi,\xi)]
-
\mathbb E_{x_\infty^\pi}\!\left[
    \cvar_{\delta,\xi\sim F}[c(x_\infty^\pi,\xi)]
\right]
=: \mathrm{SR}_{\delta,\infty}(\pi,F).
\]
The limit is the sample-risk premium of the limiting action. The same comparison holds for any integrable action independent of the future outcome, so the strict inequality is recorded on its own.
\begin{lemma}
\label{lem:strict_sample_risk_premium}
Under the hypotheses of Proposition~\ref{prop:endpoints}, if the conditional
tails induced by $x^\pi$ are misaligned, then
\[
    \cvar_{\delta,x^\pi,\xi}
    \left[c(x^\pi,\xi)\right]
    >
    \mathbb{E}_{x^\pi}\!\left[
        \cvar_{\delta,\xi\sim F}
        \left[c(x^\pi,\xi)\right]
    \right].
\]
Consequently, the sample-risk premium of $\pi$ is strictly positive.
\end{lemma}
The proof is in Appendix~\ref{prf:strict_sample_risk_premium}. The condition
\[
\mathbb P\!\left(
t\in\mathcal T_\delta(x_\infty^\pi)
\right)<1
\qquad\text{for every }t\in\mathbb R
\]
is precisely the condition that the conditional tails induced by
$x_\infty^\pi$ are misaligned. Since $c(x_\infty^\pi,\xi)$ is integrable and
$x_\infty^\pi$ is independent of $\xi$,
Lemma~\ref{lem:strict_sample_risk_premium} gives
\[
    \mathrm{SR}_{\delta,\infty}(\pi,F)>0.
\]
Consequently,
\[
    \mathrm{PS}_{\delta,n}(\pi,F)=o(1),
    \qquad
    \mathrm{SR}_{\delta,n}(\pi,F)=\Theta(1).
\]

\subsection{Proof of Lemma~\ref{lem:strict_sample_risk_premium}}
\label{prf:strict_sample_risk_premium}

This is the contrapositive of part~1 of Proposition~\ref{prop:endpoints}. In the
notation of Appendix~\ref{prf:endpoints}, let $t^*$ attain the variational
representation of $\cvar_\delta[c(x^\pi,\xi)]$, which exists by integrability; as shown
there, $t^*$ also minimizes $G$. Misalignment gives
$\mathbb P\!\left(t^*\notin\mathcal T_\delta(x^\pi)\right)>0$, and on that event
$\phi_{x^\pi}(t^*)>\min_{s}\phi_{x^\pi}(s)$. Since the integrand defining $G$ is nonnegative,
$\SRcomp{\pi}{F}=G(t^*)>0$, which is the claimed strict inequality.

\subsection{Proof of Proposition~\ref{prop:separation_inconsistency}}\label{prf:separation_inconsistency}

Write $x_F^*$ for the unique optimal action. For any policy $\pi$ and any
$\varepsilon>0$, part~2 of Proposition~\ref{prop:endpoints} gives that the
integrand
\[
    \cvar_{\delta,\xi\sim F}[c(x^\pi,\xi)]
    -
    \opt_\delta(F)
\]
is nonnegative everywhere, and by definition of $\kappa_F(\varepsilon)$ it is
at least $\kappa_F(\varepsilon)$ on the event
$\{|x^\pi-x_F^*|\geq\varepsilon\}$. Hence
\[
    \mathrm{PS}_{\delta,n}(\pi,F)
    ~~\geq~~
    \kappa_F(\varepsilon)\cdot
    \mathbb P\!\left(|x^\pi-x_F^*|\geq\varepsilon\right).
\]
Under the hypothesis $\kappa_F(\varepsilon)>0$, if for some $\varepsilon>0$ and
$\eta>0$ one has
\[
\text{{\(\displaystyle \liminf_{n\to\infty}
    \mathbb P\!\left(|x^{\pi}-x_F^*|\geq\varepsilon\right)
    \geq\eta,\)}}
\]
the displayed inequality gives
\[
\text{{\(\displaystyle \liminf_{n\to\infty}
    \mathrm{PS}_{\delta,n}(\pi,F)
    ~~\geq~~
    \eta\,\kappa_F(\varepsilon)>0,\)}}
\]
hence {$\mathrm{PS}_{\delta,n}(\pi,F)=\Omega(1)$}.

 {For the consistency claim, suppose $\mathrm{PS}_{\delta,n}(\pi,F)\to0$.} Fix $\varepsilon>0$. Then
$\kappa_F(\varepsilon)>0$, and
\[
\text{{\(\displaystyle \mathbb P\!\left(|x^{\pi}-x_F^*|\geq\varepsilon\right)
    ~~\leq~~
    \frac{\mathrm{PS}_{\delta,n}(\pi,F)}{\kappa_F(\varepsilon)}
    ~~\longrightarrow~~
    0.\)}}
\]
Since $\varepsilon>0$ was arbitrary, {$x^{\pi}\to x_F^*$} in probability.

\section{Post-sample consistency of SAA}\label{apx:saa_post_sample_consistency}

\subsection{Proof of Proposition~\ref{prop:saa_post_sample_consistency}}

Write
\(\widehat Q_n(x)=\cvar_{\delta,\xi\sim\hat F_n}[c(x,\xi)]\), and write
\(\phi_n(x,t)\) for the empirical counterpart of the variational objective
\(\phiv{F}{x}{t}\). The argument is a uniform law of large numbers on the
variational objective, passed to the infimum over the threshold and then to
the minimizer. Boundedness localizes the threshold. Compactness and
almost-sure continuity give the uniform law.

\paragraph{The threshold lies in \([-M,M]\).}
The envelope \(\sup_{x\in\mathcal{X}}|c(x,\xi)|\leq M\) holds for
\(F\)-almost every \(\xi\), so \(c(x,\xi)\in[-M,M]\) for every
\(x\in\mathcal{X}\), outside a single null set. For \(t>M\),
\((c(x,\xi)-t)^+=0\), hence \(\phiv{F}{x}{t}=t>M=\phiv{F}{x}{M}\).
For \(t<-M\), \((c(x,\xi)-t)^+=c(x,\xi)-t\), and
\[
    \phiv{F}{x}{t}
    =
    \frac{\mathbb{E}[c(x,\xi)]}{1-\delta}
    -
    \frac{\delta}{1-\delta}\,t,
\]
which is strictly decreasing in \(t\). Thus \(\phiv{F}{x}{t}>\phiv{F}{x}{-M}\)
whenever \(t<-M\). The same identities hold pathwise for
\(\phi_n\), because almost every sample lies in the envelope. Therefore both
infima are attained inside \([-M,M]\):
\[
    \Qc{F}{x}
    =
    \inf_{t\in[-M,M]}\phiv{F}{x}{t},
    \qquad
    \widehat Q_n(x)
    =
    \inf_{t\in[-M,M]}\phi_n(x,t).
\]

\paragraph{Uniform convergence of the variational objective.}
Let \(\Theta=\mathcal{X}\times[-M,M]\), which is compact, and write a generic
point as \(\theta=(x,t)\). Define
\[
    f(\theta,\xi)
    =
    \bigl(c(x,\xi)-t\bigr)^+.
\]
Almost-sure continuity of \(x\mapsto c(x,\xi)\) makes \(\theta\mapsto f(\theta,\xi)\)
continuous for \(F\)-almost every \(\xi\), and \(0\leq f\leq 2M\). For
\(\eta>0\) let
\[
    \omega(\xi,\eta)
    =
    \sup\bigl\{
        |f(\theta,\xi)-f(\theta',\xi)|
        :
        \theta,\theta'\in\Theta,\;
        |\theta-\theta'|\leq\eta
    \bigr\},
\]
where the supremum may be taken over a countable dense set of pairs and then
agrees with the supremum over all pairs, on the almost-sure set where
\(f(\cdot,\xi)\) is continuous. Thus \(\omega(\xi,\eta)\) is measurable,
\(0\leq\omega(\xi,\eta)\leq 4M\), and \(\omega(\xi,\eta)\to0\) for
\(F\)-almost every \(\xi\) as \(\eta\to0\), by uniform continuity on the
compact set \(\Theta\). Dominated convergence gives
\(\mathbb{E}[\omega(\xi,\eta)]\to0\).

Fix \(\eta>0\) and cover \(\Theta\) by finitely many balls of radius \(\eta\)
with centers \(\theta_1,\ldots,\theta_J\). For an arbitrary \(\theta\), choose
a center \(\tau(\theta)\) within distance \(\eta\). Splitting the empirical
average at that center and using
\(|f(\theta,\xi)-f(\tau(\theta),\xi)|\leq\omega(\xi,\eta)\) gives
\begin{align*}
    \bigl|\phi_n(x,t)-\phiv{F}{x}{t}\bigr|
    &\leq
    \frac{1}{1-\delta}\Biggl(
        \frac1n\sum_{i=1}^n \omega(\xi_i,\eta)
        +
        \max_{1\leq j\leq J}
        \Biggl|
            \frac1n\sum_{i=1}^n f(\theta_j,\xi_i)
            -
            \mathbb{E}[f(\theta_j,\xi)]
        \Biggr|
        +
        \mathbb{E}[\omega(\xi,\eta)]
    \Biggr).
\end{align*}
The right-hand side does not depend on \(\theta\). The strong law sends the
average of \(\omega(\cdot,\eta)\) to its expectation and each of the finitely
many centered averages to zero. Hence, almost surely,
\[
    \limsup_{n\to\infty}
    \sup_{\theta\in\Theta}
    \bigl|\phi_n(x,t)-\phiv{F}{x}{t}\bigr|
    \leq
    \frac{2}{1-\delta}\,\mathbb{E}[\omega(\xi,\eta)].
\]
The null set can be taken independent of \(\eta=1/k\) for \(k\in\mathbb{N}\).
The left-hand side does not depend on \(k\), and the right-hand side tends to
zero with \(k\), so the limsup is zero. That is,
\[
    \sup_{x\in\mathcal{X},\,t\in[-M,M]}
    \bigl|\phi_n(x,t)-\phiv{F}{x}{t}\bigr|
    \longrightarrow 0
    \qquad\text{almost surely}.
\]
An infimum over \(t\) moves by at most the uniform distance of the two
objectives, so
\[
    \sup_{x\in\mathcal{X}}
    \bigl|\widehat Q_n(x)-\Qc{F}{x}\bigr|
    \longrightarrow 0
    \qquad\text{almost surely}.
\]
{Whenever $|(x,t)-(y,s)|\le\eta$,}
\[
\text{{$\displaystyle \bigl|\phiv{F}{x}{t}-\phiv{F}{y}{s}\bigr|\le |t-s|+\frac{\mathbb{E}[\omega(\xi,\eta)]}{1-\delta}.$}}
\]
{The right-hand side tends to zero with $\eta$, so $\phi_F$ is jointly continuous on $\mathcal X\times[-M,M]$. Minimizing over the compact threshold interval gives continuity of $\Qc{F}{\cdot}$ on $\mathcal X$.}
A continuous function on a nonempty compact set attains its minimum, so
\(\mathcal{X}^*_F\) is nonempty and compact. On the same almost-sure set,
each map \(x\mapsto c(x,\xi_i)\) is continuous, so \(\widehat Q_n\) is
continuous and its argmin is nonempty. The smallest minimizer is a measurable
selection; the argument below applies to any measurable selection, and
\(x^{\mathrm{SAA}}\) denotes the one fixed in the text.

\enlargethispage{\baselineskip}
\paragraph{The post-sample gap.}
Let \(x_F^*\in\mathcal{X}^*_F\). Optimality of \(x^{\mathrm{SAA}}\) for \(\widehat Q_n\)
gives \(\widehat Q_n(x^{\mathrm{SAA}})\leq\widehat Q_n(x_F^*)\). Adding and subtracting
\(\widehat Q_n(x^{\mathrm{SAA}})\) and \(\widehat Q_n(x_F^*)\) and dropping that
nonpositive difference,
{\predisplaypenalty=0
\begin{align*}
    0
    &\leq
    \Qc{F}{x^{\mathrm{SAA}}}-\opt_\delta(F) \\
    &=
    \bigl(\Qc{F}{x^{\mathrm{SAA}}}-\widehat Q_n(x^{\mathrm{SAA}})\bigr)
    +
    \bigl(\widehat Q_n(x^{\mathrm{SAA}})-\widehat Q_n(x_F^*)\bigr)
    +
    \bigl(\widehat Q_n(x_F^*)-\Qc{F}{x_F^*}\bigr) \\
    &\leq
    2\sup_{x\in\mathcal{X}}
    \bigl|\widehat Q_n(x)-\Qc{F}{x}\bigr|.
\end{align*}
}
The right-hand side tends to zero almost surely, so the random post-sample
gap vanishes almost surely. The gap is at most \(2M\). Bounded convergence
gives
\[
    \mathrm{PS}_{\delta,n}(\pi^{\mathrm{SAA}},F)
    =
    \mathbb{E}\bigl[\Qc{F}{x^{\mathrm{SAA}}}\bigr]
    -
    \opt_\delta(F)
    \longrightarrow 0.
\]

\section{\texorpdfstring{{SAA: asymptotics and finite-sample calculations}}{SAA: asymptotics and finite-sample calculations}}\label{prf:saa_concentration_rate}

{For the concentration proof, write \(x^*=x_F^*\), \(q^*=q_F(x^*)\), and \(Q''=Q_F''(x^*)\). Let \(a_F\) be the cutoff slope defined in~\eqref{eqn:saa_variance}. Write \(\Delta_n=x^{\mathrm{SAA}}-x^*\), and \(\widehat Q_n(x)=\cvar_{\delta,\xi\sim\widehat F_n}[c(x,\xi)]\). Let \(M\) be the uniform loss bound in Proposition~\ref{prop:saa_post_sample_consistency}. For an integrable function \(f\), use \(Pf=\mathbb E[f(\xi)]\) and \(P_nf=n^{-1}\sum_{i=1}^n f(\xi_i)\).}

\begin{lemma}[Quadratic gap]\label{lem:saa_quadratic_gap}
{Let \(x^*\) be an interior minimizer of \(Q_F\), and suppose \(Q_F\) is twice differentiable there with \(0<Q''<\infty\). Then there are a neighborhood \(\mathcal N\) of \(x^*\) and constants \(0<c\le C<\infty\) such that, for every \(x\in\mathcal N\),}
{\[
c\,|x-x^*|^2
~~\leq~~
\Qc{F}{x}-\Qc{F}{x^*}
~~\leq~~
C\,|x-x^*|^2.
\]}
\end{lemma}

\begin{proof}
{By hypothesis, \(x^*\) is an interior minimizer of \(\Qc{F}{\cdot}\) and \(0<Q''<\infty\). The first derivative vanishes at an interior minimizer, and Taylor's theorem with Peano remainder gives}
{\[
\Qc{F}{x}-\Qc{F}{x^*}
=
\tfrac12 Q''\,(x-x^*)^2
+
o\bigl((x-x^*)^2\bigr)
\]}
as \(x\to \text{{\(x^*\)}}\). Positivity and finiteness of the second derivative produce the neighborhood and the constants.
\end{proof}

\begin{lemma}[Parametric rate]\label{lem:saa_rate}
{Suppose the hypotheses of Proposition~\ref{prop:saa_concentration_rate}. With \(0<\sigma^2<\infty\) as in~\eqref{eqn:saa_variance},
\[
\sqrt n\,\Delta_n\Rightarrow\mathcal N(0,\sigma^2),
\qquad \{n\Delta_n^2\}_{n\ge1}\text{ is uniformly integrable}.
\]
Consequently, \(\Delta_n=\Theta_{\mathbb P}(n^{-1/2})\).}
\end{lemma}

\begin{proof}
{We first identify a centered score with positive variance and bound the local empirical process. This bound localizes the joint SAA minimizer and controls its tails. We obtain asymptotic linearity from the joint M-estimation theorem in the smooth regime and from a direct comparison in the atomic regime, then conclude uniform integrability.}

Well separation and the post-sample consistency of SAA give \(\text{{\(\Delta_n\)}}\to 0\) in probability.

\paragraph{The slope is a nondegenerate average.}

{We identify the score governing the action fluctuation.} By the hypothesis of the proposition the action derivative exists at \(\text{{\(x^*\)}}\) for almost every outcome, and equation~\eqref{eqn:marginal_loss_bound} forces \(|\partial_x c(\text{{\(x^*\)}},\xi)|\le \mu(\xi)\). 
{Use \(\psi=\psi_F\), with \(\psi_F\) as in~\eqref{eqn:saa_variance}, and set \(S_n=n^{-1}\sum_{i=1}^n\psi(\xi_i)\) and \(v=\operatorname{Var}(\psi(\xi))\). The bound by a constant multiple of \(1+\mu\) puts \(\psi\) in \(L^2\). We verify below that it is centered and \(v>0\); centering then gives \(\mathbb E[nS_n^2]=v\) for every \(n\).}

{In the smooth regime, the score in equation~\eqref{eqn:saa_variance} can be written as}
{\[
\psi(\xi)
=
\frac{1}{(1-\delta)}\,
\Bigl(
\partial_x c(x^*,\xi)\,
\mathbf{1}_{\{c(x^*,\xi)> q^*\}}
+
a_F\bigl(\mathbf{1}_{\{c(x^*,\xi)\le q^*\}}-\delta\bigr)
\Bigr).
\]}
{There is no atom at the cutoff, so \(\mathbb P(c(x^*,\xi)>q^*)=(1-\delta)\) and the threshold-score term has mean zero. The local envelope permits differentiation under the expectation at a fixed threshold. Since \(\phi_F(x,q^*)-Q_F(x)\) is nonnegative and vanishes at the interior point \(x^*\), its derivative there is zero. Thus
\[
\mathbb E\psi
=\frac{\mathbb E[\partial_xc(x^*,\xi)\mathbf1_{\{c(x^*,\xi)>q^*\}}]}{(1-\delta)}
=\partial_x\phi_F(x^*,q^*)
=Q_F'(x^*)=0.
\]
On \(\{c(x^*,\xi)<q^*\}\), which has probability \(\delta>0\), the score equals \(a_F\neq0\). Its variance \(v\) is therefore positive.}

In the atomic regime the cutoff sits on the atom, so the score is the derivative of the variational objective along the path that moves the action and the atom level together. 

{Here \(a_F\) is the derivative of the atom level at \(x^*\).} The criterion along that path is \(\ell(x)+(c(x,\xi)-\ell(x))^+/(1-\delta)\). The isolation limits~\eqref{eqn:atom_isolation} force the action derivative on the atom to equal \(a_F\). 

{If, for some \(\varepsilon>0\), the event \(E_\varepsilon:=\{c(x^*,\xi)=q^*,\ \partial_xc(x^*,\xi)-a_F>\varepsilon\}\) has positive probability, differentiability implies
\[
\mathbf1_{E_\varepsilon}\mathbf1_{\{c(x^*+h,\xi)>\ell(x^*+h)+\varepsilon h/2\}}
\longrightarrow\mathbf1_{E_\varepsilon}
\qquad(h\downarrow0).
\]
By continuity, the original strict upper-tail outcomes also remain above this boundary asymptotically. Dominated convergence on these two disjoint events gives a limit inferior of the upper-tail probability at least \(\mathbb P(c(x^*,\xi)>q^*)+\mathbb P(E_\varepsilon)\), contradicting the first isolation limit along \(u=\varepsilon h/2\). If the derivative difference is below \(-\varepsilon\) on a positive-probability subset of the atom, the analogous argument removes that mass from the event in the second isolation limit, while outcomes initially below the atom remain below the moving boundary asymptotically. This contradicts the second limit. Hence \(\partial_xc=a_F\) almost surely on the oracle atom.}

Outcomes off the atom may still cross the moving level. 
{For each increment \(h\), let \(C_h\) be the event that a loss initially off the oracle atom crosses the moving level between \(x^*\) and \(x^*+h\). Off the oracle atom, \(c(x^*,\xi)\neq\ell(x^*)\); continuity of \(c(\cdot,\xi)\) and \(\ell\) makes the crossing indicator tend to zero. Dominated convergence therefore gives \(\mathbb P(C_h)\to0\). The difference quotients are dominated by a constant multiple of \(1+\mu\), so dominated convergence controls both these crossings and the ordinary action-derivative remainder. On the oracle atom, derivative alignment instead gives \(c(x^*+h,\xi)-\ell(x^*+h)=o(|h|)\) almost surely. Thus the moving-path remainder is \(o(|h|)\) in \(L^2\).} 

Where the positive part does not switch, the derivative equals \(a_F\) on \(\{c\le \text{{\(q^*\)}}\}\) and \((\partial_x c-\delta a_F)/(1-\delta)\) on \(\{c>\text{{\(q^*\)}}\}\). The score is therefore
{\[
\psi(\xi)
~~=~~
a_F
+
\frac{\partial_x c(x^*,\xi)-a_F}{(1-\delta)}\,
\mathbf{1}_{\{c(x^*,\xi)>q^*\}}.
\]}
It equals \(a_F\) on the atom. Since \(a_F\neq 0\) and the atom has positive probability, \(\psi\) is not almost surely zero. Dominated convergence under~\eqref{eqn:marginal_loss_bound} passes the derivative under the expectation, so \(\psi\) is centered because \(Q_F'(\text{{\(x^*\)}})=0\), and a centered square-integrable random variable that is not almost surely zero has variance \(v>0\). \(S_n\) is its sample average.

In both regimes the central limit theorem gives \(\sqrt{n}\,S_n\Rightarrow\mathcal{N}(0,v)\) with \(v>0\). In particular \(S_n=O_{\mathbb{P}}(n^{-1/2})\) and \(S_n\) is not \(o_{\mathbb{P}}(n^{-1/2})\).

\paragraph{\texorpdfstring{{A local empirical-process bound.}}{A local empirical-process bound.}}

{Write \(f_{x,t}(\xi)=t+(c(x,\xi)-t)^+/(1-\delta)\). The local bound~\eqref{eqn:marginal_loss_bound} gives an envelope \(\mu_f(\xi)=C(1+\mu(\xi))\) with \(\mathbb E[\mu_f^{2+\eta}]<\infty\), such that, for all actions \(x,y\) in the neighborhood of~\eqref{eqn:marginal_loss_bound} and all thresholds \(t,t'\),
\[
|f_{x,t}-f_{y,t'}|\le \mu_f(\xi)(|x-y|+|t-t'|).
\]
Let \(\mathcal B_\lambda\) be the square of radius \(\lambda\) centered at \((x^*,q^*)\), and define
\[
G_n(x,t)=\sqrt n(P_n-P)(f_{x,t}-f_{x^*,q^*}).
\]}

{\emph{Local maximal bound.} For all sufficiently small \(\lambda>0\) and all \(n\),}

\begin{equation}\label{eqn:saa_local_maximal}
\text{{\(\displaystyle \mathbb E\!\left[\sup_{\mathcal B_\lambda}|G_n|^{2+\eta}\right]
\le C_\eta\lambda^{2+\eta}\mathbb E[\mu_f^{2+\eta}],\)}}
\end{equation}

{where \(C_\eta\) is independent of \(n\) and \(\lambda\).}

{The local Lipschitz envelope makes \(G_n\) almost surely continuous, so its supremum equals the supremum over the countable union of dyadic grids and is measurable. To prove the bound, symmetrize \(G_n\) and use dyadic grids with spacing \(d_k=\lambda2^{-k}\). Connect each point to \((x^*,q^*)\) through its successive grid projections. Conditionally on the sample, each Rademacher increment across a grid edge is sub-Gaussian with variance proxy at most \(C(P_n\mu_f^2)d_k^2\). There are at most \(C4^k\) edges at level \(k\), so the conditional \(L^{2+\eta}\) norm of their maximum is at most \(C_\eta d_k\sqrt{k+1}(P_n\mu_f^2)^{1/2}\). The Lipschitz bound makes the terminal grid remainder vanish for each fixed \(n\). Minkowski's inequality, moment symmetrization, and the convergence of \(\sum_k2^{-k}\sqrt{k+1}\) therefore bound the left-hand side of~\eqref{eqn:saa_local_maximal} by \(C_\eta\lambda^{2+\eta}\mathbb E[(P_n\mu_f^2)^{(2+\eta)/2}]\). Jensen's inequality gives \(\mathbb E[(P_n\mu_f^2)^{(2+\eta)/2}]\le\mathbb E[\mu_f^{2+\eta}]\), proving the claim.}

\paragraph{\texorpdfstring{{Localization at the root-$n$ scale.}}{Localization at the root-n scale.}}
{Bounded loss confines every minimizing threshold to \([-M,M]\), as shown in Appendix~\ref{apx:saa_post_sample_consistency}.}
Let \(t_n(x)\) minimize \(t\mapsto P_n f_{x,t}\) and write \(t_n=t_n(x^{\mathrm{SAA}})\). Since \(\widehat Q_n(x^{\mathrm{SAA}})\le\widehat Q_n(\text{{\(x^*\)}})\le P_n f_{\text{{\(x^*\)}},\text{{\(q^*\)}}}\),
{\[
P f_{x^{\mathrm{SAA}},t_n}-P f_{x^*,q^*}
\leq
n^{-1/2}\,\bigl|G_n(x^{\mathrm{SAA}},t_n)\bigr|.
\]}
{The left-hand side is the quadratic action gap plus the threshold excess above \(q_F(x^{\mathrm{SAA}})\). The minimizing threshold at \(x^*\) is unique: the local quadratic threshold gap in Lemma~\ref{lem:smooth_exponents}, or strict straddling in Lemma~\ref{lem:atom_exponents}, together with convexity in the threshold, gives the singleton \(\{q^*\}\). Since the oracle action is unique, the joint population minimizer is \((x^*,q^*)\). Continuity on the compact set \(\mathcal X\times[-M,M]\) and the uniform law proved in Appendix~\ref{apx:saa_post_sample_consistency} then give joint consistency of \((x^{\mathrm{SAA}},t_n)\). Choose a sufficiently small fixed neighborhood. There \(|q_F(x)-q^*|\le(|a_F|+1)|x-x^*|\), and the threshold excess is at least a positive multiple of \(|t-q_F(x)|^\gamma\), with \(\gamma=2\) in the smooth regime and \(\gamma=1\) in the atomic regime. On this bounded neighborhood the linear atomic bound also dominates a quadratic bound. Consequently the joint population gap is at least a positive multiple of
\[
|x-x^*|^2+|t-q^*|^2
\]
in both regimes. Write}
{\[
R_n=\max\bigl(|\Delta_n|,\,|t_n-q^*|/(|a_F|+1)\bigr)
\]}
{and \(\lambda_j=2^j n^{-1/2}\). On an annulus \(\{\lambda_{j-1}<R_n\le\lambda_j\}\) inside a sufficiently small fixed neighborhood, the basic inequality requires
\[
\sup_{\mathcal B_{C\lambda_j}}|G_n|\ge c\sqrt n\,\lambda_{j-1}^2.
\]
Bound~\eqref{eqn:saa_local_maximal} and Markov's inequality make its probability at most \(C_\eta2^{-j(2+\eta)}\). Choose a fixed radius \(r\) small enough that the bounds hold on all the squares needed to cover \(\{R_n\le r\}\). Summing the annuli, including the last annulus intersected with this set, gives
\[
\mathbb P(R_n>2^J n^{-1/2},\,R_n\le r)\le C_\eta2^{-J(2+\eta)}.
\]
Joint consistency controls \(\{R_n>r\}\). Thus \(\Delta_n=O_{\mathbb P}(n^{-1/2})\) and \(t_n-q^*=O_{\mathbb P}(n^{-1/2})\).}

\paragraph{\texorpdfstring{{Asymptotic linearity.}}{Asymptotic linearity.}}
{Joint localization and \(S_n=O_{\mathbb P}(n^{-1/2})\) place the minimizer and the candidate \(h_n=-S_n/Q''\) in a fixed root-\(n\) window with arbitrarily high probability. We prove the expansions on each window \(|h|,|u|\le R_0n^{-1/2}\), with \(x=x^*+h\) and \(t=q^*+u\); the joint integrand is locally Lipschitz in these coordinates.}

The population half expands in both coordinates together. Lemma~\ref{lem:saa_quadratic_gap} gives \(\Qc{F}{\text{{\(x^*\)}}+h}-\Qc{F}{\text{{\(x^*\)}}}=\tfrac12 \text{{\(Q''\)}}h^2+o(h^2)\). In the smooth regime the proof of Lemma~\ref{lem:smooth_exponents} gives threshold curvature {\(g/(2(1-\delta))\)}, with \(g=g_{\text{{\(x^*\)}}}(\text{{\(q^*\)}})\), and \(u(h):=q_F(\text{{\(x^*\)}}+h)-\text{{\(q^*\)}}=a_F h+o(|h|)\). Uniformly for \(|h|\) and \(|u|\) of order \(n^{-1/2}\),
{\[
P f_{x^*+h,\,q^*+u}
=
\Qc{F}{x^*}
+
\tfrac12 Q''\,h^2
+
\frac{g}{2(1-\delta)}(u-u(h))^2
+
o(n^{-1}).
\]}
In the atomic regime Lemma~\ref{lem:atom_exponents} gives \(q_F=\ell\) near the oracle and a linear threshold excess of slope at least half of \(\min\{s_+,s_-\}>0\), with the slopes of~\eqref{eqn:atom_slopes}. The quadratic term in \(h\) is again the lemma above.

{In the smooth regime, apply the minimization form of \citet[Theorem~5.23]{VanDerVaart1998} to the joint criterion $f_{x,t}$. Joint consistency was established above. There is no atom at $q^*$, so the criterion is almost surely differentiable at $(x^*,q^*)$; the local Lipschitz envelope is a constant multiple of $1+\mu$ and lies in $L^2$. The population expansion above has positive definite matrix
\[
\mathsf H=\begin{pmatrix}Q''+b a_F^2&-b a_F\\-b a_F&b\end{pmatrix},
\qquad b=g/(1-\delta),\qquad \det\mathsf H=Q''b>0.
\]
Exact SAA minimization satisfies the theorem's optimization-error condition. With
\[
\dot f(\xi)=\begin{pmatrix}
\partial_xc(x^*,\xi)\mathbf1_{\{c(x^*,\xi)>q^*\}}/(1-\delta)\\
1-\mathbf1_{\{c(x^*,\xi)>q^*\}}/(1-\delta)
\end{pmatrix},
\]
the theorem gives the joint linear representation with influence $-\mathsf H^{-1}\dot f$. Its action coordinate is $-\psi_F/Q''$, since the first row of $\mathsf H^{-1}$ is $(1,a_F)/Q''$. Thus
\[
\sqrt n\,(x^{\mathrm{SAA}}-x^*)=-\frac{\sqrt n\,S_n}{Q''}+o_{\mathbb P}(1).
\]
}

{\emph{Stochastic remainder bound.} Suppose \(r_{0,0}=0\), the class \(r_{h,w}\) is locally Lipschitz in \((h,w)\) with envelope \(C\mu_f\), and \(\sup_{|h|,|w|\le\lambda}\|r_{h,w}\|_2=o(\lambda)\). For every fixed \(0<R_0<\infty\),}

\begin{equation}\label{eqn:saa_stochastic_remainder}
\text{{\(\displaystyle \sup_{|h|,|w|\le R_0/\sqrt n}|(P_n-P)r_{h,w}|=o_{\mathbb P}(n^{-1}).\)}}
\end{equation}

{To see this, put \(\lambda=R_0/\sqrt n\) and normalize the class as \(r_{\lambda\beta_1,\lambda\beta_2}/\lambda\), with \(\beta\in[-1,1]^2\). On any fixed finite grid, the expected maximum of its centered empirical process \(\sqrt n(P_n-P)\) tends to zero by the \(L^2\) assumption. After symmetrization, the fine-grid tail has expected supremum at most
\[
C\|\mu_f\|_2\sum_{k\ge k_0}2^{-k}\sqrt{k+1},
\]
uniformly in \(n\), by the same chaining argument. First let \(n\) grow on the fixed grid, then let \(k_0\) grow. Rescaling gives~\eqref{eqn:saa_stochastic_remainder}.}

{In the atomic regime use relative threshold coordinates:
\[
\widetilde f_{h,w}=f_{x^*+h,\,\ell(x^*+h)+w},\qquad
B=1-\frac{\mathbf1_{\{c(x^*,\xi)>q^*\}}}{(1-\delta)},\qquad
D=\frac{\mathbf1_{\{c(x^*,\xi)=q^*\}}}{(1-\delta)}.
\]
Retain the threshold kink. The integrand expansion is
\[
\widetilde f_{h,w}-\widetilde f_{0,0}
=h\psi+wB+(-w)^+D+r_{h,w}.
\]
Off the oracle atom, continuity makes the sign of the hinge constant on a sufficiently small square for each outcome. On the atom, derivative alignment gives \(c(x^*+h,\xi)-\ell(x^*+h)=o(|h|)\); the Lipschitz property of the positive part controls the remainder uniformly in \(w\). Differentiability therefore yields
\[
\sup_{|h|,|w|\le\lambda}\frac{|r_{h,w}(\xi)|}{\lambda}\longrightarrow0
\quad\text{almost surely}.
\]
This supremum is bounded by a constant multiple of \(1+\mu\). Dominated convergence therefore gives
\[
\mathbb E\!\left[\sup_{|h|,|w|\le\lambda}|r_{h,w}|^2\right]=o(\lambda^2),
\qquad
\sup_{|h|,|w|\le\lambda}\|r_{h,w}\|_2=o(\lambda).
\]
The second bound follows from the first. Since \(\ell\) is continuously differentiable, the remainder also has a Lipschitz envelope \(C\mu_f\). Bound~\eqref{eqn:saa_stochastic_remainder} therefore applies.}

{Keep the population threshold excess exact:
\[
T(h,w)=P(\widetilde f_{h,w}-\widetilde f_{h,0})
\ge c_{\mathrm{at}}|w|
\]
locally, for some \(c_{\mathrm{at}}>0\), by Lemma~\ref{lem:atom_exponents}. With the centered averages \(B_n=(P_n-P)B\) and \(D_n=(P_n-P)D\), the empirical expansion is
\[
P_n\widetilde f_{h,w}-P_n\widetilde f_{0,0}
=\frac{Q''}{2}h^2+hS_n+T(h,w)+wB_n+(-w)^+D_n
+o_{\mathbb P}(n^{-1}).
\]
The indicators are bounded, so \(|B_n|+|D_n|=O_{\mathbb P}(n^{-1/2})\). With probability tending to one, the threshold terms are therefore at least \(c_{\mathrm{at}}|w|/2\). Joint localization and local Lipschitz continuity of \(\ell\) give \(w_n=t_n-\ell(x^{\mathrm{SAA}})=O_{\mathbb P}(n^{-1/2})\). Compare this joint minimizer with the feasible candidate \((h_n,0)=(-S_n/Q'',0)\). Optimality and completion of the action square give
\[
\frac{Q''}{2}(\Delta_n-h_n)^2+\frac{c_{\mathrm{at}}}{2}|w_n|
\le o_{\mathbb P}(n^{-1}).
\]
Thus \(\Delta_n-h_n=o_{\mathbb P}(n^{-1/2})\) and \(w_n=o_{\mathbb P}(n^{-1})\).}

{ The atomic comparison holds on each fixed root-$n$ window; tightness lets its radius increase after taking $n$ to infinity. Together with the smooth linear representation, this gives in both regimes}
{\[
\bigl(\Delta_n+S_n/Q''\bigr)^2
=
o_{\mathbb{P}}(n^{-1}).
\]}
Hence
{\[
\sqrt{n}\,\Delta_n
~~=~~
-\frac{\sqrt{n}\,S_n}{Q''}
+
o_{\mathbb{P}}(1).
\]}
{The first term converges in distribution to \(\mathcal N(0,v/(Q'')^2)=\mathcal N(0,\sigma^2)\), with \(0<\sigma^2<\infty\) as in~\eqref{eqn:saa_variance}. Since this normal distribution has no atom at zero, \(\Delta_n=\Theta_{\mathbb P}(n^{-1/2})\).}

\paragraph{Uniform integrability.}
{Use the annuli from the localization step. On the \(j\)-th annulus inside \(\{R_n\le r\}\), \(n\Delta_n^2\le nR_n^2\le4^j\), while its probability is at most \(C_\eta2^{-j(2+\eta)}\). Thus the tail contribution from \(j\ge J\) is bounded, uniformly in \(n\), by
\[
C_\eta\sum_{j\ge J}2^{-j\eta}\longrightarrow0.
\]
For the escape event, set
\[
W_n=\sup_{x\in\mathcal X,\,|t|\le M}|(P_n-P)f_{x,t}|.
\]
The bounded uniform law from Appendix~\ref{apx:saa_post_sample_consistency} gives \(\mathbb E[W_n]\to0\). Replacing one observation changes \(W_n\) by at most \(2M/(n(1-\delta))\). Bounded differences therefore give an exponentially small probability that \(W_n\) exceeds any fixed positive constant, for all sufficiently large \(n\). On the compact joint parameter set the continuous population objective has the unique minimizing pair \((x^*,q^*)\), so \(\{R_n>r\}\) forces such a fixed deviation by the basic inequality. Since \(\mathcal X\) is bounded,
\(\mathbb E[n\Delta_n^2\mathbf1_{\{R_n>r\}}]\to0\).
Combining this with the annular tail bound proves uniform integrability of \(\{n\Delta_n^2\}\).}
\end{proof}

\subsection{Proof of Proposition~\ref{prop:saa_concentration_rate}\texorpdfstring{{ and the regret rates}}{ and the regret rates}}

{Lemma~\ref{lem:saa_rate} gives the normal limit, uniform integrability, and the stated rate in probability.}

{The normal limit and uniform integrability give~\eqref{eqn:concentration} with \(r_n=n^{-1/2}\). For a fixed action neighborhood \(A_n=\{|\Delta_n|\le r\}\), uniform integrability implies
\[
n\mathbb P(A_n^c)\le r^{-2}\mathbb E[n\Delta_n^2\mathbf1_{A_n^c}]\longrightarrow0.
\]
Bounded loss gives \(|Q_F(x)-Q_F(x^*)|\le2M\) and \(|q_F(x)|\le M\). The escape remainders in Lemma~\ref{lem:orders} are therefore \(O(\mathbb P(A_n^c))=o(n^{-1})\). The proofs of Theorems~\ref{prop:asymptotic_characterization} and~\ref{cor:atom_category} apply with these bounds replacing the global Lipschitz estimates for those remainders. They give \(\mathrm{PS},\mathrm{SR}=\Theta(n^{-1})\) in the smooth regime, and \(\mathrm{PS}=\Theta(n^{-1})\), \(\mathrm{SR}=\Theta(n^{-1/2})\) in the atomic regime. The same replacement applies to the coefficient limits in Propositions~\ref{prop:smooth_brackets} and~\ref{prop:atom_brackets} and hence to Corollary~\ref{cor:rho_limit} in the smooth regime.}

\subsection{\texorpdfstring{{Finite-sample computation for the smooth examples}}{Finite-sample computation for the smooth examples}}\label{apx:saa_smooth_finite}

{Table~\ref{tab:saa_smooth_finite} uses the truncated distributions of an exponential outcome conditional on $\xi\le12$ and a lognormal outcome conditional on $\xi\le2000$, with the parameters in Section~\ref{sec:saa_finite}. There is no mass at the truncation bound. At \(\delta=0.95\), conditioning the lognormal distribution on \(\xi\le2000\) changes \(K\) from approximately \(428.07\) for the untruncated distribution in Table~\ref{tab:K_delta} to \(418.42\) in Table~\ref{tab:saa_smooth_finite}. For squared loss on $[0,B]$, bounded outcomes ensure bounded loss and a bounded action-Lipschitz envelope. The objective $Q_F$ is strongly convex, since subtracting $x^2/2$ from each loss leaves an affine function of $x$ whose CVaR is convex. The computed oracles are $x^*\approx1.97151$ and $x^*\approx30.81134$, respectively. Their critical outcome pairs $x^*\pm\sqrt{2q^*}$ are approximately $(0.03048,3.91254)$ and $(0.04188,61.58080)$, both strictly inside the respective supports, with positive unequal densities. Proposition~\ref{prop:K_squared} gives the local curvature and nonzero cutoff slope; compactness and the smooth squared loss give the remaining hypotheses of Proposition~\ref{prop:saa_concentration_rate}.}

{For each outcome distribution and $n\in\{100,400,1600\}$, generate $R=12{,}000$ independent training samples and compute their SAA actions $x_1^{\mathrm{SAA}},\ldots,x_R^{\mathrm{SAA}}$. With $1-\delta=0.05$, $(1-\delta)n$ is an integer, so $m=(1-\delta)n$ in the SAA formula and the empirical CVaR is the average of these $m$ largest sample losses. For a sorted sample, these losses come from the two ends of the sample, so the objective is piecewise quadratic in $x$. On each piece, its minimizer is the mean of the selected tail outcomes projected onto that piece. Minimizing these quadratic pieces gives the SAA action. All actions are retained individually for risk evaluation. The accompanying computational supplement includes \texttt{smooth\_saa.py}, whose \texttt{mc} mode reproduces these calculations. It initializes one random generator per outcome distribution, with seeds \(20261025\) for the exponential and \(20261026\) for the lognormal, and generates the three rows in increasing order of \(n\).}

{Population evaluation uses analytic truncated moments. For $t\ge0$, put $d=\sqrt{2t}$ and $m_j(x,d)=\mathbb E[\xi^j\mathbf1_{\{|\xi-x|>d\}}]$, $j=0,1,2$. Then
\[
\phi_F(x,t)=t+\frac{m_2(x,d)-2xm_1(x,d)+(x^2-2t)m_0(x,d)}{2(1-\delta)}.
\]
The conditional cutoff solves $\mathbb P(|\xi-x|\le\sqrt{2q_F(x)})=\delta$, and $Q_F(x)=\phi_F(x,q_F(x))$. The common threshold $\widehat t$ minimizes $R^{-1}\sum_{r=1}^R\phi_F(x_r^{\mathrm{SAA}},t)$ and solves
\[
\frac1R\sum_{r=1}^R\mathbb P_\xi\bigl(|\xi-x_r^{\mathrm{SAA}}|\le\sqrt{2\widehat t}\bigr)=\delta,
\]
where each probability is under the truncated outcome distribution with $x_r^{\mathrm{SAA}}$ held fixed. With $Q^*=Q_F(x^*)$, define
\[
a_r=Q_F(x_r^{\mathrm{SAA}})-Q^*,\qquad
b_r=\phi_F(x_r^{\mathrm{SAA}},\widehat t)-Q_F(x_r^{\mathrm{SAA}}).
\]
The Monte Carlo estimates are $\widehat{\mathrm{PS}}=R^{-1}\sum_r a_r$, $\widehat{\mathrm{SR}}=R^{-1}\sum_r b_r$, and $\widehat\rho=\widehat{\mathrm{SR}}/\widehat{\mathrm{PS}}$.}

{The standard errors quantify only variation across the simulated training samples. The envelope derivative at the minimizing common threshold and the delta method give the first-order ratio influence values
\[
w_r=\frac{b_r-\widehat\rho\,a_r}{\widehat{\mathrm{PS}}},\qquad
\sum_{r=1}^R w_r=0.
\]
We report the approximate Monte Carlo standard errors
\[
\operatorname{se}_{\mathrm{MC}}(\widehat\rho)
=\left(\frac{\sum_{r=1}^R w_r^2}{R(R-1)}\right)^{1/2},\qquad
\operatorname{se}_{\mathrm{MC}}(\mathrm{share})
=\frac{\operatorname{se}_{\mathrm{MC}}(\widehat\rho)}{(1+\widehat\rho)^2}.
\]
Share standard errors are multiplied by $100$ in the table to express them in percentage points.}

{Minimizing over the common threshold using the same replications makes \(\widehat{\mathrm{SR}}\) downward biased. The supplement's \texttt{bias\_diagnostic.py} estimates the leading term of a second-order expansion of this optimization bias. Across the six rows, its estimated magnitude is \(0.3\%\)--\(1.3\%\) of the Monte Carlo standard error of \(\widehat{\mathrm{SR}}\).}

\subsection{\texorpdfstring{{Finite-sample calculations for the uniform newsvendor}}{Finite-sample uniform newsvendor calculations}}\label{apx:saa_finite}

{For $\xi\sim U[0,1]$, $c(x,\xi)=x/2-\min(x,\xi)$, and $1-\delta=0.1$, direct integration gives
\[
Q_F(x)=\begin{cases}
5x^2-x/2,&0\le x\le(1-\delta),\\
x/2-(1-\delta)/2,&(1-\delta)\le x\le1.
\end{cases}
\]
Hence $x_F^*=0.05$, $Q_F(x_F^*)=-0.0125$, and $Q_F''(x_F^*)=10$. Near the oracle, the cutoff is $\ell(x)=-x/2$, with atom mass $1-x$. At the oracle, $p=0.95$ and $\tau=0.9/0.95=18/19$. The continuous demand distribution gives the isolation limits. The action set is compact, the loss is bounded and continuous, its slopes have magnitude $1/2$, and its action derivative exists almost surely at the oracle. These facts verify the assumptions of Proposition~\ref{prop:saa_concentration_rate}.}

{The one-sided derivatives of the empirical objective give the quantile optimality condition $\hat F_n(x-)\le(1-\delta)/2\le\hat F_n(x)$. Choosing the smallest such order gives $x^{\mathrm{SAA}}=\xi_{(k)}$, $k=\lceil n/20\rceil$, including a consistent selection when the empirical minimizer is not unique. Its Beta distribution yields all calculations in Table~\ref{tab:saa_finite}. Define the truncated moments
\[
M_j(h)=\frac{(k)_j}{(n+1)_j}I_h(k+j,n+1-k),\qquad
T_j(h)=\frac{(k)_j}{(n+1)_j}-M_j(h),
\]
where $(a)_j$ is the rising factorial, $(a)_0=1$, and $I_h$ is the regularized incomplete Beta function. These are $\mathbb E[(x^{\mathrm{SAA}})^j\mathbf1_{\{x^{\mathrm{SAA}}\le h\}}]$ and $\mathbb E[(x^{\mathrm{SAA}})^j\mathbf1_{\{x^{\mathrm{SAA}}>h\}}]$. In particular,
\[
\mathbb E Q_F(x^{\mathrm{SAA}})=5M_2(1-\delta)-\tfrac12M_1(1-\delta)+\tfrac12T_1(1-\delta)-\tfrac{1-\delta}{2}T_0(1-\delta).
\]}

{For the sample sizes in Table~\ref{tab:saa_finite}, the combined-loss quantile $t_n^{\mathrm{comb}}$ is in $(-1/2,0)$. Conditional on $x^{\mathrm{SAA}}=x$ and for $t$ in this interval, the loss CDF is zero when $x<-2t$ and is $1-x/2+t$ when $x>-2t$. Thus $t_n^{\mathrm{comb}}$ solves
\[
(1+t)T_0(-2t)-\tfrac12T_1(-2t)=0.9.
\]
For $h=-2t$, integrating the conditional positive part gives
\[
\begin{aligned}
J_n(t)&:=\mathbb E[(c(x^{\mathrm{SAA}},\xi)-t)^+]\\
&=\tfrac12M_2(h)-\tfrac12M_1(h)-tM_0(h)\\
&\quad+\tfrac18T_2(h)-\tfrac12tT_1(h)+\tfrac12t^2T_0(h).
\end{aligned}
\]
The combined-risk value is $t_n^{\mathrm{comb}}+J_n(t_n^{\mathrm{comb}})/(1-\delta)$. Subtracting $\mathbb E Q_F(x^{\mathrm{SAA}})$ gives SR, while $\mathrm{PS}=\mathbb E Q_F(x^{\mathrm{SAA}})+0.0125$. Numerical evaluation of these special functions and the scalar quantile equation produces the reported finite-sample values.}

{Finally, the score in~\eqref{eqn:saa_variance} is $\psi_F(\xi)=-1/2+10\mathbf1_{\{\xi<0.05\}}$. It takes the value $9.5$ with probability $0.05$ and $-1/2$ otherwise, so $\operatorname{Var}(\psi_F(\xi))=4.75$. Since $Q_F''(x_F^*)=10$,
\[
\sigma^2=\frac{\operatorname{Var}(\psi_F(\xi))}{Q_F''(x_F^*)^2}=0.05(0.95).
\]
Let $\Phi$ and $\varphi$ denote the standard normal distribution function and density. Then
\[
n\,\mathrm{PS}\to0.2375,\qquad
\sqrt n\,\mathrm{SR}\to
\frac{0.95(1/2)\sqrt{0.05(0.95)}\,\varphi(\Phi^{-1}(18/19))}{0.1}
\approx0.1112168.
\]
The limiting ratio is $\rho_n/\sqrt n\to0.1112168/0.2375\approx0.4682813$, and the premium share tends to one.}

\end{document}